%% file: nteasee_facct25.tex
\documentclass[sigconf]{acmart}
\AtBeginDocument{%
  }

\usepackage{multirow}
\usepackage{longtable}
\usepackage{pdfpages}
\usepackage{subcaption}

\copyrightyear{2025}
\acmYear{2025}
\setcopyright{cc}
\setcctype{by}
\acmConference[FAccT '25]{The 2025 ACM Conference on Fairness, Accountability, and Transparency}{June 23--26, 2025}{Athens, Greece}
\acmBooktitle{The 2025 ACM Conference on Fairness, Accountability, and Transparency (FAccT '25), June 23--26, 2025, Athens, Greece}\acmDOI{10.1145/3715275.3732160}
\acmISBN{979-8-4007-1482-5/2025/06}
\begin{document}


\title{Nteasee: Understanding Needs in AI for Health in Africa - \\A Mixed-Methods Study of Expert and General Population Perspectives}

\author{Mercy Nyamewaa Asiedu}
\authornote{Authors contributed equally to this research.}
\email{masiedu@google.com}
\affiliation{%
  \institution{Google Research}
  \country{USA}
}

\author{Iskandar Haykel}
\authornotemark[1]
\email{alexander.haykel@gmail.com}
\authornote{Work done while affiliated with Google.}
\affiliation{%
  \institution{Americans for Responsible innovation}
  \country{USA}
}

\author{Awa Dieng}
\email{awadieng@google.com}
\authornotemark[1]
\affiliation{%
  \institution{Google DeepMind}
  \country{Canada}
}

\author{Kerrie Kauer}
\email{kkauer@google.com}
\affiliation{%
  \institution{Google}
  \country{USA}
\email{}
}
\author{Tousif Ahmed}
\email{ahmedtousif@google.com}
\affiliation{%
  \institution{Google}
  \country{USA}
\email{}
}
\author{Florence Ofori}
\email{fofori@asu.edu}
\authornotemark[2]
\affiliation{%
  \institution{Arizona State University}
  \country{USA}
\email{}
}
\author{Charisma Chan}
\email{charismachan@google.com}
\affiliation{%
  \institution{Google}
  \country{Switzerland}
\email{}
}
\author{Stephen Pfohl}
\email{spfohl@google.com}
\affiliation{%
  \institution{Google Research}
  \country{USA}
\email{}
}
\author{Negar Rostamzadeh}
\email{nrostamzadeh@google.com}
\affiliation{%
  \institution{Google Research}
  \country{Canada}
\email{}
}
\author{Katherine Heller}
\email{kheller@google.com}
\affiliation{%
  \institution{Google Research}
  \country{USA}
\email{}
}

\renewcommand{\shortauthors}{Asiedu, Haykel, Dieng et al.}

\begin{abstract}
Artificial Intelligence (AI) for health has the potential to significantly change and improve healthcare. However in most African countries, identifying culturally and contextually attuned approaches for deploying these solutions is not well understood. 
To bridge this gap, we conduct a qualitative study to investigate the best practices, fairness indicators, and potential biases to mitigate when deploying AI for health in African countries. Additionally, we explore opportunities where AI could make a positive impact in health. 
We use a mixed-methods approach combining in-depth interviews (IDIs) and surveys. We conduct 1.5-2 hour long IDIs with 50 experts in health, policy, and AI across 17 countries, and, through an inductive approach, we conduct a qualitative thematic analysis on expert IDI responses. We administer a blinded 30-minute survey with case studies to 672 general population participants across 5 countries in Africa (Ghana, South Africa, Rwanda, Kenya and Nigeria), and analyze responses on quantitative scales, statistically comparing responses by country, age, gender, and level of familiarity with AI. We thematically summarize open-ended responses from surveys.
Our results find generally positive attitudes, high levels of trust, accompanied by moderate levels of concern among general population participants for AI usage for health in Africa. This contrasts with expert responses, where major themes revolved around trust/mistrust, ethical concerns, and systemic barriers to integration, among others. This work presents the first-of-its-kind qualitative research study of the potential of AI for health in Africa from an algorithmic fairness angle, with perspectives from both experts and the general population. 
We hope that this work guides policymakers and drives home the need for further research and the inclusion of general population perspectives in decision-making around AI usage.
\end{abstract}

\ccsdesc[500]{Computing methodologies~Artificial intelligence}
\ccsdesc[500]{Social and professional topics~Computing / technology policy}
\ccsdesc[500]{Human-centered computing}

\keywords{algorithmic fairness, bias, health equity, policy, need-finding, human experiences, norms, practices, participatory design}

\maketitle

\input{1_introduction}

\input{2_methods_surveys}

\input{3_methods_IDIs}

\input{4_results_surveys}

\input{5b_short_result_IDIs}

\input{6_discussion}

\input{7_conclusion}

\input{8_reflexivity}

\begin{acks}
The authors would like to acknowledge the insights and feedback from Maya Leventer-Roberts, Sunny Jansen, Chirag Nagpal and Vishwali Mhasawade. We would also like to thank Gutcheck for their rigor, timeliness and reach in administering and analyzing the general population surveys, and D'Well for their transcription services. This study was funded by Google LLC and/or subsidiary thereof (Google). All the authors are or were employees of Google and may own stock as a part of a standard compensation package.
\end{acks}

\bibliographystyle{ACM-Reference-Format}
\bibliography{references}

\appendix
\input{99_appendix}

\end{document}

%% file: 1_introduction.tex
\section{Introduction}
\label{1_introduction}

Increasingly, advancements in AI technologies including large language models (LLMs) are promising to have significant impact across  a variety of healthcare delivery and services, such as improved diagnostics and precision medicine, predictive analytics and preventative care, personalized treatment, telemedicine, remote monitoring, and improved healthcare accessibility  \cite{Topol2019, Goldberg2024, patel2017correlating, Azizi2021, Singhal2023,jiang2017artificial, Yagnik2024}. More research is actively being done to discover the best use cases for AI in healthcare along with how to actualize those use cases through research examining human experiences, needs, perceptions and interactions. However, much of this work focuses on high income countries (HICs), and relies on assumptions about healthcare systems, practices, and health attitudes, which in many instances may be understood as belonging to distinctly Global North contexts \cite{Wahl2018}. Conversely, in the context of low-and-middle income countries (LMICs), it is often the case that healthcare systems, contexts, practices, and prevailing health attitudes are importantly different from HICs in the Global North \cite{Yu2024}. While there is the potential for AI to revolutionize healthcare, overlooking geo-cultural and contextual knowledge could disproportionately and negatively impact human lives. 
This work aims to help fill this gap by investigating the best practices, fairness indicators, and potential biases to mitigate when deploying AI for health in African countries, as well as explore opportunities where AI could make a positive impact in health.

\input{figure_latex/overview}

\paragraph{Contributions:} In this study, we set out to explore what considerations may be relevant to deploying AI health solutions in low-and-middle income countries (LMICs) in Global South contexts. In that vein, our specific and original research focus is on the deployment of AI health solutions in African countries. The research questions we seek to answer include (1) what are currently the most significant determinants of health inequities, and perspectives on addressing them?; (2) what are present perceptions around AI and healthcare and opportunities for development?; (3) what are the perceived or envisioned effects of colonial history on AI?; (4) what are algorithmic fairness and bias considerations for African contexts specifically?; and (5) what are community-driven approaches to representative data generation and design and deployment of AI systems?

In order to answer these questions, we interviewed experts across the machine learning, healthcare, and health equity spheres, working within or around AI and health in African countries. In doing so, we drew out their insights across a variety of themes and dimensions involved in developing, deploying, and regulating AI solutions for health in African countries. We also deployed survey questionnaires with general population participants from five countries in Africa–Ghana, Rwanda, Nigeria, Kenya and South Africa–interrogating their attitudes towards AI health solutions including the use of case studies around large language models. We compiled and analyzed these insights through an inductive approach, presenting them for a multi-stakeholder audience, namely technology and health policymakers, healthcare professionals, and machine learning developers and deployers. To our knowledge, this is the first study that utilizes rigorous mixed-methods research with multinational, multidisciplinary experts and general population participants on AI for health in the Global South and to provide specific, actionable, best practices, and geo-contextual fairness and bias considerations.

Our work offers several novel insights regarding specific contextual challenges and opportunities for AI use in health in Africa. We suggest methods to build trust in AI systems to support adoption in the African context. Additionally, we identify that while experts have concerns about AI which may be indeed valid, general population participants are more accepting of AI adoption, emphasizing the need for policy makers to consider general population perspectives. This finding challenges the common assumption that the general population agrees with expert notions of data colonialism, fairness, and use of internationally developed AI tools. 

This work advances a picture of the landscape of AI for health in African countries, delineating experience, needs, and perceptions of AI tools, which we hope may aid stakeholders in navigating the path towards efficient and impactful evaluation, governance and integration of AI health solutions across the African region. 

This being said, we acknowledge the limitations of our study and the insights it offers, among them the circumscribed range of African nations and linguistic and cultural contexts we investigate. Our aim through this study is to advance the discussion on the integration of AI solutions within such contexts, and beyond this to galvanize further critical research supporting the successful integration of AI health solutions across the Global South,  and in doing so advance the discourse on AI, equity, and global health.
 
\section{Related Work}
Given that most of the historical data, models and problems in AI for healthcare are explored in Global North contexts, Global South geo-contexts are relatively underexplored. To address this issue, a slew of recent research has sought to advance our understanding of the challenges and solutions for successful development and integration of AI into Global South health systems. At the cross-regional level, \citet{Okolo2022, Kong2023, Chakraborty2024, Zuhair2024} advance insights from Global South contexts for improving the AI health regulatory landscape and for integrating AI into underserved and under-resourced health systems. At the regional level,  \citet{Sambasivan2021} have observed that mainstream conceptions of algorithmic fairness for model evaluation (both in health contexts and beyond) require retailoring when considering AI solutions in India. Bhatt et al. \cite{Bhatt2022} charted a framework for re-contextualizing natural language processing (NLP) fairness research in an Indian context, with the caveat that this can apply to other Global South geo-contexts. Concerning the African region specifically, research for successful AI integration into health systems traverses several subject-matter areas. The question of AI policy readiness and development \cite{Arakpogun2021,Gwagwa2021, otaigbe2022scaling, Etori2023, Diallo2024} have seen much attention recently, with some common themes being the need for robust data infrastructure, the importance of context-specific AI readiness assessments, and the critical role of international collaboration and capacity building in ensuring equitable AI development across the African continent. Better understanding challenges and solutions for AI in African health systems has also been a major point of focus, cutting across a range of topics such as addressing health inequities \cite{Qoseem2024}, enhancing existing healthcare resources and infrastructure \cite{Owoyemi2020, Oladipo2024, Tshimula2024}, and improving healthcare accessibility \cite{Sallstrom2019, Wakunuma2022, Mbunge2023}, especially for under-resourced rural areas \cite{Guo2018, Arefin2024}. In a related vein, there is a growing interest in the prospect of \textit{algorithmic colonization}, where imported AI technologies from the Global North fail to align with local realities and even contribute to the perpetuation of Global South disparities \cite{Mohamed2020, Birhane2020, Asiedu2024}. The concern over algorithmic colonization highlights the growing importance of developing decolonial approaches to AI integration \cite{Adams2021, Ayana2024}, which are not only technically sound but also optimized for cultural attunement and equitable access.

The current literature underscores both the potential and challenges of AI in transforming healthcare in the African region. Previous work has solicited expert stakeholders to help identify conditions for developing AI solutions tailored to African contexts \cite{Lopes2021}. Nevertheless, there remains a critical need to further include and grow the range of experts involved in building our understanding of policy and development considerations for successful AI integration in African health systems. Similarly, it is vital to also include the perspectives of general population stakeholders, who represent the ones ultimately directly affected by the use of AI health technologies.

%% file: figure_latex/overview.tex
\begin{figure*}[!h]
         \centering
         \includegraphics[width=.9\textwidth]{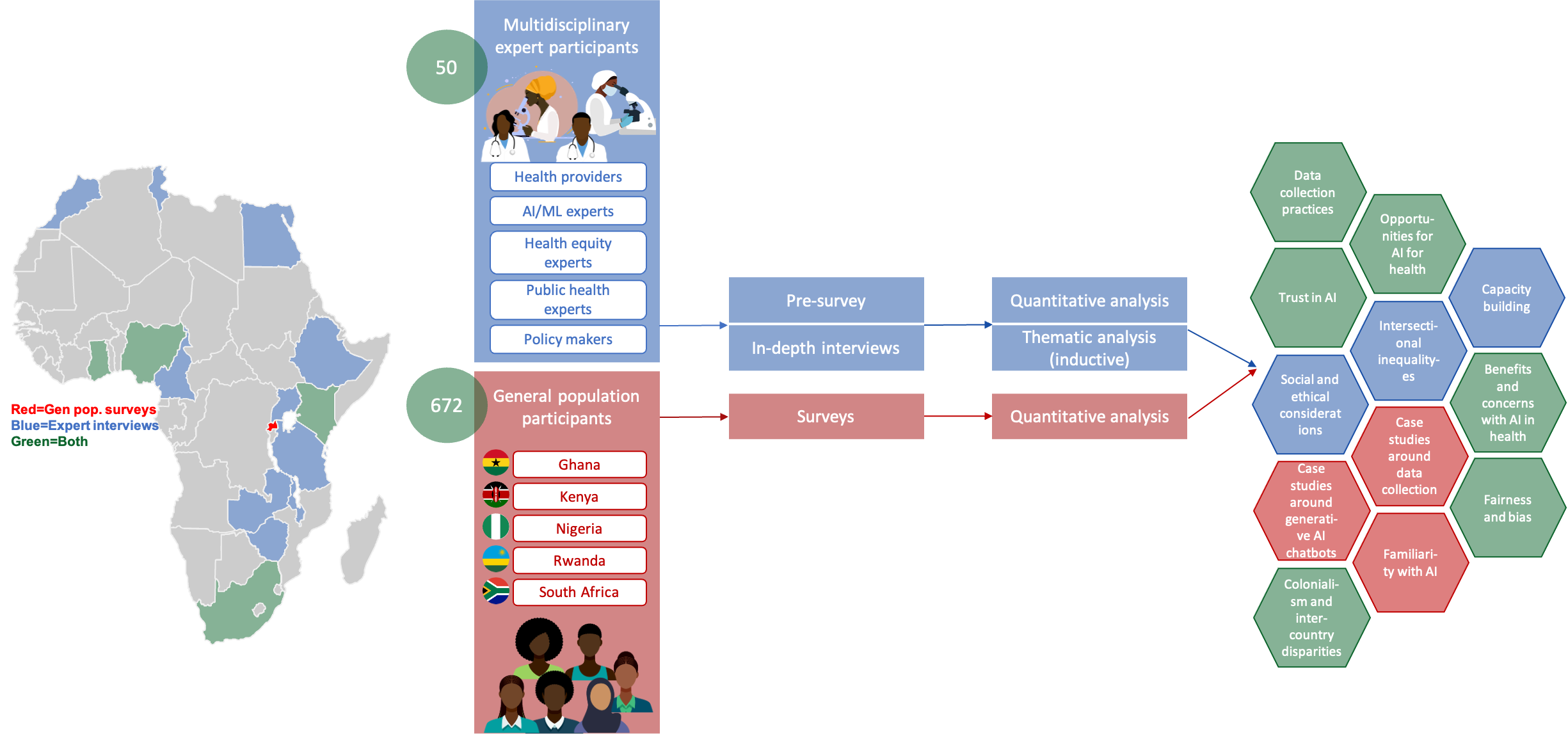}
        \caption{Nteasee* study overview of participant distribution within Africa, methods and summary of findings. \footnotesize {Experts residing in countries outside of Africa were also interviewed as was one expert in Mauritius, though this is not reflected in the figure. \\ * Nteasee is an Akan Twi word from Ghana which means “to understand”. It is also an Adinkra symbol (visual illustrations that represent proverbs, concepts and aphorisms) which stands for cooperation and understanding \cite{adjei2018adinkra}.}}
        \label{fig:overview}
\end{figure*}

%% file: 2_methods_surveys.tex
\section{Methods}
\label{methods}
\input{tables/participant_demographics}

We conducted a survey of general population (Gen Pop.) participants and in-depth interviews (IDIs) with pre-interview surveys of expert participants to understand perceptions of AI for health in Africa. While health is a broad area and it is important to perform this study across different specific sub-specialties such as radiology, oncology, etc, we focus on a broad approach of health due to the paucity of original mixed-methods research in this space. This approach allowed the expert interviewees the flexibility to discuss which specific applications of AI they believe are most urgent and needed in the African context. Further details of the two arms of the study are in the subsequent sections. We analyze the survey and interview results independently and also analyze the intersections of the outcomes, comparing general population viewpoints to expert viewpoints.  \vspace{-.4cm}

\subsection{General population surveys}

\subsubsection{\bf Survey overview and participants}
This part of the study was conducted via a blinded online survey facilitated by GutCheck, between November 14 to November 21, 2023. The survey was conducted in five countries in Africa: Ghana (n=125), Rwanda (n=125), Nigeria (n=169), Kenya (n=125), and South Africa (n=128), with a total of 672 general population respondents completing the survey. The sample size was pre-determined to achieve a margin of error of 7\%-8\% at a 95\% confidence level for each country, and a margin of error of 3\%-4\% at 95\% confidence level for the total population (all countries combined). Enrollment was completed once we reached the sample size per country.

The survey was conducted in English only. Specific eligibility criteria for the respondents were (1) 18+ years of age (2); Have attained at least a high school diploma or equivalent; (3) Attest that they were able to read and understand complex and technical text in English “very well” (4); Be at least somewhat familiar with the term “AI”/Artificial Intelligence. These eligibility criteria were set to ensure we were obtaining high quality, well informed responses. The survey took an average of 30 minutes to complete. For quality control, the vendor incorporated automated speeder and straightliner checks in place. They also conducted manual checks on the open-ended questions, read through the responses as they were coming in and removed respondents that entered gibberish, responses that were very short, responses clearly indicating respondents were not paying attention, or responses which were identical to those of another respondent (indicating the use of a bot or survey farm). The participant demographics for the general population survey can be found in Table \ref{tab:participant_demographics}.

\subsubsection{\bf Survey procedure}
Respondents were first screened for eligibility. We asked eligible respondents to provide demographic information and then proceeded to ask questions about understanding of and familiarity with AI and ML and its benefits, trust in AI interventions for health, algorithmic fairness and biases, and intersections between AI, health, and colonialism. Additionally, we studied use cases of large language models (LLMs) due to their emerging relevance. Towards this, we conducted case studies - providing hypothetical scenarios of a user interacting with an LLM chatbot for health to understand conditions for acceptance, useability, and believability. We provided versions of these scenarios that were augmented to include location, gender and religion, to understand whether the above metrics changed with additional contextual responses. Finally we asked respondents to indicate specific non-personal hypothetical questions they would ask an AI chatbot about health. The full set of questions asked in the survey can be found in Appendix \ref{supp:survey_ques}.

\subsubsection{\bf Statistical analysis}
 We performed ANOVA to analyze statistical differences in responses by (1) country, (2) gender, (3) age range, and (4) level of familiarity with AI. We present the overall outcomes of the responses in the main text of the paper but detail the statistical comparison outcomes in Appendix \ref{supp:per_country_genpop}.

%% file: tables/participant_demographics.tex
\begin{table}[!htbp]
    \footnotesize
    \centering

    \caption{\textbf{ Participant Demographics for Experts IDI and General Population Survey Respondents}}
        \begin{tabular}{p{0.15\linewidth} | p{0.25\linewidth} | p{0.2\linewidth} p{0.2\linewidth}}
        \toprule
        Demographic & Subgroup & Expert Participants (n=50) & Gen Pop. Participants (n=672) \\
        \midrule
        \multirow{1}{2cm}{Gender}
        & Female & 19 (38\%) & 271 (40.3\%)  \\
        & Male & 30 (60\%) & 401 (59.7\%) \\
        & Non-binary & 1 (2\%) & 0 \\
        
        \midrule
        \multirow{1}{2cm}{Age}  
        & 18-20 & 0 & 56 (8.3\%) \\
        &  21-29 & 11 (22\%) & 330 (49.1\%)  \\
        & 30-39 & 25 (50\%) & 201 (29.9\%) \\
        &  40-49 & 12 (24\%) & 66 (9.8\%) \\
        & 50-59 & 2 (4\%) & 14 (2.1\%) \\
        
        \midrule
        \multirow{1}{2cm}{Countries of Residence} 
        & South Africa & 5 (10\%) & 128 (19.0\%) \\
        &  Kenya & 8 (16\%) & 125 (18.6\%) \\
        &  Nigeria & 6 (12\%) & 169 (25.2\%)\\
        &  Ghana & 3 (6\%) & 125 (18.6\%) \\
        &  Rwanda  & - & 125 (18.6\%) \\
        &  Other countries in Africa (e.g., Egypt, Tunisia, and others) & 17 (34\%) & - \\
        & Other countries outside Africa (e.g. USA, Germany)
          & 11 (22\%)& - \\

        \midrule
        \multirow{1}{2cm}{Highest level \\ of education} 
        & High school or less & 3 (6\%) & 105 (15.6\%) \\
        &  Associate’s & 1 (2\%) & - \\
        &  Bachelor's & 9 (18\%) & 393 (58.4\%) \\
        &  Masters & 21 (42\%) & 72 (10.7\%) \\
        &  Doctorate  & 8 (16\%)& 5 (0.7\%) \\
        &  Professional (eg.JD, MD) & 8 (16\%) & - \\
        &  Other (eg. Vocational)
          & - & 97 (14.4\%) \\
        
        \midrule
        \multirow{1}{2cm}{General \\ population \\ employment \\ status} 
        & Employed & -& 308 (60.2\%) \\
        & Self-Employed & - & 156 (12.5\%) \\
        &  Student & - & 125 (17.2\%) \\
        &  Others & - & 83 (12.4\%) \\
        
        \midrule
        \multirow{1}{2cm}{Expert \\ specialty} 
        & AI/ML & 12 (24\%) & - \\
        & Medical Doctors & 12 (24\%) & - \\
        &  Public health & 7 (14\%) & - \\
        &  Other digital health & 19 (38\%) & - \\
        
        \midrule
        \multirow{1}{2cm}{Expert years \\ of relevant \\ work \\ experience} 
        & 1-2 years & 8 (16\%) & - \\
        & 3-5 years & 8  (16\%) & - \\
        & 5-10 years & 13 (26\%) & - \\
        & 10+ years & 20 (40\%) & - \\

         \bottomrule
    \end{tabular}
\label{tab:participant_demographics}
\end{table}

%% file: 3_methods_IDIs.tex
\subsection{Expert in-depth interviews}
\subsubsection{\bf IDI Expert Participant Selection}
To recruit experts, we reached out to people in our networks using an email draft which included a link to fill out a participation interest survey for the interviews. 
We also shared the study description and fliers 
with a link/QR code to the form in AI communities and group chats, and on social media pages including LinkedIn. Participants were recruited between October 24 2023 and January 30th 2024.

Since the in-depth interviews (IDIs) specifically focused on experts, we recruited participants who had expertise in AI and health, practicing health providers, public health researchers, health policy experts, and entrepreneurs. We had 445 interested participants sign up to take part in the study. We went through five rounds of short-listing to narrow down the participants based on responses in the participant interest survey. Participants were selected based on (1) years of experience, prioritizing \textgreater 10 years of experience, followed by 5+ years of experience in their respective fields, (2) field of practice–we ensured there was a balanced representation of clinicians, public health, and AI researchers, (3) focus of research/work on Africa, and (4) country of residence, prioritizing African residents. We also included non-African residents who were African diaspora and/or had \textgreater 10 years of experience working in Africa. Demographics of the participants as well as relevant criteria such as years of experience can be found in Table \ref{tab:participant_demographics}. Expertise in AI Trust/safety/responsible AI was not a selection criteria for experts, and none of the experts we interviewed indicated specifically that they were in the field of AI Trust/safety/responsible AI.

\subsubsection{\bf IDI Procedure}

After agreeing to informed consent, the participants filled out a brief demographic survey. 
The interview was then scheduled and then conducted online over Google Video Conference (GVC) and audio-recorded. There were two individual moderators throughout the total qualitative interviews, and each interview was conducted with 1 moderator and 1 participant. During the interview, a moderator engaged with the participants in in-depth, semi-structured interviews that lasted approximately 90-120 minutes. Before each interview, the interviewer thoroughly explained the interview process, emphasized confidentiality and addressed any questions or concerns raised by the participants. After each interview, the voice recordings were saved as MP3 files and no identifying data (\textit{i.e.}, participants’ imagery) was retained. Voice recordings were transcribed by a vendor (D’ Well), and the transcripts stored in a secure data bucket without any PII. The full interview guide can be found in Appendix \ref{supp:interview_ques}.

The focus of the interviews explored the participants’ perspectives on (a) equity in healthcare; (b) historical biases in health; (c) AI in healthcare; (d) algorithmic fairness framing and applications; (e) stereotypes, sensitive attributes, and vulnerable sub-populations; and (f) an understanding of existing definitions of fairness. Further, we explored (g) perceptions of colonialism and the Global North/South divide. 

\subsubsection{\bf Data Analysis}
The team met either weekly or bi-weekly for six months to discuss the data process, the insights that were emerging, and engaged in critical discussion to ensure that the analysis codes accurately reflected the data. The research team proceeded through several steps during the data analysis that included familiarization with the data, open coding, and axial coding. Each member of the thematic analysis team engaged in a round of open coding. 
In the second phase, each member then engaged in axial coding and the categories that emerged from open coding were further developed and refined by exploring the relationships between them. The research team discussed all axial codes and higher-order themes until consensus was achieved. Lastly transcripts were swapped such that each was reviewed by at least 1 other researcher to ensure trustworthiness and reliability. Two of the researchers finalized codes and conducted a category audit to ensure standards of trustworthiness and reliability were met. Through this iterative process five higher order themes and subthemes emerged from the data.

%% file: 4_results_surveys.tex
\section{Results}
\label{sec:results_surveys}

\subsection{General population quantitative survey results}
 
\subsubsection{\bf Trust, benefits and concerns on AI tools:} 
\begin{figure}[!htbp]
    \centering
    \begin{subfigure}{\columnwidth}
        \centering
        \includegraphics[width=\linewidth]{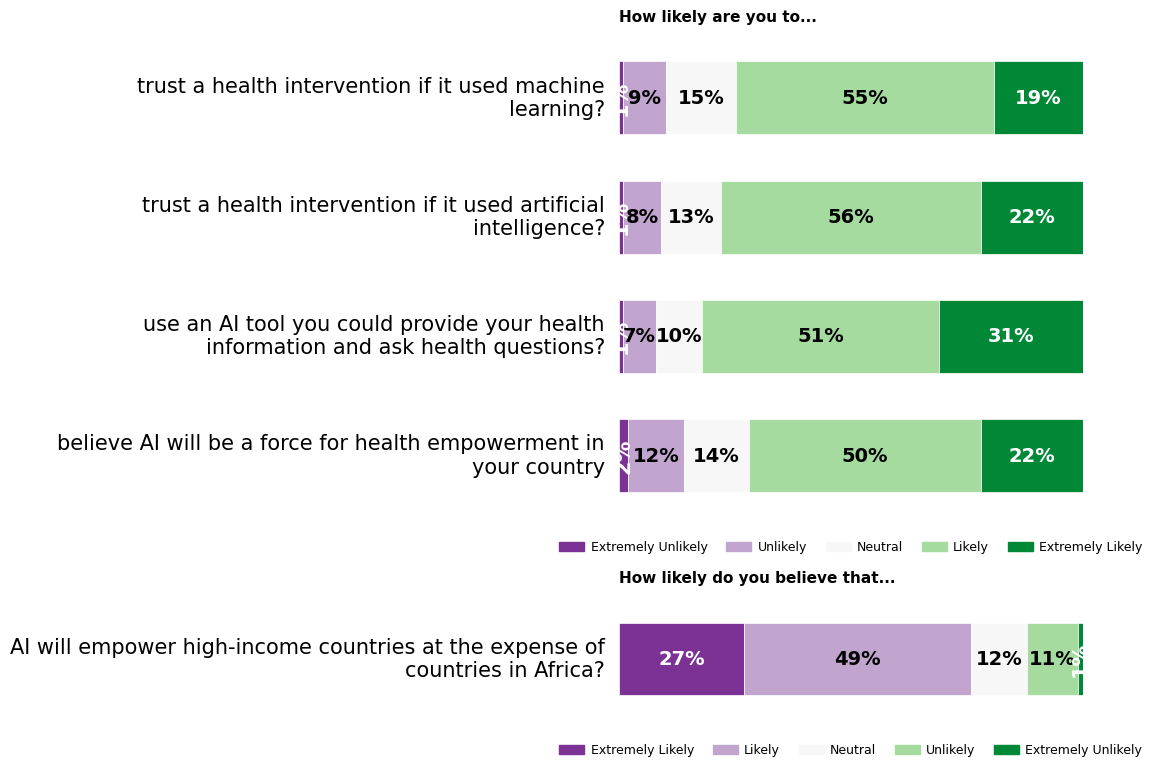}
        \caption{}
    \end{subfigure} 
    \vspace{1em} 
    \begin{subfigure}{\columnwidth}
        \centering
    \includegraphics[width=.95\linewidth]{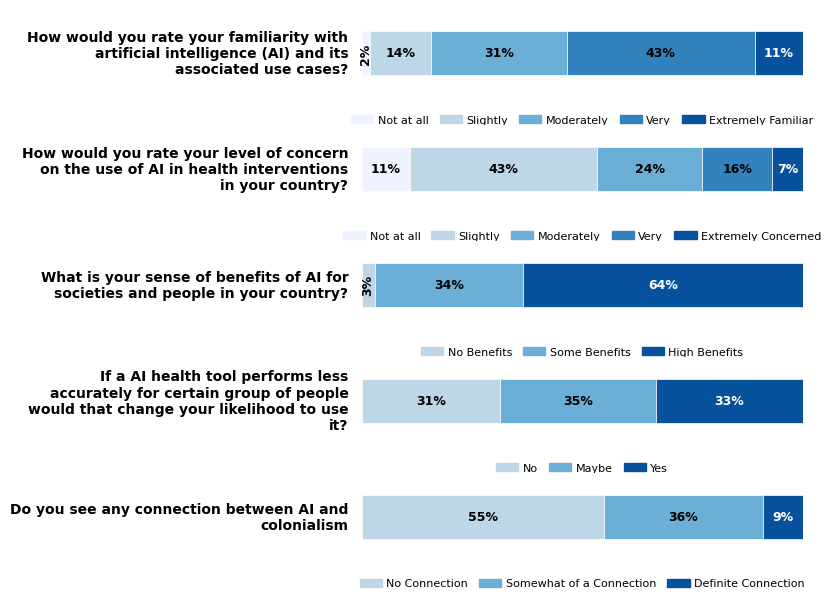}
        \caption{}
        \vspace{-.8cm}
    \end{subfigure}
    \caption{General population responses to trust in AI, benefits, concerns, country-level disparities, and colonialism. a) Responses
to questions poised in a likert scale format b) Responses to questions poised in a semantic differential scales format.}
\label{fig:genpopsurvey}
    \vspace{-.5cm}
\end{figure}

Gen Pop. respondents had high levels of trust for AI definitions with most indicating they were likely or extremely likely to trust a health intervention if it used AI (74-78\%). Seventy-two percent of the Gen Pop. believed that AI  was likely or extremely likely to be a force for health empowerment in their countries and about 98\% indicated that AI for health had benefits for people in their country. However, while acknowledging the potential benefits of AI,  89\% also acknowledged concerns around AI use for health with 47\% indicating that they were moderately, very or extremely concerned around the use of AI for health. When asked whether they would use an AI tool that had been shown to perform less accurately for a group of people, only 33\% indicated they would still use it, with 35\% indicating they might use it depending, and 31\% saying they would not use it at all (Fig.\ref{fig:genpopsurvey}).

\subsubsection{\bf Benefits to country, Inter-country disparities and colonialism:} While most participants (98\%) generally indicated that AI potentially had high benefits (64\%) or some benefits (34\%) for societies and people in their country, 66\% also believed that AI would empower high-income countries at the expense of countries in Africa, with 45\% of Gen Pop. indicating that they saw some connection between AI and colonialism (Fig.\ref{fig:genpopsurvey}). In terms of benefits, making tasks faster and easier was the most dominant response - ``\textit{AI presents the opportunity to make life easier, quicker and generally better by the elimination of human errors}'' [Ghana, 21-29]. Most concerns revolved around access to knowledge of AI- ``\textit{Most Africans do not know much about new technologies and this will greatly affect the use of AI in my country}'' [Nigeria, 30-39] , and job loss ``\textit{It means job losses. If an AI or robot can do what a human being can. It means there will be less need for human labor}'' [South Africa, 30-39], though there were indications by other respondents that AI will also create job- ``\textit{AI may automate certain tasks, it also has the potential to create new jobs and industries.}''[Nigeria, 30-39]

\subsubsection{\bf Case study around perceptions of AI chatbots for health}
We provided respondents with a case study on LLM interactions. Specifically, respondents were shown a question by a ``user'' posed towards an LLM, and the response by an LLM model. Respondents were shown a base question describing the user’s symptoms and request for advice about what to do next and the LLM response to the question. They were asked to assume they were the user and respond to useability and adherence to the LLM recommendations. Note that the Gen Pop. participants are not considered experts and obtaining their responses on whether or not the LLM answered the question and level of accuracy was primarily to gauge the perceptions users had towards the LLM generated response. These responses do not provide an indication of how accurate the LLM was. Figure \ref{fig:casestudy_chatbot_a} shows the results as a percentage with 61\% of respondents indicating the response was fully accurate, and 48\% indicating it fully answered the question (Fig. \ref{fig:casestudy_chatbot_a}). However, a significant portion of respondents (52\%) indicated it only partially answered the question (48\%) or did not answer the question at all (4\%), while 38\% indicated the response was only partially accurate and 1\% indicated it was not accurate. Most respondents (79\%) indicated that the diagnosis was in the list of suggestions. On usability, most respondents (64\%) described the length of the response as just right, while 33\% indicated it was too long. Only 3\% felt the response was too short (Fig. \ref{fig:casestudy_chatbot_a}). 

\input{figure_latex/casestudy_chatbot}

Respondents were then provided augmentations to the questions that included the location of the user and the change in LLM response based on the location (Fig. \ref{fig:casestudy_chatbot_b}). Most Gen Pop. respondents (54\%) felt that the response including location context was helpful, 38\% indicated it was somewhat helpful and 8\% indicated it was not helpful. Respondents were then provided augmentations that added gender and religion of the user to the question and the corresponding LLM contextual response (Fig. \ref{fig:casestudy_chatbot_b}). Sixteen per cent indicated this addition was not helpful, 39\% indicated it was somewhat helpful or not helpful, and 45\% indicated that it was very helpful. We looked at adherence to the LLM recommendations on next steps and found that irrespective of augmentation (none/base, location, gender and religion), most would follow LLM recommendations to see a doctor, nurse, or other health practitioner as next steps (23\%-35\%), followed by getting plenty of rest (14\%-19\%), and drinking plenty of fluids (13\%-17\%) (Fig. \ref{fig:casestudy_chatbot_c}). Taking over-the-counter medications came next with the base response, however with local augmentation recommendation, asking neighbors or friends was the next most popular response, and with gender and religious augmentation, seeking spiritual guidance from religious leaders was the next most popular response Fig. \ref{fig:casestudy_chatbot_c}. Only about 5-8\% indicated they would seek information from elsewhere, indicating high trust in LLM generated recommendations.

\subsubsection{\bf Case studies around data collection projects for development of AI tools:} 
\begin{figure*}[!ht]
     \centering
     \includegraphics[width=.9\textwidth]{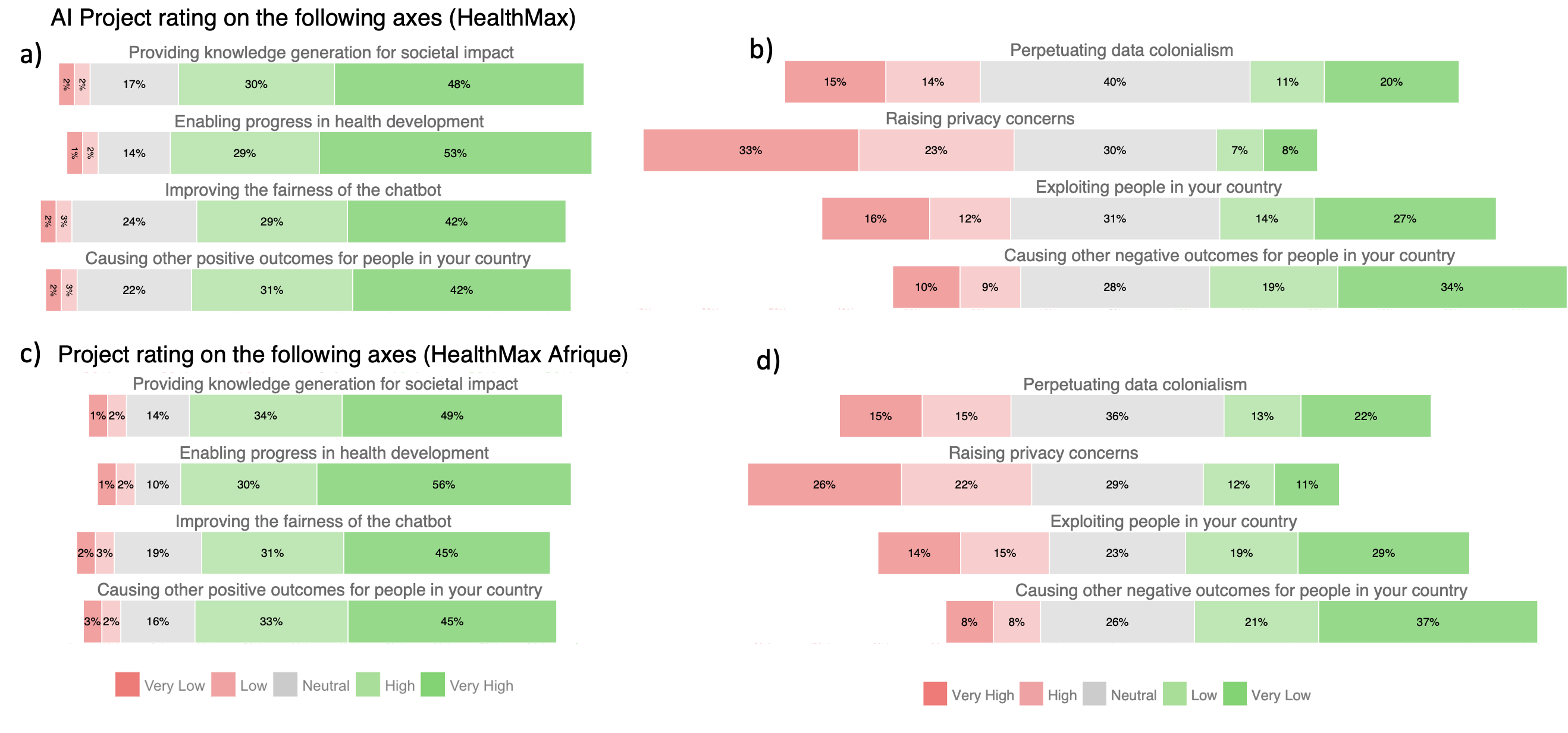}
    \caption{\textbf{Participant responses to case studies around data collection projects for development of AI tools.} Respondents were randomly assigned to rate HealthMax, an international organization (a,b) or HealthMax Afrique, a local organization (c,d), along positive axes (a,c) and negative axes (b,d)}
    \vspace{-.3cm}
    \label{fig:healthmax}
\end{figure*}

We provided respondents with a scenario in which an institute/company was collecting data from people in their country to develop AI tools. See Appendix \ref{supp:survey_ques} for specific questions. Respondents were randomly divided into an arm that was told that the institution was an international organization (HealthMax) and another arm that was told that the institution was a local organization based in their country (HealthMax Afrique). Respondents were asked to indicate how likely they were to participate in the project and then asked to rate the project along positive axes - societal impact, health development, improving fairness, other positive outcomes, and along negative axes - data colonization, privacy concerns, exploitation and other negative outcomes. Overall, Gen Pop. respondents viewed the projects positively regardless of local vs. international branding (Fig. \ref{fig:healthmax}). There was an average of 5\% increase in positive ratings when the project was branded locally (Fig. \ref{fig:healthmax}a,c). The highest change in negative ratings were on privacy concerns with 56\% of respondents rating privacy concerns as high or very high for HealthMax international, compared to 46\% of respondents for HealthMax Afrique. There were no significant differences in negative ratings along the lines of colonialism, exploitation, and other negative outcomes between local and international (Fig. \ref{fig:healthmax}b,d).

 \subsection{Expert responses to pre-interview surveys}
 \subsubsection{\bf Expert responses on best and worst data collection practices:} Prior to participating in the in-depth interviews, experts were asked to fill out pre-surveys where they listed best and worst data collection practices. Responses are summarized in Table \ref{tab:best_worst}.
 
 \begin{table}[hbt!]
    \footnotesize
    \centering
    \caption{\textbf{Expert best and worst data collection practices}}
        \begin{tabular}{p{0.45\linewidth} | p{0.45\linewidth}}
        \toprule
        \textbf{Best Practices} & \textbf{Bad Practices}\\
        \midrule
        Working with local personnel/in-country researchers/on-the-ground organizations & Tokenistic involvement of in-country researchers (not as true collaborators)\\
       \midrule
        Engaging/consulting with local communities to sensitize them to the reason and use of data collection & Biased data from crowdsourcing hospital data, online data sourcing\\
       \midrule
        Obtaining informed consent and ethical considerations and approvals, country specific policies and guidelines & Taking data without consent, or with improper consent/withholding information\\
      \midrule
        Considering cultural differences in data collection practices  & Exploiting/underpaying data labellers/manipulation\\
      \midrule
        Developing better quality tools for data collection & Coercion  \\
        \midrule
        Ensuring that data collected provides benefits to local respondents & Models inaccessible to locations where data has been sourced from \\
       \midrule
       Representative wide variety of samples from different regions in Africa and within countries & Over-reliance on community health workers\\
     \midrule
        Appropriate compensation/incentives for data use &  Lack of information to respondents on data use \\
      \midrule
        Adequate data security, participant  confidentiality, and privacy & Lack of involvement of local specialists with local knowledge\\
    \bottomrule
    \end{tabular}
\label{tab:best_worst}
\end{table}

\subsubsection{\bf Expert responses to the impact of current barriers to AI for clinical usage:}
Clinical experts who were interviewed were also asked pre-interview survey questions around how they would use AI developed in a different country to diagnose their patients. They were told the limitations of the model were that (1) it requires historical health records to reach the best performance it can, (2) it was developed using radiology images from a different device than what you use, (3) it has been shown to perform worse for African-Americans. Experts were asked how they would approach using the tool and which limitations would impact their decision the most. Most experts indicated that they would be cautious but evaluate it for their specific use cases or use it for more general use cases. Some specifically mentioned that experts, particularly clinicians in Africa, are used to working with foreign-developed software tools, and tend to use them only as a preliminary diagnosis/guide, and make sure to take into account local contexts - ``\textit{All the software and medical device we use have been developed in USA or Europe…The outputs are indicative only and assists in formulating a preliminary diagnosis.}'' [Medical Doctor, 30-39, Mauritius], ``\textit{I would be cautious in my use of the tool and would not use it for critical cases. I would use it for cases that are more routine and in situations where the cost of misdiagnosis would not be so dire.}'' [Medical Doctor, 21-29, Ghana].
For most experts, the option that would most impact their decision to use the tool would be that it performs worse for African Americans (n=9 clinicians), as most considered their patients to be ``\textit{demographically and genetically similar to African Americans}''. The second most impactful option would be that the tool requires historical health records to reach the best performance it can (n=4), due to the prevalence of paper-based records and lack of electronic health records making it challenging to obtain digital historical data. However, experts indicated that some countries do have EHRs and historical data. Experts generally were least concerned about the data coming from a different device as they felt confident that algorithms could more easily generalize across devices.

%% file: figure_latex/casestudy_chatbot.tex
\begin{figure}[!htbp]
\centering
  \centering
  \includegraphics[width=\columnwidth]{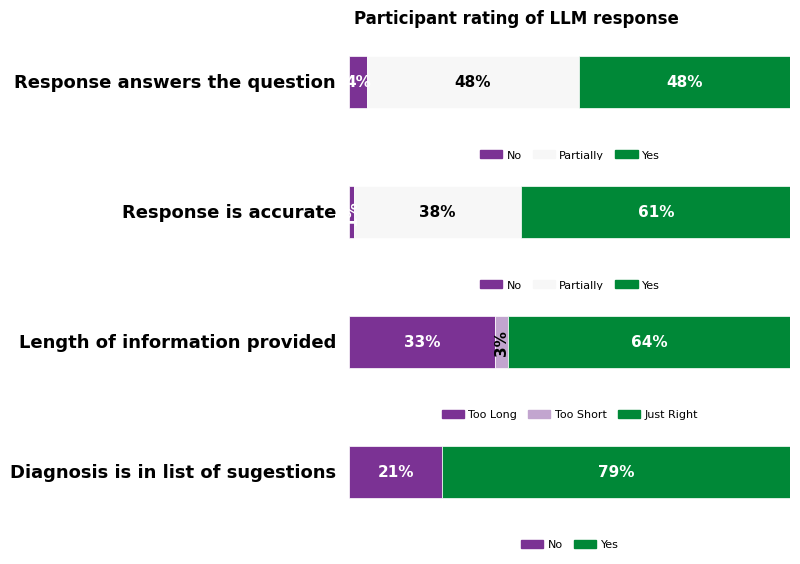}
  \caption{\textbf{Gen Pop. participant responses to chatbot health recommendations:}  Respondents rating LLM responses for answer level, accuracy, length of information, and whether or not the diagnosis is in the list of suggestions. \vspace{-.5cm}}
\label{fig:casestudy_chatbot_a}
\end{figure}%
\begin{figure}[!htbp]
  \centering
  \includegraphics[width=\columnwidth]{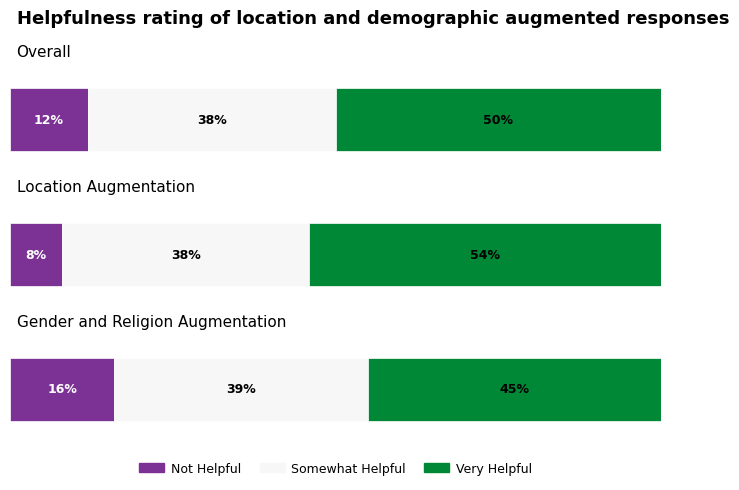}
  \caption{\textbf{Gen Pop. participant responses to chatbot health recommendations:} Rating of helpfulness of the added local augmentations and religion and gender augmentations}
  \vspace{-.3cm}
\label{fig:casestudy_chatbot_b}
\end{figure}

\begin{figure*}[!htbp]
    \centering
    \includegraphics[width=.9\textwidth]{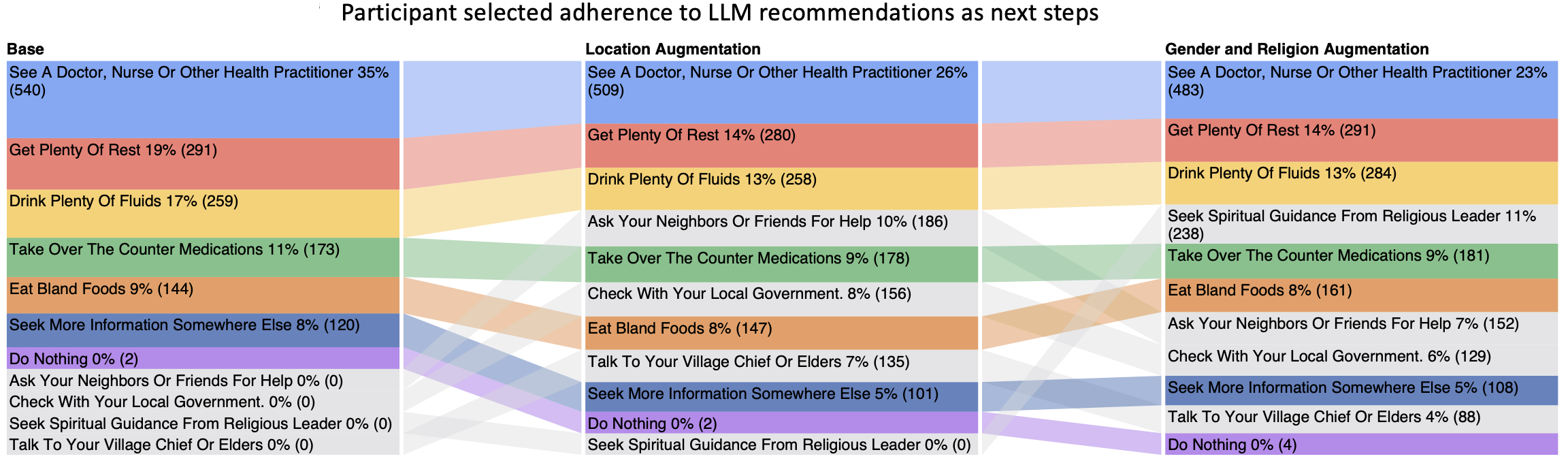}
        \caption{\textbf{Gen Pop. participant responses to chatbot health recommendations:} Adherence level to LLM-provided recommendations as next steps and how they change with different augmentations. Note that the Gen Pop. population are not considered experts and their responses on whether or not the LLM answered the question and level of accuracy was primarily to gauge the perceptions users had towards the LLM-generated response and not to provide indication of how accurate the LLM was. Note: the last 4 options in the base recommendations of (c) did not appear in the base follow up options. The last option in the location augmentation did not appear in the location followup options.}
     \label{fig:casestudy_chatbot_c}
\end{figure*}

%% file: 5b_short_result_IDIs.tex
\label{sec:thematic_analysis}
\subsection{Findings of expert qualitative interviews}
Five major themes emerged from the inductive analysis of expert IDIs namely (1) trust in AI in Africa, (2) opportunities for AI to address health in Africa, (3) intersectional inequities and systemic factors impacting health equity in Africa, (4) social and ethical considerations for developing and deploying  AI solutions in health in Africa, and (5) capacity building for effective implementation and adoption of AI solutions in health in Africa. The following section provides a summary of each theme and sub themes with representative quotes. A full writeup of the in-depth thematic analysis can be found in \ref{supp:full_themes}.

\subsubsection{\bf Trust In AI in Africa}
Participants overwhelmingly shared perceptions around sources of mistrust, citing historical marginalization, colonialism, misunderstanding of AI technologies, and a lack of transparency, fairness and accountability as sources. As P25, a Medical Doctor from Nigeria put it, \textit{"a lot of times things that are seen to be coming from the Global North are viewed with a lot of skepticism."} Many participants also highlighted how the legacy of colonialism has fostered a belief that Western innovations, even those aimed at improving healthcare may have ulterior motives, such as continued control, or data exploitation. As P006, a Public Health Researcher in Tanzania stated,\textit{..."so it can also be perceived that way, that these people are still trying to continue to control us, tell us what to do, predict how things are going to be in terms of our health and everything”.} Through the lens of colonialism and marginalization in Africa, other themes related to misunderstanding of technology and lack of transparency were salient among these experts. As P030 a public health research observed, \textit{“There is a lot of misinformation and myths around AI”}. These baseline concerns over the use of AI in healthcare are exacerbated by perceptions of biases and lack of accountability with automated systems. In order to tackle these concerns around trust, participants identified the need to build trust for AI use through community-driven approaches, and they proposed methods to foster trust through commitments, education, and good data practices. Most participants stressed the importance of involving local communities to ensure culturally and contextually relevant solutions for AI for health in Africa, which will facilitate adoption of AI. This entails involving communities directly in the problem scoping and development phase, as stated by P059, a Community Health Worker from Kenya  \textit{"if the communities are engaged at the point of the system then they can provide background information on the region the communities come from, the kind of cases that have been there, the kind of environment they live in and even the cultural perspectives that affect health"}. The data illustrated how essential it is to address concerns about data ownership and sovereignty, ensuring data is used responsibly, and that building trust must include establishing long-term commitments, engaging with stakeholders to showcase the benefits of using AI tools in health, and providing assurance of the effectiveness of these models. 

\subsubsection{\bf Opportunities for AI to Address Health in Africa}
Another  major  theme  that  emerged  was  the  expert  suggestions  of  opportunities  for  AI  to address health concerns in Africa, encompassing their views on the desired features of AI solutions for better adoption. As one public health researcher from Nigeria explained, \textit{ “I think that will be a game changer instead of taking time off these repetitive administrative tasks that the healthcare system [is] currently burdened with and allowing health workers to work at the scope of practice”.} Overall, experts highlighted AI’s potential to democratize healthcare, namely the idea that AI could offer greater autonomy to African communities, reducing reliance on outside support by providing tools for self-diagnosis, prevention, and community-based care.
P038, a Machine Learning Practitioner/Engineer summarized this sentiment, stating, \textit{“We  believe  that  providers  and  the  AI  can  work  really  well  together. The AI is not looking to replace their expertise, but when they work together, they’re actually augmenting each other”.  }Participants suggested opportunities for AI to address health concerns in Africa, encompassing their views on the desired features of AI solutions for better adoption. In addition to being contextually relevant, participants emphasized that accuracy and consistency  are  crucial  factors  before  deploying  AI  tools  in  health.

\subsubsection{\bf Intersectional Inequities and Systemic Factors Impacting Health Equity in Africa}

Experts described various factors, both individual and systemic, that impact health equity in Africa, as well as proposals for addressing existing inequities. Experts broadly identified social determinants of health, infrastructure limitations, and socioeconomic disparities stemming from the history of colonialism as key contributors to health inequities. As P018,a Medical Doctor, noted, \textit{“Income is a major factor that contributes to health inequities. People who are poor are less likely to have access to healthcare, and they are less likely to receive
quality  healthcare  when  they  do  have  access”}. A common theme from the experts centered around the a social-economic stratified system where lower economic areas could greatly benefit from AI in healthcare, but concerns about equitable models in poorer communities was a top of mind concern. These were further compounded by the intersection of other historically oppressed and marginalized identities and belief systems. For example, gender identity was frequently cited as a source of inequity in healthcare access. As P073, a Medical doctor in Nigeria summarized, \textit{"There are some places where the cultural norms do not allow for equitable access to quality healthcare. So for instance, we have some cultures that prohibit a woman . . . to seek healthcare in the health facilities”}. The education disparities, in addition to diversity of languages spoken on the continent, also led to further disenfranchisement from groups who do not speak the same language or have access to education, creating additional barriers to healthcare. Some  participants also mentioned inequity in healthcare access for sexual minorities, mostly due to stigma and discrimination: \textit{``when you actually think about the LGBTQ communities, they actually face a lot of health inequities in general because of stigma and judgment. Those are the key population groups,  I think, in Africa that are impacted by the health inequities''} - P038 [30-39, USA, Machine Learning Practitioner/Engineer]. Race, ethnicity and disability status also emerged as axis of identity that were negatively impacted by healthcare and several participants expressed hope and healthy skepticism for the role AI could play in mitigating these forms of discrimination. Experts stressed that while targeted interventions can address specific barriers like geographical limitations or lack of funding, there is a need for a comprehensive approach to tackle health inequities in Africa in a sustainable way. The role of governments and private institutions were also seen as critical to addressing health inequities in Africa. Participants, for the most part, pointed to the critical role of government and policy in addressing health inequities in Africa. Many called for better health resource allocation in order to \textit{“avail the healthcare service throughout the population on equitable basis” }(P008 [30-39,  Ethiopia,  Public Health Researcher]). Participants also described AI-aided decision making as a solution for improved health outcomes.

\subsubsection{\bf Social and Ethical Considerations} Experts articulated the importance of addressing social and ethical factors when introducing AI technologies in healthcare in Africa. One such considerations is the need to develop localized solutions that consider social behaviours and cultural sensitivities. This includes \textit{“understanding the context like the social behavior, the context of healthcare in that particular country, the leadership structure as well” }(P006 [30-39,  Tanzania,  Public Health Researcher]) and \textit{“making sure the model eliminates all these”} (P006 [30-39, Tanzania, Public Health Researcher]). Many participants highlighted the necessity of developing localized solutions that are sensitive to social behaviors, advocating for community engagement throughout the development and deployment of AI solutions. As P038, a Machine Learning Practitioner/Engineer noted, \textit{“One of the things is having someone local on the ground.  I do believe that if you’re going to have a program that’s deployed in a particular community, you need someone who’s actually familiar with that community involved in the program right from the get go”. } Experts also emphasized addressing existing inequities by deploying AI equitably and prioritizing language accessibility. The data underscores that the safe and effective implementation of AI in healthcare in Africa hinges on ethical considerations, bias mitigation strategies, establishing clear regulatory frameworks, and addressing data representativeness, scarcity, collection and privacy. Many participants stressed the need to ensure that AI is deployed in a way that reaches vulnerable populations who are most in need of the technology, as well as taking steps to ensure accessibility in terms of language. P010 [30-39, Kenya, Research and Policy Analyst] noted how AI solution providers need to \textit{“really look at having a balance between market profitability and how best to ensure that, other than profits, . . . how does this really get into the general population who need it the most”.} 

\subsubsection{\bf Capacity Building for Effective Implementation and Adoption of AI Solutions in Health in Africa}
Experts highlighted strategies to overcome the challenges associated with developing and integrating AI into the healthcare sector in Africa. These strategies encompass enhancing data collection and infrastructure, bridging the digital literacy gap, fostering strong local institutions, and implementing comprehensive training programs and awareness campaigns. This includes promoting open data initiatives and encouraging data sharing among the scientific community as P028 Head of Engineering in South African stated, \textit{“I think the open data movement of trying to promote, publishing more data and making that widely available, can go a long way to building better data sets”.} Experts indicated that such efforts are crucial for nurturing the African AI ecosystem and building public trust for successful AI adoption in Healthcare in Africa. Investing in local data infrastructure, including “cloud hosting” and data storage units was a common sentiment among participants. For P056, Public Health Researcher, \textit{“we need to develop our own databases, our own data storage units to store our own data in Africa, then we’ll be able to use this data to develop algorithms that will . . . benefit the local communities”.} Most participants believe that collaboration across stakeholders – including users, local communities, and educational institutions – is vital in driving this development. As P010 a Research and Policy Analyst from Kenya articulated, fostering partnerships and networking across these groups is essential for \textit{“helping grow the capacities in Africa, the knowledge”.} A USA based Digital Health Lead for an NGO echoed this sentiment, asserting that establishing strong \textit{“local institutions and research, whether public or private”, is “a key driver towards...successfully implementing and maintaining AI”}

%% file: 6_discussion.tex
\section{Discussion} \label{sec:discussion}
In this work, we conducted a qualitative study that leveraged in depth-interviews and surveys to understand expert and general population perspectives regarding the use of AI for health and healthcare applications in Africa. Our study emphasizes the importance of community-driven approaches to the design, implementation, and evaluation of AI systems for health in Africa that are grounded in the local contexts in which such systems are intended to be used. 

Our study identified expert and general population perspectives on factors contributing to health inequities in Africa and the potential for AI applications in healthcare to improve population health and reduce health inequities in Africa. A major recurring theme identified relates to the need for contextually-grounded, culturally-aware, and community-driven approaches. For example, this includes the need to understand specific social and structural causes of health inequities, including those related to infrastructure, healthcare access, and local political, regulatory, and governmental environments \cite{Qoseem2024, Owoyemi2020, Oladipo2024}; the need for a participatory approach to problem formulation grounded in the needs of the specific communities and contexts in which a system is intended to be used \cite{Martin2020, Suresh2024, Passi2019}; as well as contextualized approaches to conceptualizing, measuring, and mitigating issues related to algorithmic fairness, bias, and transparency \cite{Sambasivan2021, Asiedu2024}. Regarding the potential for AI to promote health equity, participants noted that AI systems and interventions designed to address specific clinical needs (\textit{e.g.}, tools for patient registration, clinical decision support, or diagnostics), structural barriers (\textit{e.g.}, healthcare access), and context-specificity (\textit{e.g.}, augmentation of LLM prompts) may be effective. Furthermore, participants highlighted that, in order to realize this potential, it is critical to address complex structural and social challenges related to infrastructure and capacity building, trust and community attitudes in the context of colonial history, as well as technical limitations related to biases in datasets and AI systems.

The study identified critical gaps in perspectives on issues between general population respondents and expert interview participants, indicating a need for both participatory inclusion of general population perspectives in the design and evaluation of AI systems and a need for broader education regarding potential harms and failure modes. For example, we find that while a majority of general population survey respondents reported a high-degree of trust in AI systems used for health applications and believed AI was likely to be a force for health empowerment, expert interviewees attributed the untrustworthiness and potential for harm of AI to several factors, including colonial history, marginalization, and lack of transparency, fairness, or accountability. Similarly, while the majority of general population respondents did not feel that an AI tool would perform differently for them compared to other people, expert survey respondents indicated that all attributes studied were likely to exhibit AI tool performance disparities and provided detailed exploration of contributing factors through the in-depth interviews.

\paragraph{\bf Call to Action}: Based on our findings, we present a call to action for AI practitioners, policy makers and researchers in Africa and around the world to consider when deploying AI tools for health in Africa.
\begin{itemize}
    \item {\it Increase the use of participatory research at a local level}: Participatory research at the local level is the most direct way to develop AI solutions that are grounded in the local contexts and address issues around trust, effective solutions, data representativeness and ethical data collection. Such participatory approach can help bridge the gap between expert concerns and general population optimism thus fostering better adoption. 
    \item {\it Fund and develop infrastructure for local data centers and compute, given the sensitive nature of health data}: Local infrastructure is paramount to address concerns around data sovereignty, ownership, security and privacy as well as improve access to representative datasets for building more unbiased tools.    
    \item {\it Build capacity to develop home-grown AI for health datasets and tools}: this is key to building culturally aware AI solutions that incorporate local knowledge and customs. 
    \item {\it Apply the expert-sourced recommendations we delineate in Table \ref{tab:best_worst} for data collection practices}
    \item {\it Improve education and awareness of both AI opportunities and locally-relevant fairness considerations.}: this is important to bridge the gap between expert concerns and general population optimism towards AI in healthcare in Africa. 
\end{itemize}

\paragraph{\bf Limitations}: A limitation of our work is the restriction of the population surveyed to one that possessed advanced English comprehension with a relatively high degree of technical literacy and education. We believed this to be necessary in order to engage most effectively with the survey. However, we acknowledge that this may limit the extent to which the findings of this study generalize. As such, it is important that future work extends our approach with more accessible methods such as on-the-ground verbal interactions that seek to understand the perspectives of those not well-represented in our study, and extend it to local languages. Furthermore, as our study primarily leverages qualitative insights from interviews and surveys, an important area of future research involves integration of these findings with real-world evidence of the relationship between specific axes of disparities and AI-enabled health applications.

%% file: 7_conclusion.tex
\section{Conclusion} \label{sec:conclusion}
This work demonstrates perspectives on AI for health in Africa, contrasting experts and gen. pop insights on opportunities, fairness and bias considerations and thought cases around the use of LLM-enabled chatbots. We find generally positive, optimistic outlooks on AI for health from general population participants, whereas expert participants express both optimism and  caution.  We hope that this work and the materials used in it as building blocks for future research that can expand this work on geographic locations, and population subgroups.

%% file: 8_reflexivity.tex
\section{Reflexitivity}
Two researchers work external to the researching organization, but have a combined extensive background in bioethics, philosophy, and AI for social good in Africa. Five of the other researchers work internally at the organization in various departments. One has an academic background in Sociology and Cultural Studies, one has a background in biomedical engineering, AI and global health, three have backgrounds in machine learning, statistics and algorithmic fairness in machine learning. We also varied in age, race, national origin, and sexual orientation which influenced our experiences with, and perceptions of, health and AI. Two of the co-first authors are of African origin, as is one of the additional authors. Three of the study's authors are of South Asian and/or Middle Eastern origin. Three of the study's authors are of North American origin.

%% file: 99_appendix.tex
\clearpage
\appendix
\renewcommand{\thesection}{A.\arabic{section}}
\renewcommand{\thefigure}{A.\arabic{figure}}
\renewcommand{\thetable}{A.\arabic{table}}
\renewcommand{\theequation}{A.\arabic{equation}}

\setcounter{section}{0}
\setcounter{figure}{0}
\setcounter{table}{0}
\setcounter{equation}{0}
\setcounter{page}{1}

\noindent \textbf{\LARGE{Appendix}}\\
\normalfont

\section{General population quantitative survey results}
\subsection{AI definitons and familiarity}
\begin{figure*}[!htp]
     \centering
     \includegraphics[width=\textwidth]{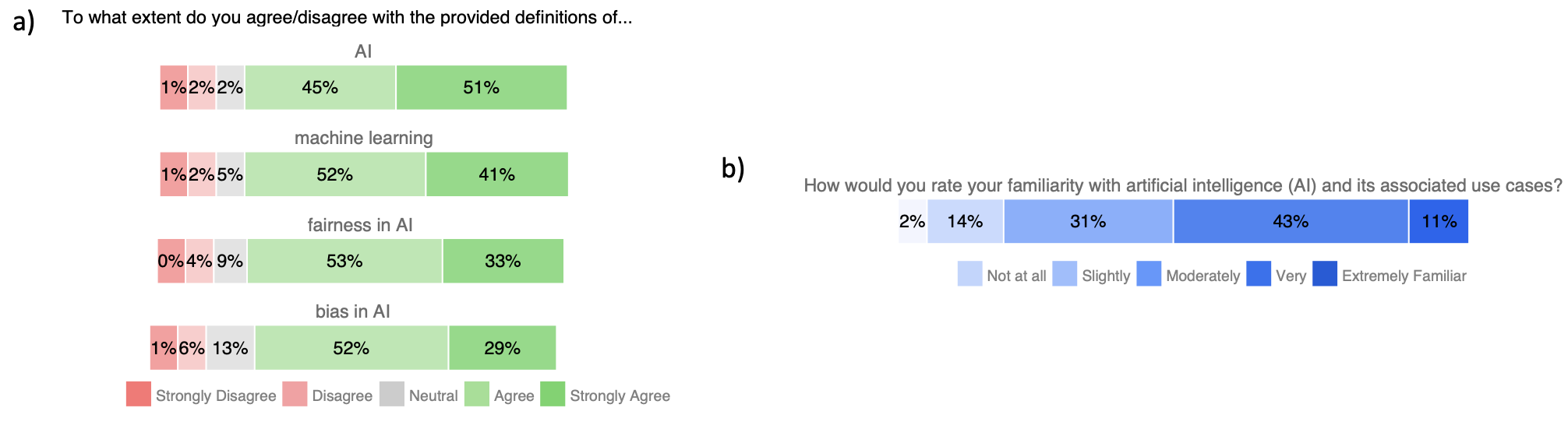}
    \caption{General population responses to likert scale questions on agreement with AI definitions and familiarity with AI. a) Responses to questions poised in a likert scale format b) Responses to questions poised in a semantic differential scales format.}
    \label{genpopsurveyapp}
\end{figure*}

\paragraph{Agreement with provided definitions:} We found that the Gen Pop. respondents generally agreed (81-96\%) with the standard definitions provided for AI, Machine Learning, Fairness and bias (Fig. \ref{genpopsurveyapp}). Please see Appendix \ref{supp:survey_ques} for the specific definitions provided. Of the provided definitions, bias in AI had the highest levels of strongly disagree, disagree and neutral. 
 
\paragraph{Familiarity with AI and use cases:} Most of the Gen Pop. respondents were familiar with AI and its associated use cases with 11\% indicating they were familiar with AI and use it regularly, 43\% indicating that they had read a lot about AI and used it themselves, 31\% indicating they were moderately familiar with AI but have not read much about it or used it themselves, 14\% indicating they had heard of AI conceptually but had not used it, and only 2\% indicating that they were not familiar at all  (Fig.\ref{genpopsurveyapp}b). We purposefully screened for respondents who had some familiarity of AI and could therefore provide knowledgeable, informed answers/opinions to the follow on questions. However this also biases our results against the section of the population which may be less educated and less aware of AI.

\subsection{Types of questions respondents would ask an AI Chatbot:} Respondents  were asked to optionally list 3 examples of questions they would ask an AI chatbot. This generated 520 questions which were analyzed and grouped into categories in a word cloud (Fig. \ref{fig:wordcloud}). The top categories around which questions fell into were 1) achieving health goals, \textit{e.g.}, ``\textit{...I want to start a fitness routine. What exercises and workout schedule would be most effective for achieving my specific health goals.}'' [Rwanda, 30-39, Male]; (2) causes of diseases, \textit{e.g.}, \textit{``\textit{A sharp pain in the upper left side of the body. What are the possible causes and solutions?}''} [Ghana, 30-29]; and (3) symptoms of diseases, \textit{e.g.}, \textit{``\textit{What are the symptoms of a certain medical condition?}''} [Kenya, 21-29].

\input{figure_latex/wordcloud_healthques}

\subsection{Comparisons of expert and general population survey on AI axes of disparities}
\paragraph{Axes of disparities:} Experts were asked to rate how likely different potential axes of disparities would cause AI to perform differently for sub-groups in Africa while Gen Pop. respondents were asked to rate how likely different demographics would cause AI to perform differently for the participant themselves. Results are reported in Fig. \ref{fig:axes_disparities}. Overall most experts (71\%) indicated that all demographics were likely or very likely to be axes of AI disparities/AI biases for sub-groups in Africa, with literacy, race, language, rural vs. urban location, and national income level being the top 5 axes of disparities. Most Gen Pop. respondents did not feel that AI tools would perform differently for them compared to other people. In agreement with experts, Gen Pop. indicated that literacy had the highest level of causing an AI tool to perform differently. Given that our respondents were mostly from educated backgrounds, 32\% indicated that their literacy level would allow an AI tool for health to perform better for them. Country of residence, country of birth, race, and colonial history of country of origin had the highest percentage of respondents indicating that they would cause an AI tool to perform worse for them (Fig. \ref{fig:axes_disparities}). 
\input{figure_latex/axes_of_disparities}
\newpage

\input{appendix_materials/appendix_latex/ai_def}

\input{appendix_materials/appendix_latex/ml_def_agree}
\input{appendix_materials/appendix_latex/ai_familiarity}
\input{appendix_materials/appendix_latex/ai_benefits}

\input{appendix_materials/appendix_latex/ai_benefits_famil}

\input{appendix_materials/appendix_latex/ai_health_concerns}
\input{appendix_materials/appendix_latex/trust_aiml}
\input{appendix_materials/appendix_latex/likelihood_kasa}
\input{appendix_materials/appendix_latex/ai_colonial}

\input{appendix_materials/appendix_latex/ai_empower}
\input{appendix_materials/appendix_latex/ai_empower_hic}
\input{appendix_materials/appendix_latex/fairness_def}
\input{appendix_materials/appendix_latex/bias_def_agreement}
\input{appendix_materials/appendix_latex/ai_bias_percep}

\input{appendix_materials/appendix_latex/healthmax_part}
\input{appendix_materials/appendix_latex/thought_case_1}
\input{appendix_materials/appendix_latex/chatbot_ques_length}
\input{appendix_materials/appendix_latex/chatbot_nextsteps}
\input{appendix_materials/appendix_latex/thought_case_2}
\input{appendix_materials/appendix_latex/location_helpful}
\input{appendix_materials/appendix_latex/location_nextsteps}
\input{appendix_materials/appendix_latex/thought_case_3}
\input{appendix_materials/appendix_latex/gender_religion}
\input{appendix_materials/appendix_latex/gender_religion_nextsteps}

\input{4b_survey_comparison}

\clearpage
\input{5_results_IDIs}

\newpage
\section{Study materials} 
\subsection{survey questions} 
\label{supp:survey_ques}
\includepdf[pages=-]{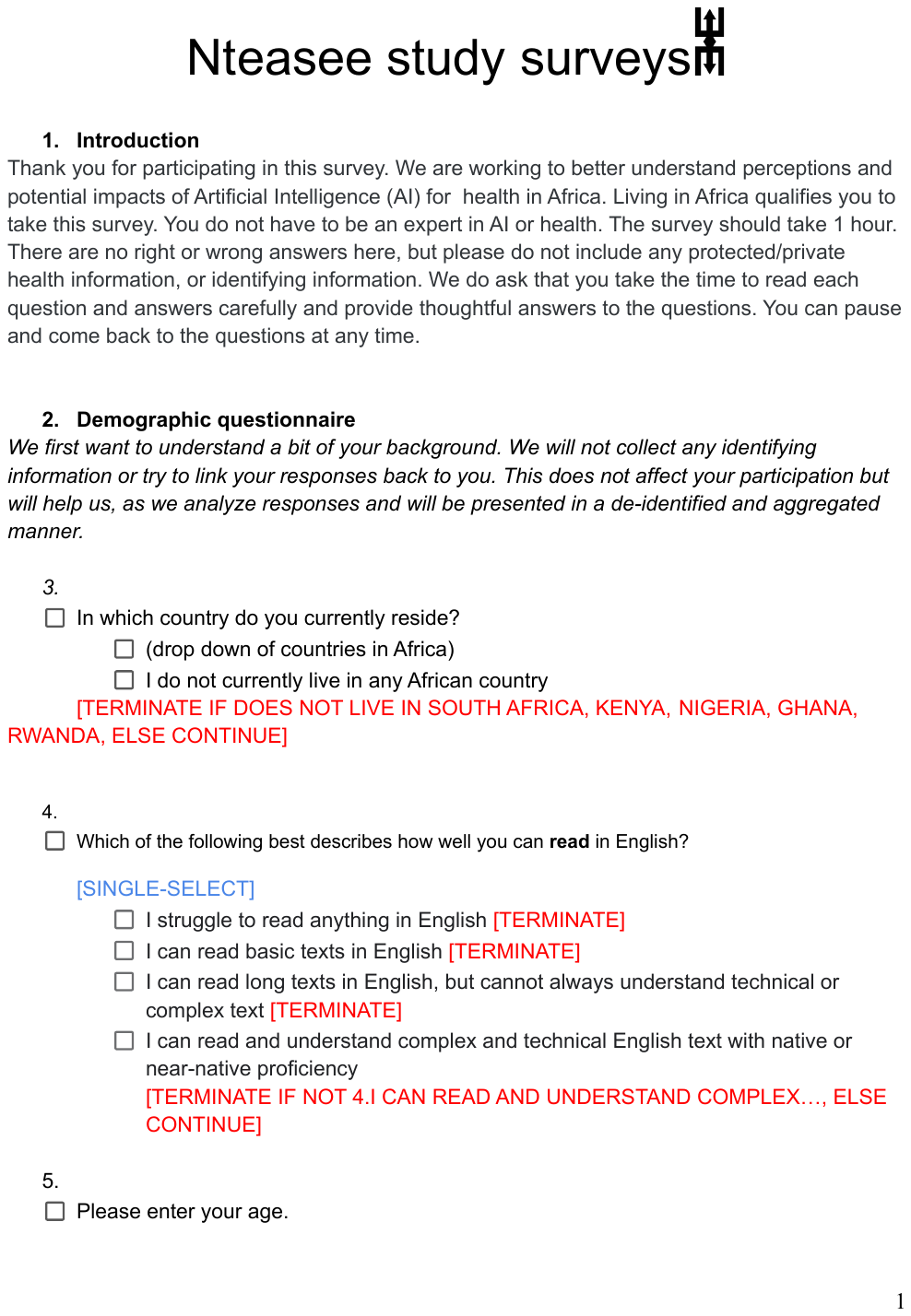}

\subsection{In depth interview guides} 
\label{supp:interview_ques}
\includepdf[pages=-]{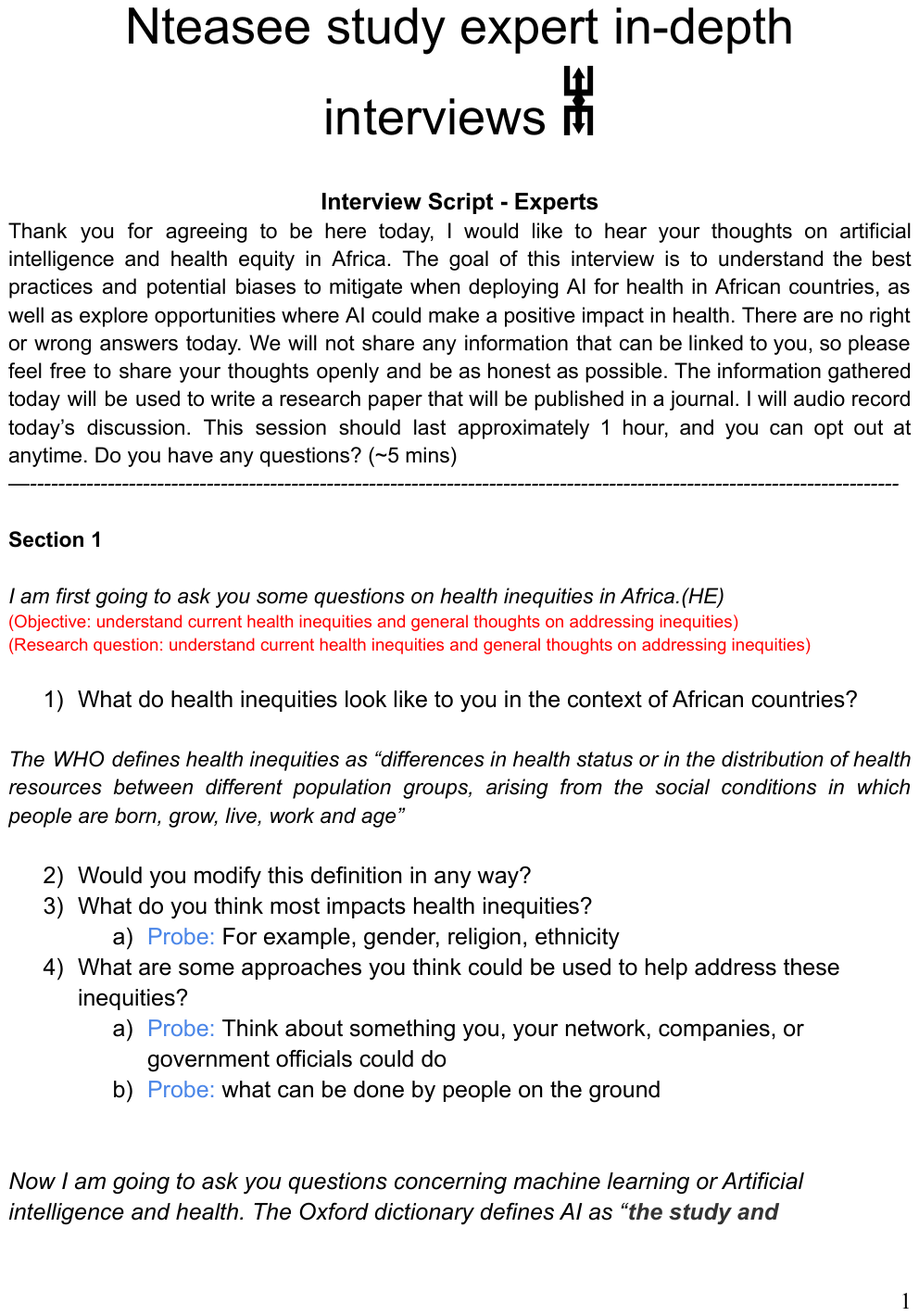}

%% file: figure_latex/wordcloud_healthques.tex
\begin{figure*}[!htbp]
         \centering
         \includegraphics[width=.8\textwidth]{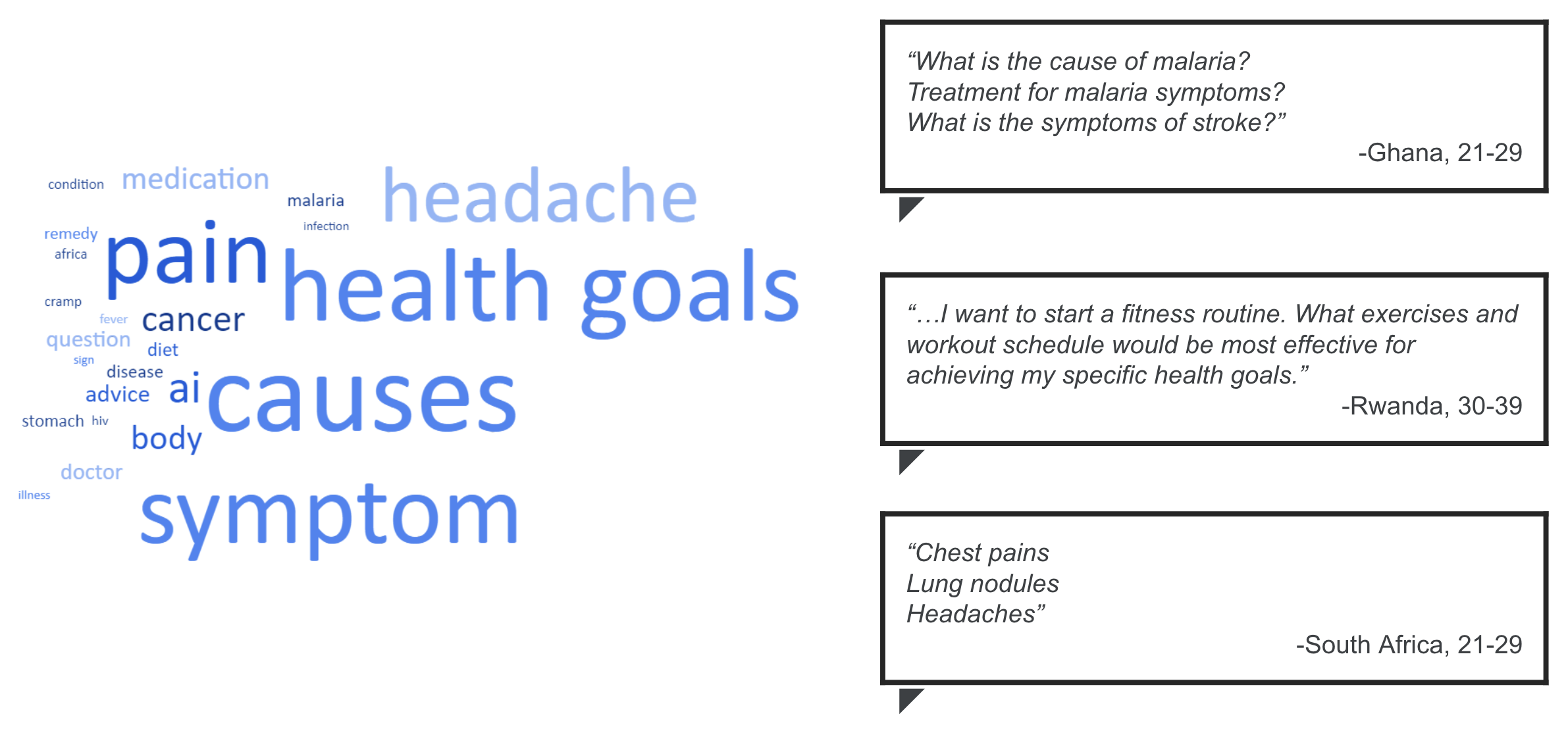}
        \caption{Word cloud distribution of question types that Gen Pop. respondents would ask an AI chatbot, with examples of questions submitted. Larger size indicates high occurrence of questions type.}
        \label{fig:wordcloud}
\end{figure*}

%% file: figure_latex/axes_of_disparities.tex
\begin{figure*}[!htbp]
         \centering
         \includegraphics[width=0.95\textwidth]{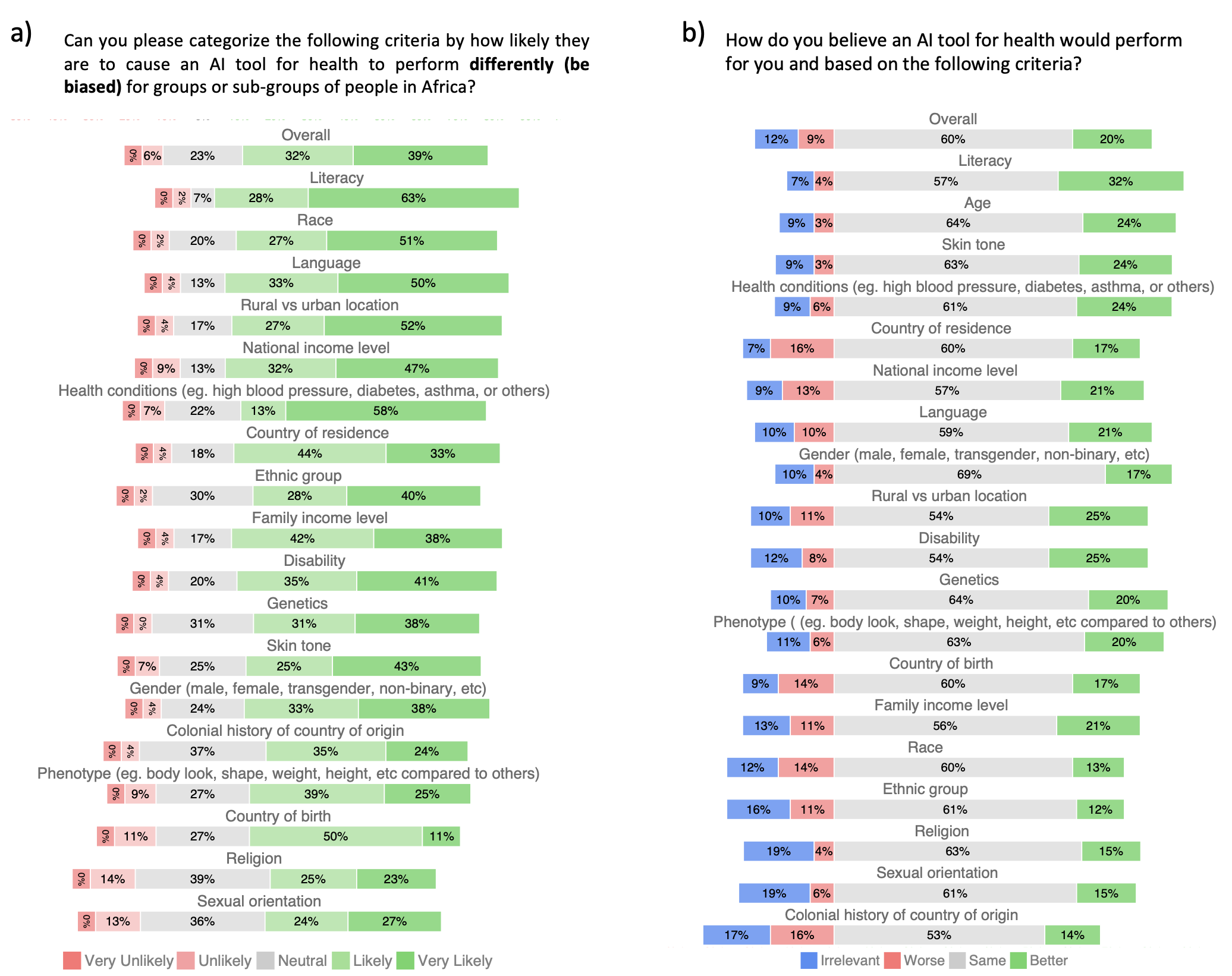}
        \caption{Perceptions from a) experts and b) Gen Pop. respondents on how likely different demographics would cause an AI tool to perform differently for people in Africa. Note that \textit{Better} is not necessarily good, it reflects
        that our participant population is better off than other people in the same country. e.g., Most of our survey participants are skewed towards high literacy and education levels compared to the general population, and so AI may perform better for them than for others.}
        \label{fig:axes_disparities}
\end{figure*}

%% file: appendix_materials/appendix_latex/ai_def.tex
\section{Per country comparison of Gen Pop survey results} 
\subsection{Per country comparison of Gen Pop survey results} 
\label{supp:per_country_genpop}
Per country general population responses to likert scale questions on agreement with AI definitions, familiarity with AI, trust, benefits, concerns, country-level disparities, and colonialism by country. 
\textcolor{red}{\textit{A/B/C/D/E}} Indicates statistical significance at 95\% confidence.

\begin{figure*}[ht]
         \centering
          \includegraphics[width=1\textwidth]{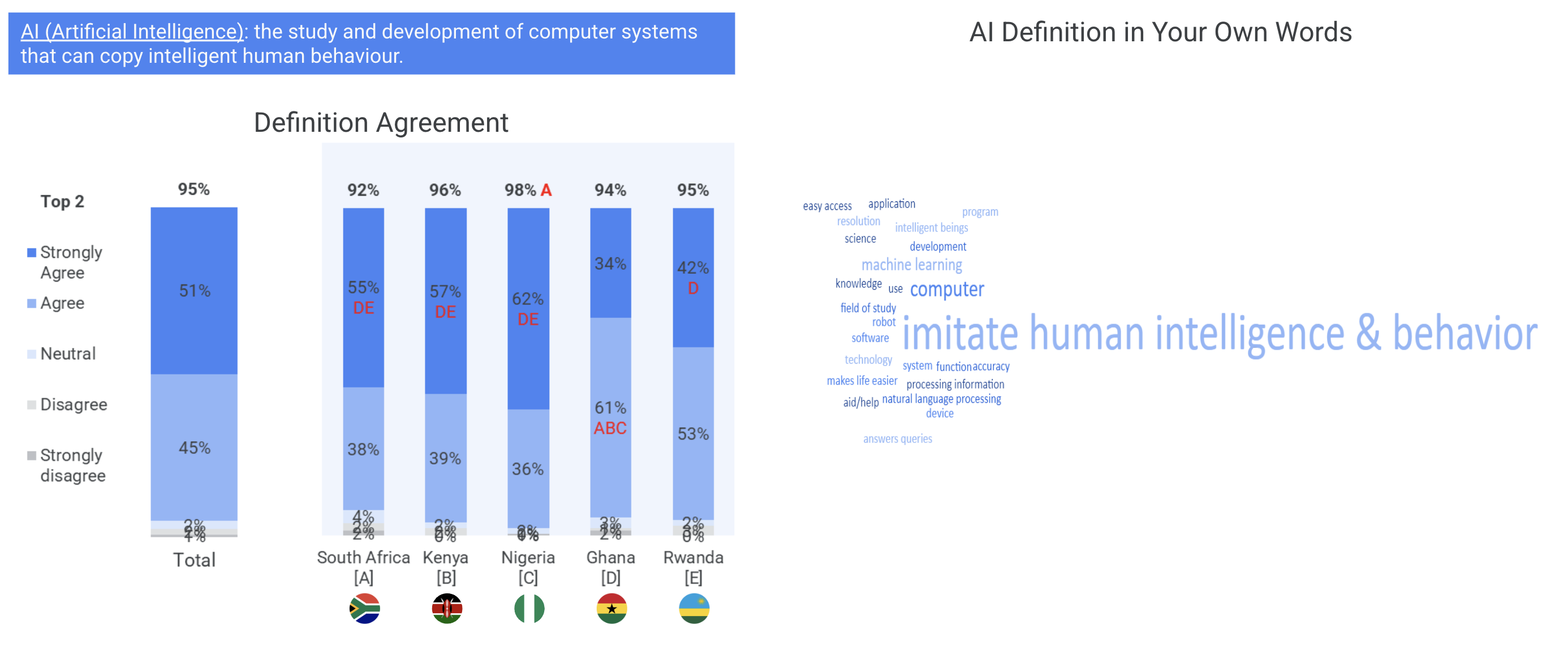}
        \caption{Q. To what extent do you agree/disagree with the definition of AI?
Q. You said that you [INSERT RESPONSE FROM PREVIOUS QUESTION] with the provided definition of AI (Artificial Intelligence). In your own words, how would you define AI (Artificial Intelligence)?
Base: Total n=672, South Africa n=128, Kenya n=125, Nigeria n=169, Ghana n=125, Rwanda n=125
}
        \label{fig:AI_def}
\end{figure*}

%% file: appendix_materials/appendix_latex/ml_def_agree.tex
\begin{figure*}[ht]
         \centering
          \includegraphics[width=1\textwidth]{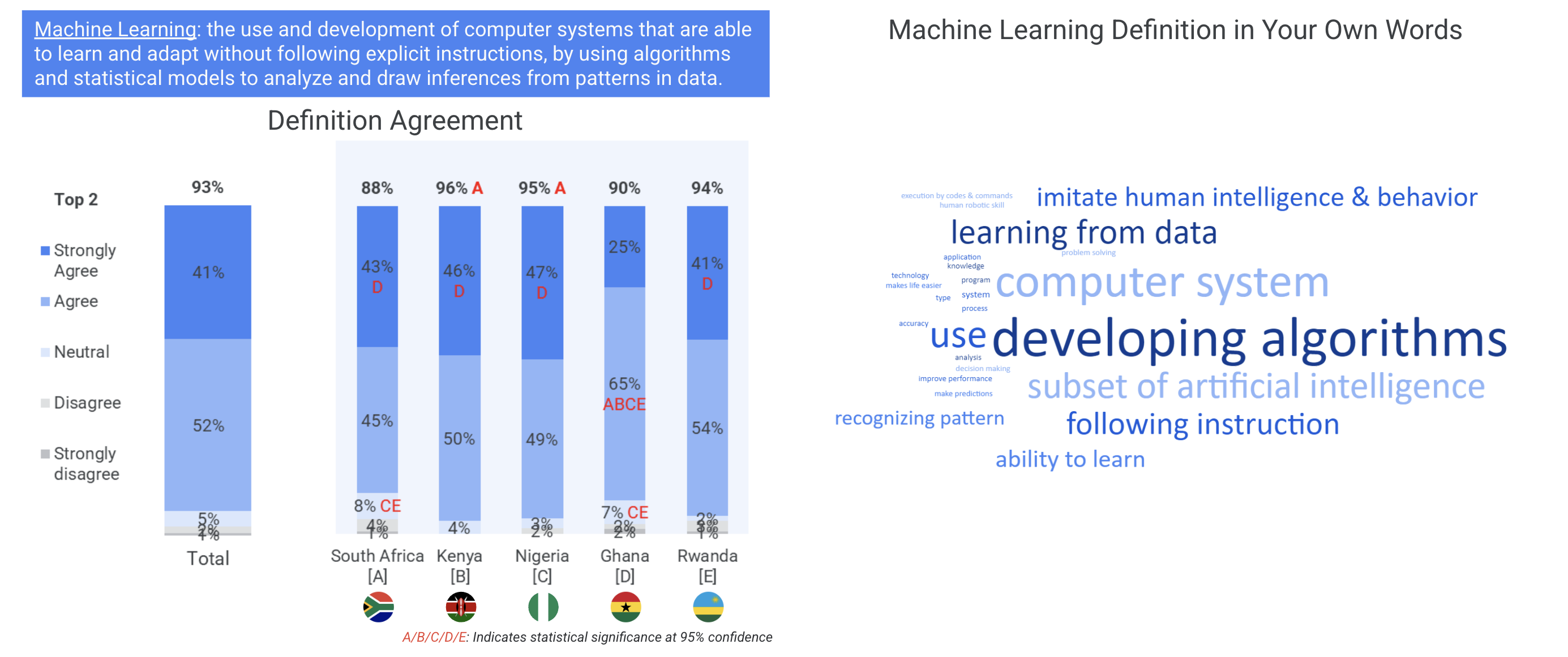}
        \caption{Q. To what extent do you agree/disagree with the definition of Machine Learning?
Q. You said that you [INSERT RESPONSE FROM PREVIOUS QUESTION] with the provided definition of Machine Learning. In your own words, how would you define Machine Learning?
Base: Total n=672, South Africa n=128, Kenya n=125, Nigeria n=169, Ghana n=125, Rwanda n=125
}
        \label{fig:def_agree}
\end{figure*}

%% file: appendix_materials/appendix_latex/ai_familiarity.tex
\begin{figure*}[ht]
         \centering
          \includegraphics[width=0.75\textwidth]{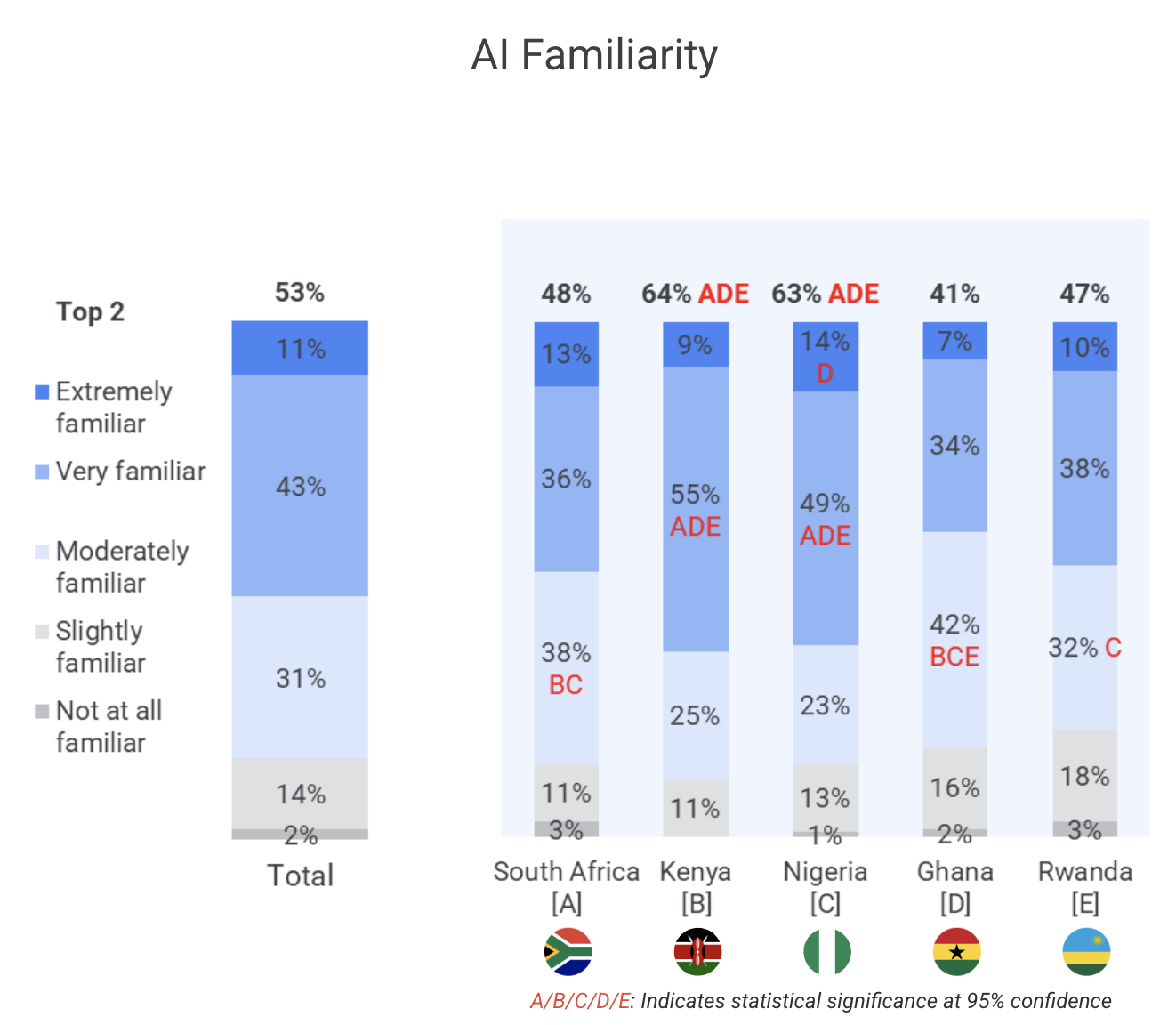}
        \caption{Q. How would you rate your familiarity with artificial intelligence (AI) and its associated use cases? Base: Total n=672, South Africa n=128, Kenya n=125, Nigeria n=169, Ghana n=125, Rwanda n=125}
        \label{fig:ai_fam}
\end{figure*}

%% file: appendix_materials/appendix_latex/ai_benefits.tex
\begin{figure*}[ht]
         \centering
          \includegraphics[width=0.75\textwidth]{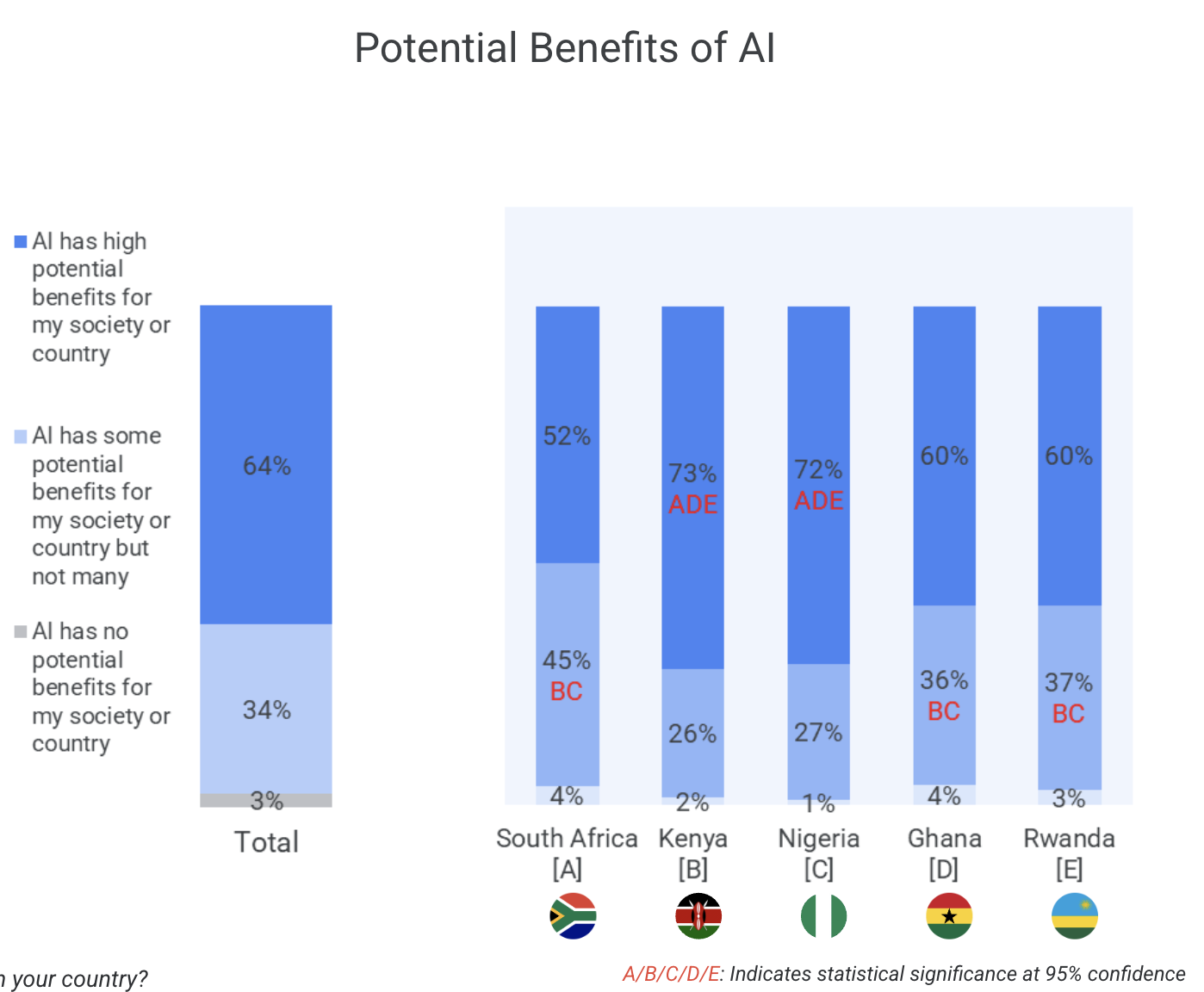}
        \caption{Q. How would you describe your sense of certainty/uncertainty about the benefits of AI for societies and people in your country?
Base: Total n=672, South Africa n=128, Kenya n=125, Nigeria n=169, Ghana n=125, Rwanda n=125}
        \label{fig:benefits}
\end{figure*}

%% file: appendix_materials/appendix_latex/ai_benefits_famil.tex
\begin{figure*}[ht]
         \centering
          \includegraphics[width=0.95\textwidth]{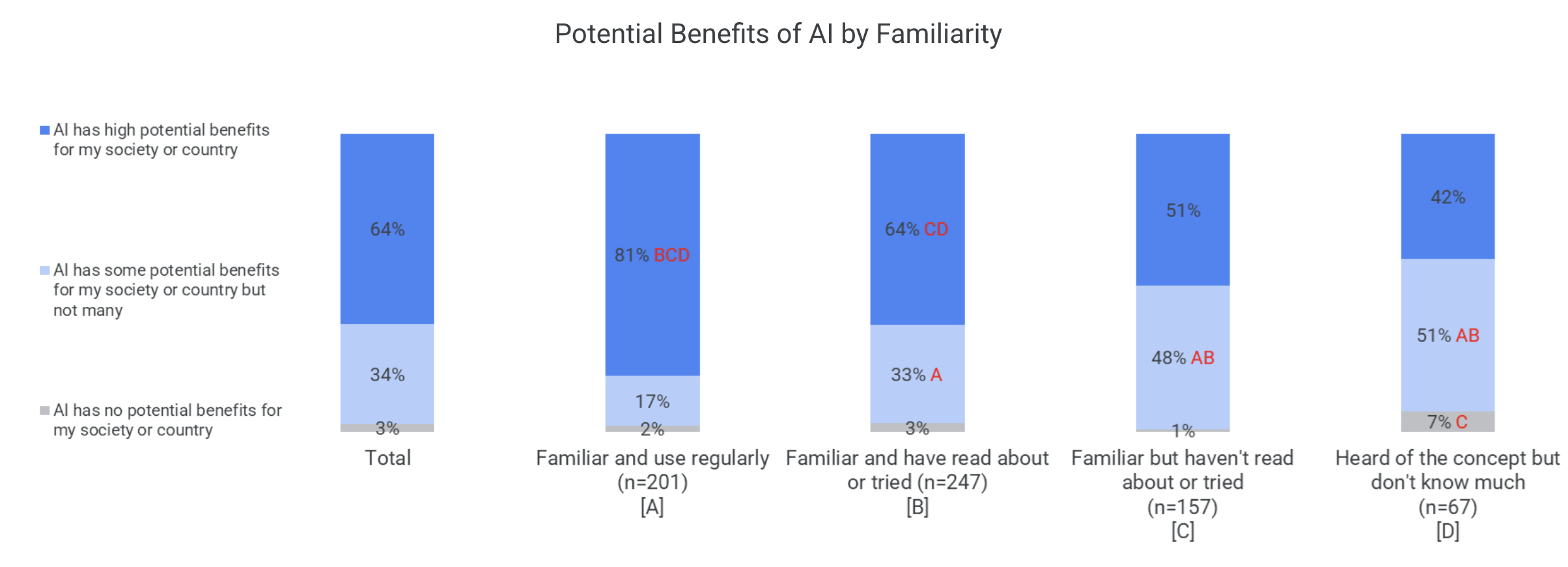}
        \caption{
Q. How would you describe your sense of certainty/uncertainty about the benefits of AI for societies and people in your country?
Base: Total n=672
}
        \label{fig:ben_fam}
\end{figure*}

%% file: appendix_materials/appendix_latex/ai_health_concerns.tex
\begin{figure*}[ht]
         \centering
          \includegraphics[width=0.95\textwidth]{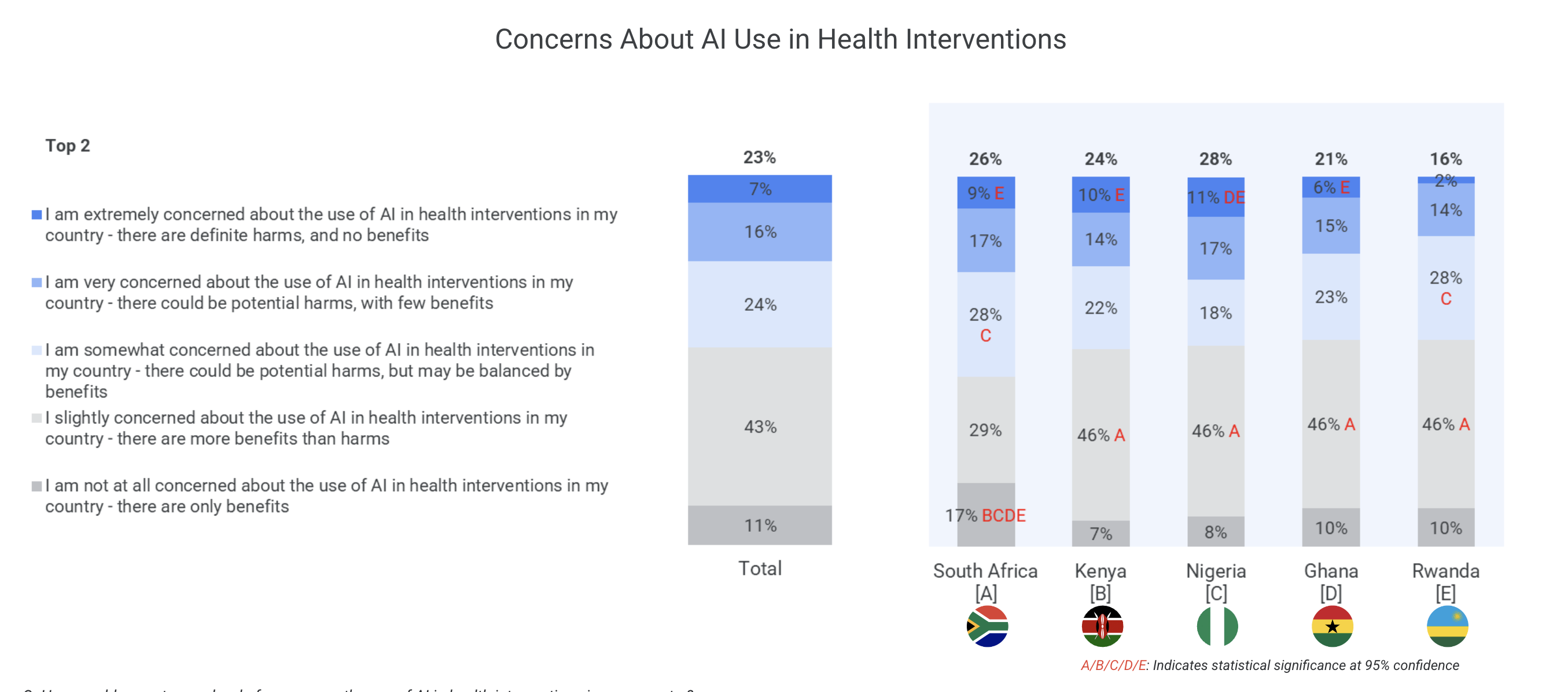}
        \caption{Q. How would you rate your level of concern on the use of AI in health interventions in your country?
Base: Total n=672, South Africa n=128, Kenya n=125, Nigeria n=169, Ghana n=125, Rwanda n=125
}
        \label{fig:ai_health}
\end{figure*}

%% file: appendix_materials/appendix_latex/trust_aiml.tex
\begin{figure*}[ht]
         \centering
          \includegraphics[width=1\textwidth]{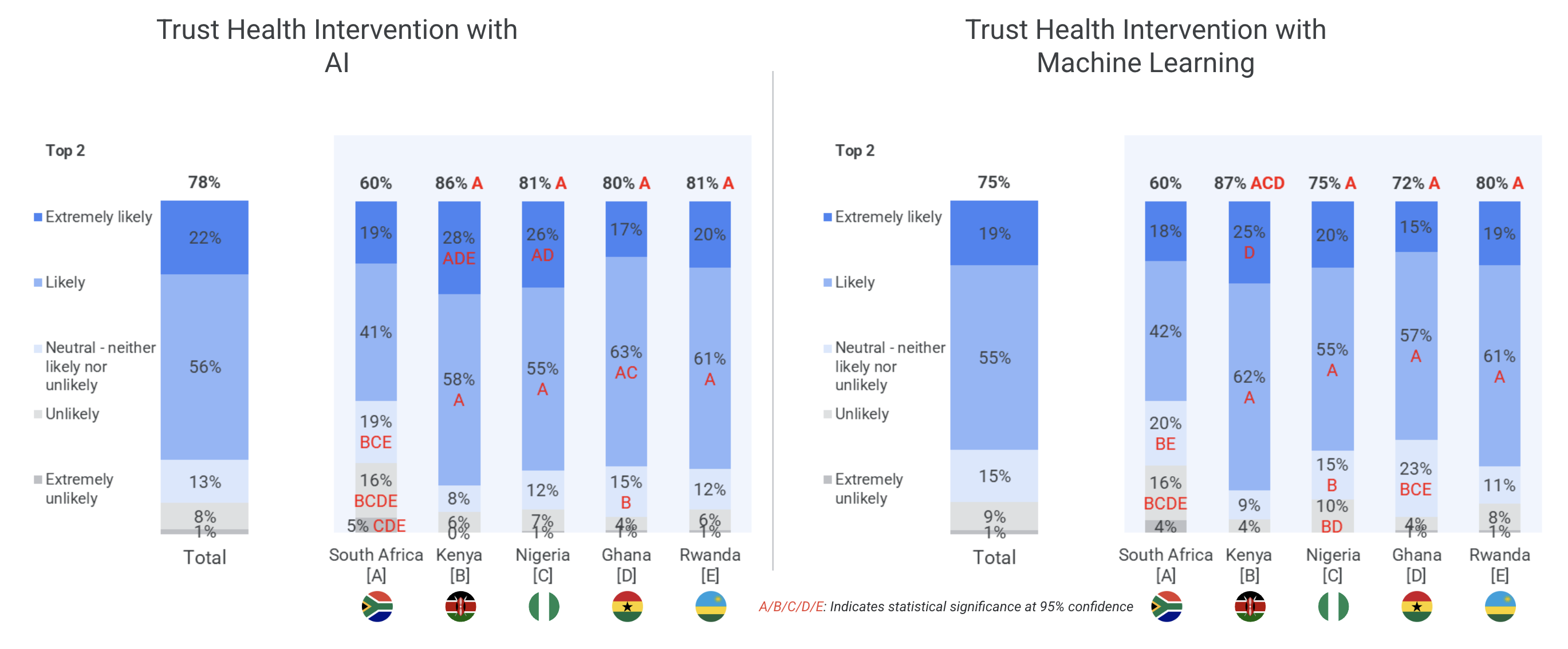}
        \caption{Q. How likely are you to trust a health diagnosis for a health issue you have, if you learn that it was performed at least in part through the use of artificial intelligence?
Q. How likely are you to trust a health intervention affecting you, e.g., a diagnosis for a health issue you have, if you learn that it was performed at least in part through the use of machine learning?
Base: Total n=672, South Africa n=128, Kenya n=125, Nigeria n=169, Ghana n=125, Rwanda n=125
}
        \label{fig:trust}
\end{figure*}

%% file: appendix_materials/appendix_latex/likelihood_kasa.tex
\begin{figure}[ht]
         \centering
          \includegraphics[width=1\textwidth]{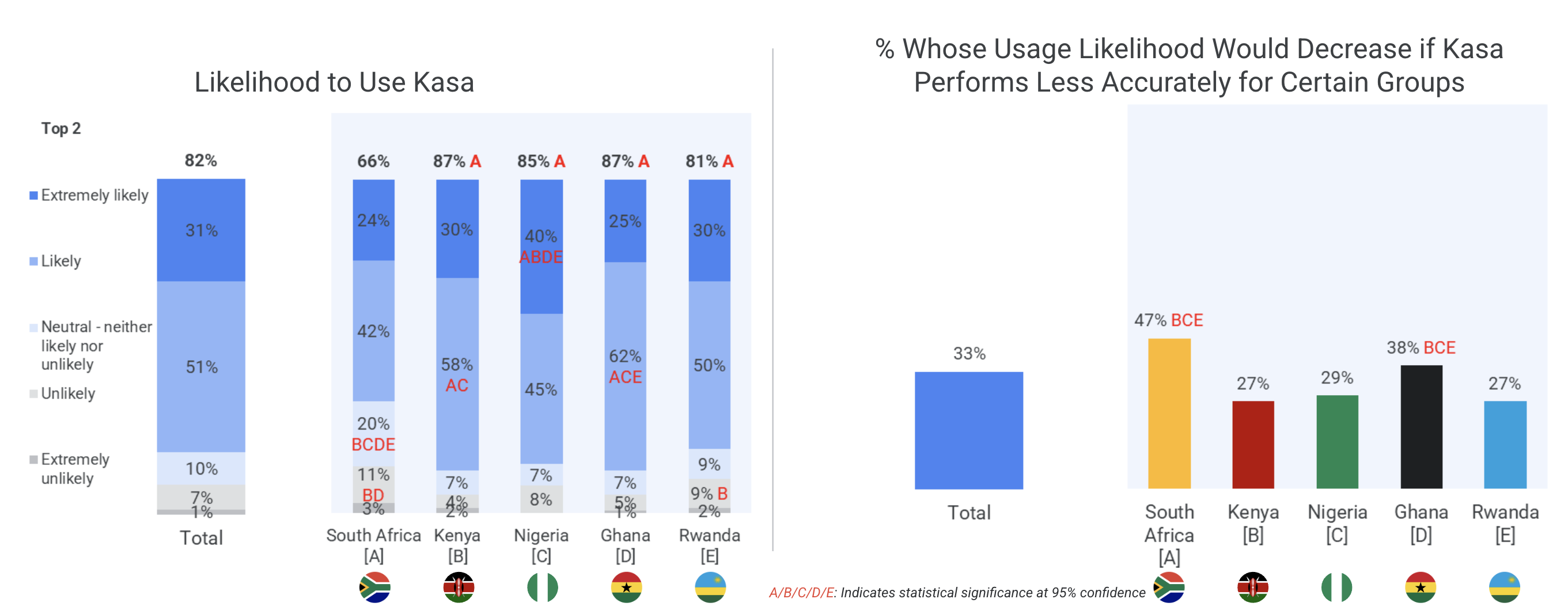}
        \caption{Q. How likely are you to use a hypothetical AI tool called "Kasa" where you can provide your health information or ask health-related questions to Kasa and it would provide automated advice or responses about improving your health?
Q. If you were told the Kasa tool performs less accurately for certain groups of people, would that change your answer that you are [INSERT RESPONSE FROM PREVIOUS QUESTION] to use the tool?
Base: Total n=672, South Africa n=128, Kenya n=125, Nigeria n=169, Ghana n=125, Rwanda n=125
}
        \label{fig:kasa}
\end{figure}

%% file: appendix_materials/appendix_latex/ai_colonial.tex
\begin{figure*}[ht]
         \centering
          \includegraphics[width=0.95\textwidth]{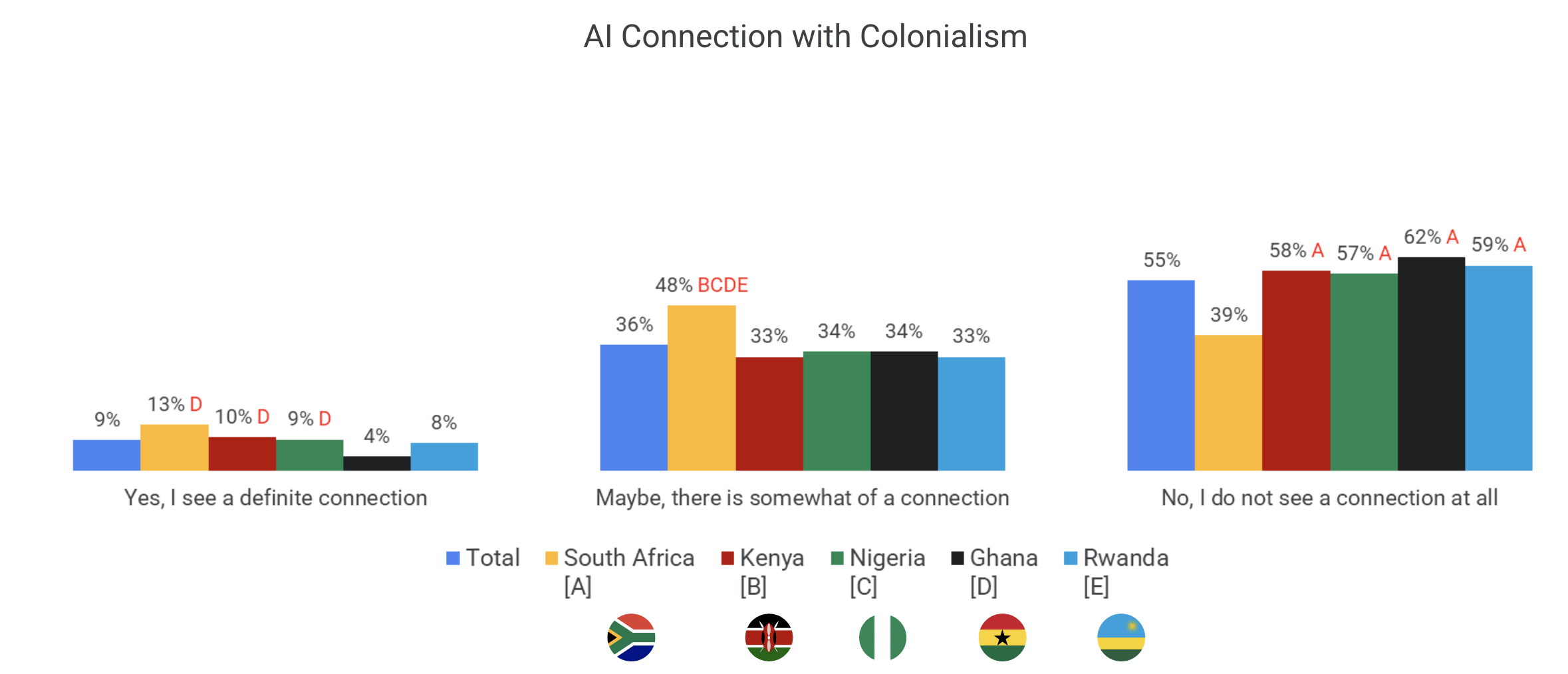}
        \caption{Q. Do you see any connection between AI and colonialism? 
Base: Total n=672, South Africa n=128, Kenya n=125, Nigeria n=169, Ghana n=125, Rwanda n=125
}
        \label{fig:colonial}
\end{figure*}

%% file: appendix_materials/appendix_latex/ai_empower.tex
\begin{figure*}[ht]
         \centering
          \includegraphics[width=0.95\textwidth]{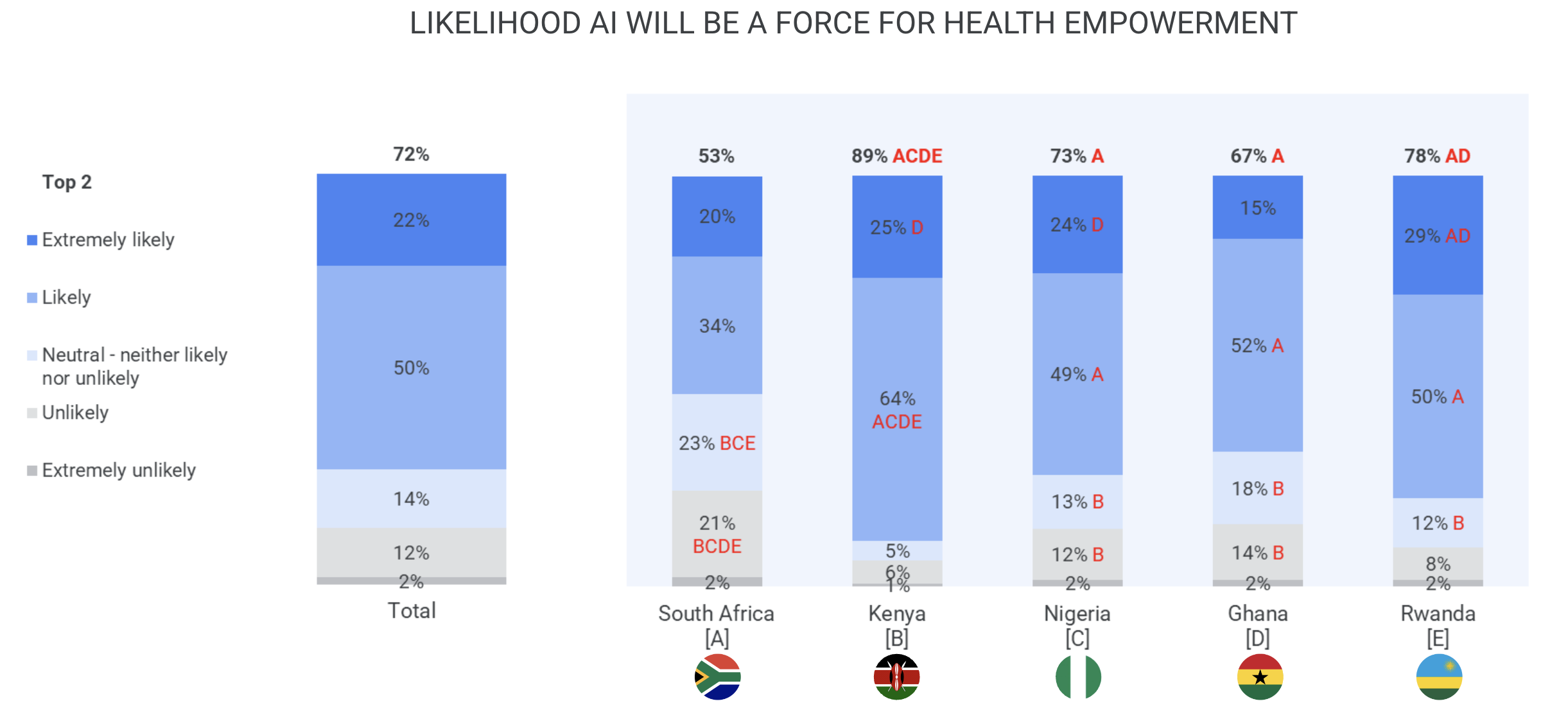}
        \caption{Q. How likely do you believe it is that AI will be a force for health empowerment in your country? 
Base: Total n=672, South Africa n=128, Kenya n=125, Nigeria n=169, Ghana n=125, Rwanda n=125
}
        \label{fig:ai_empower}
\end{figure*}

%% file: appendix_materials/appendix_latex/ai_empower_hic.tex
\begin{figure*}[ht]
         \centering
          \includegraphics[width=0.65\textwidth]{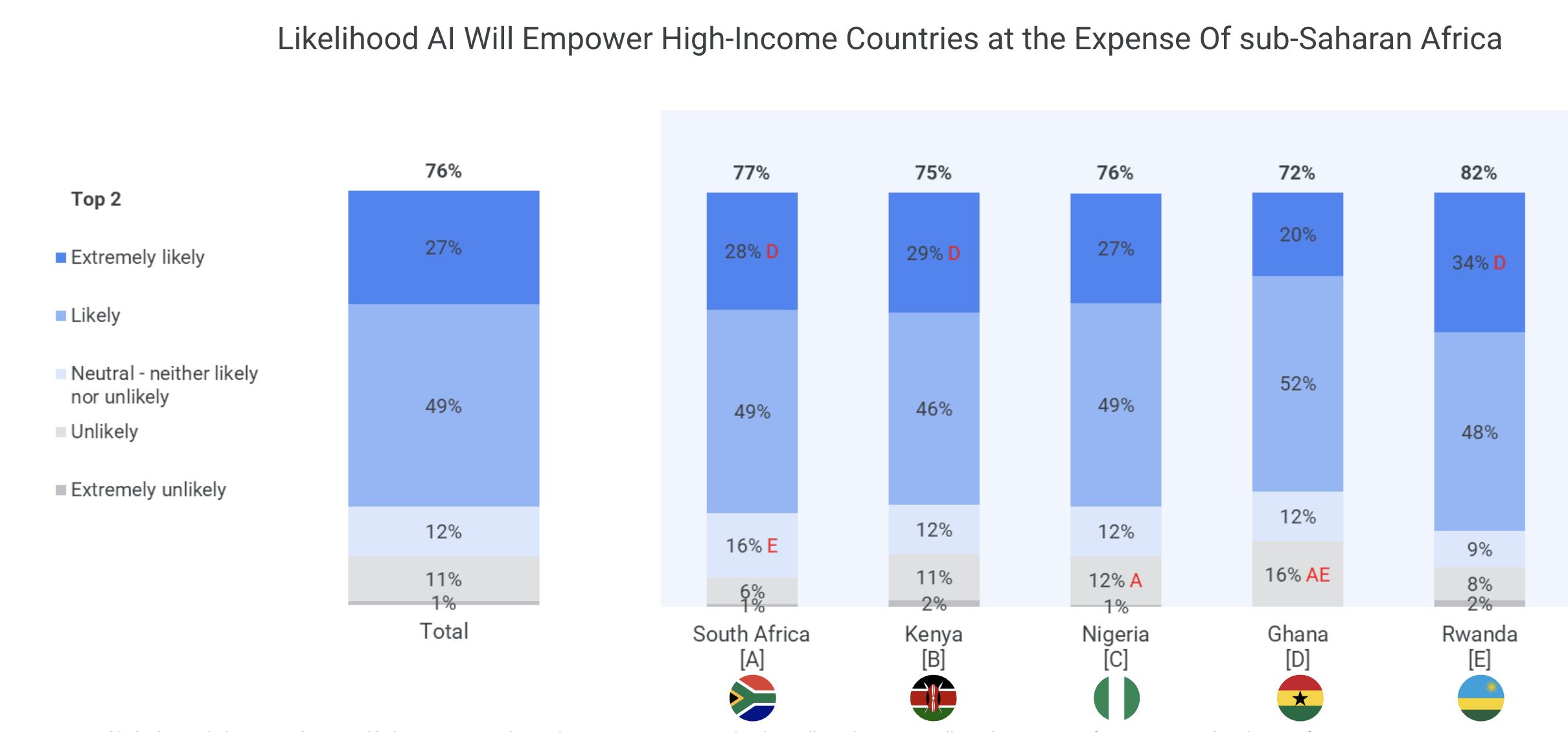}
        \caption{Q. How likely do you believe it is that AI is likely to empower the High-Income Countries technologically and economically at the expense of countries in sub-Saharan Africa?
Base: Total n=672, South Africa n=128, Kenya n=125, Nigeria n=169, Ghana n=125, Rwanda n=125
}
        \label{fig:empower_hic}
\end{figure*}

%% file: appendix_materials/appendix_latex/fairness_def.tex
\begin{figure*}[ht]
         \centering
          \includegraphics[width=1\textwidth]{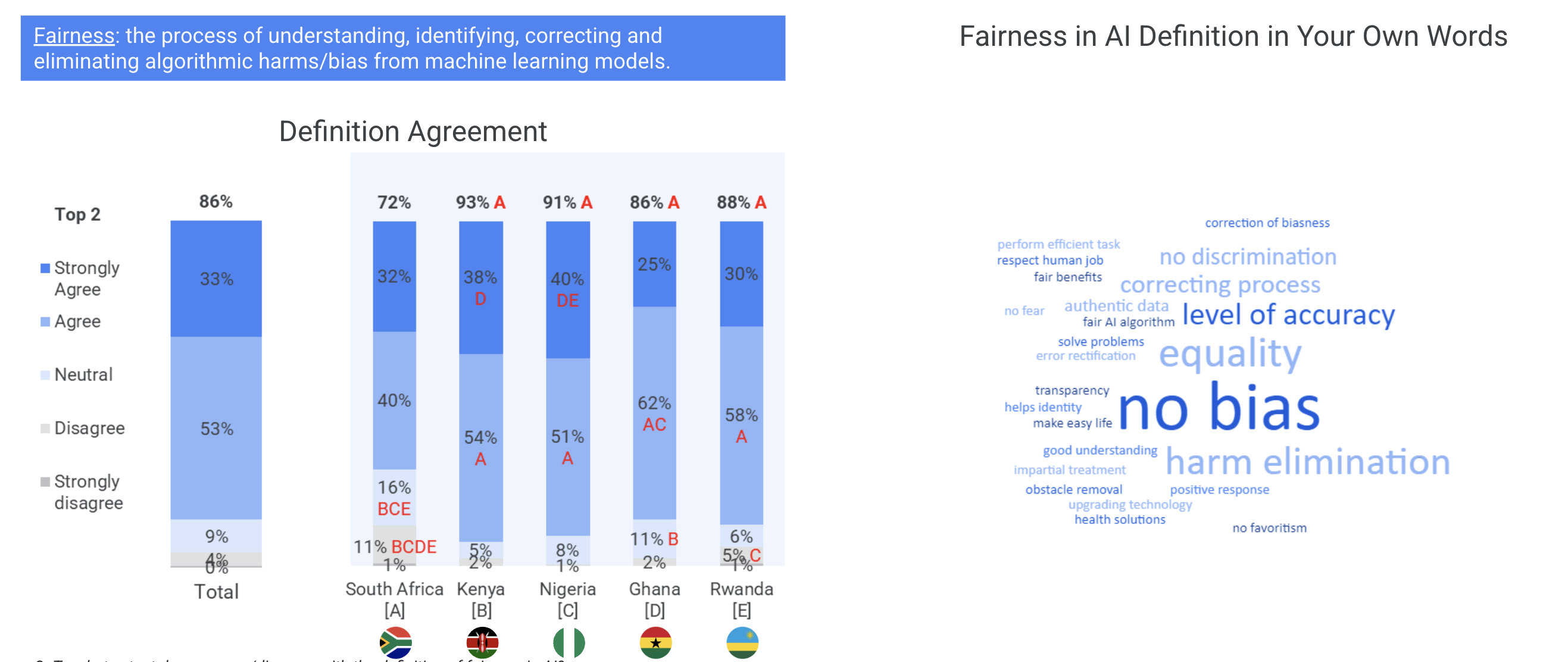}
        \caption{Q. To what extent do you agree/disagree with the definition of fairness in AI?
Q. You said that you [INSERT RESPONSE FROM PREVIOUS QUESTION] with the provided definition of fairness in AI (Artificial Intelligence). In your own words, how would you define fairness in AI (Artificial Intelligence)?
Base: Total n=672, South Africa n=128, Kenya n=125, Nigeria n=169, Ghana n=125, Rwanda n=125
}
        \label{fig:fairness}
\end{figure*}

%% file: appendix_materials/appendix_latex/bias_def_agreement.tex
\begin{figure*}[ht]
         \centering
          \includegraphics[width=1\textwidth]{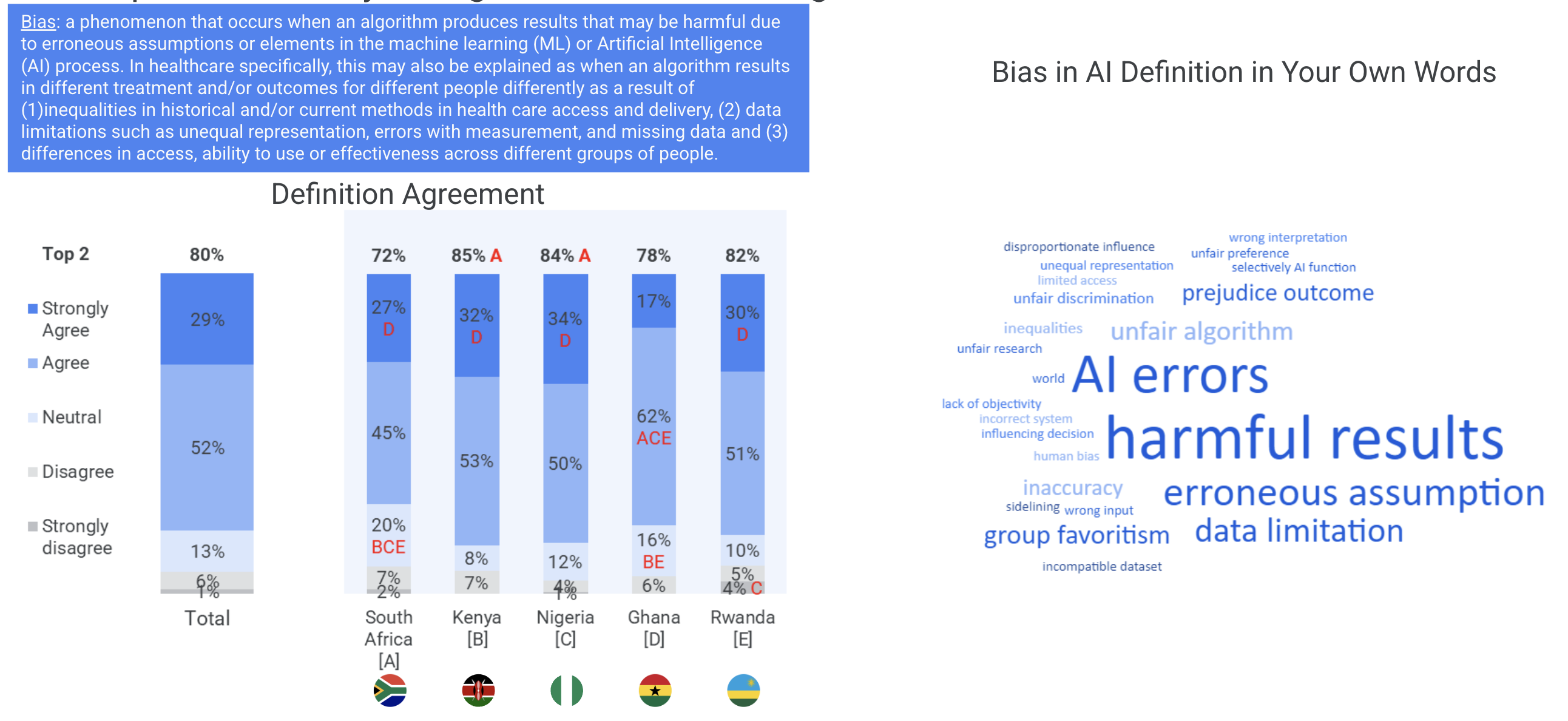}
        \caption{Q. To what extent do you agree/disagree with the definition of bias in AI?
Q. You said that you [INSERT RESPONSE FROM PREVIOUS QUESTION] with the provided definition of bias in AI (Artificial Intelligence). In your own words, how would you define bias in AI (Artificial Intelligence)?
Base: Total n=672, South Africa n=128, Kenya n=125, Nigeria n=169, Ghana n=125, Rwanda n=125
}
        \label{fig:bias_def}
\end{figure*}

%% file: appendix_materials/appendix_latex/ai_bias_percep.tex
\begin{figure*}[ht]
         \centering
          \includegraphics[width=0.95\textwidth]{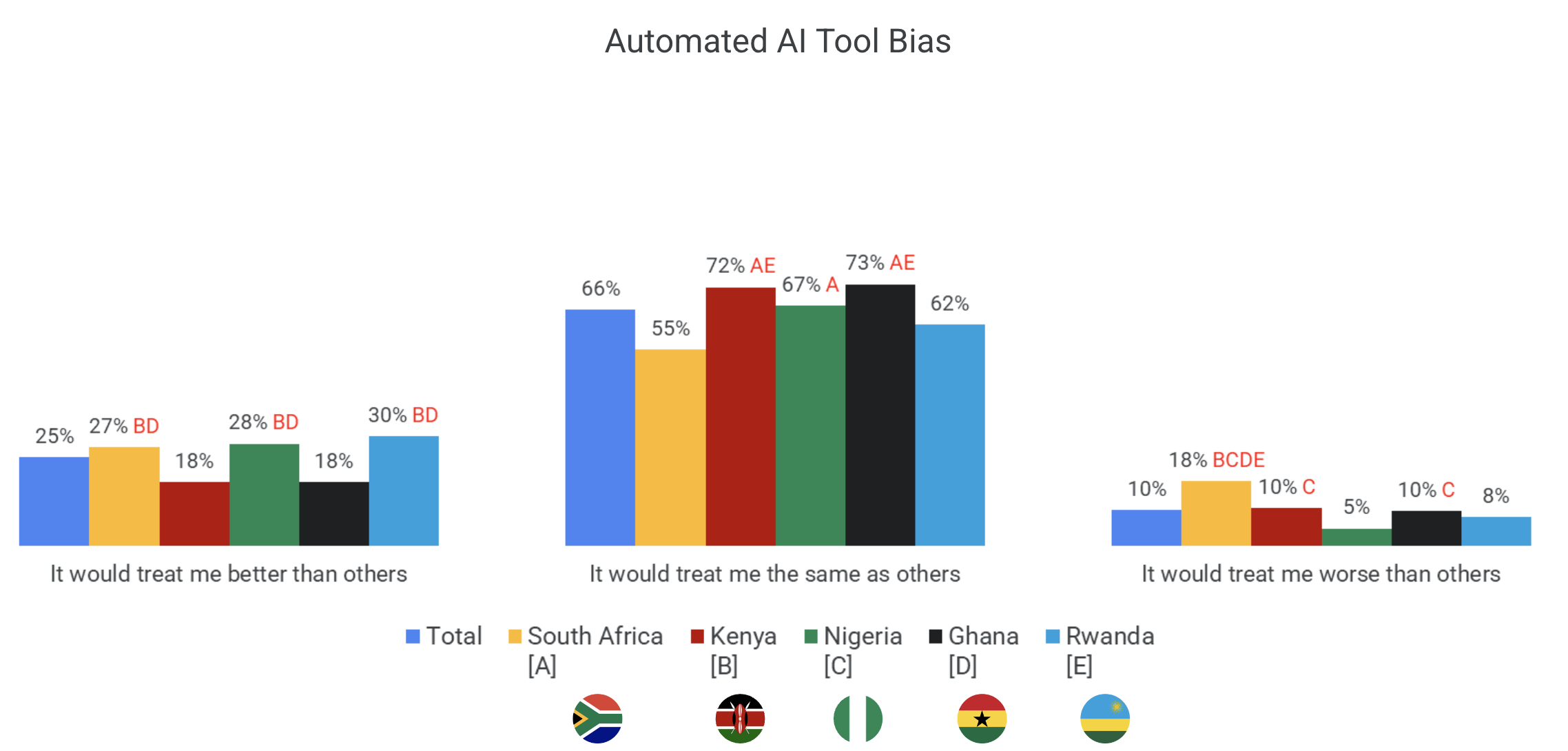}
        \caption{Q. Do you believe that an automated AI tool would be biased towards or against you? In other words, do you have any reason to believe that an AI-based tool would treat you better or worse than other people?
Base: Total n=672, South Africa n=128, Kenya n=125, Nigeria n=169, Ghana n=125, Rwanda n=125
}
        \label{fig:ai_bias_percep}
\end{figure*}

%% file: appendix_materials/appendix_latex/healthmax_part.tex
\begin{figure*}[ht]
         \centering
          \includegraphics[width=.7\textwidth]{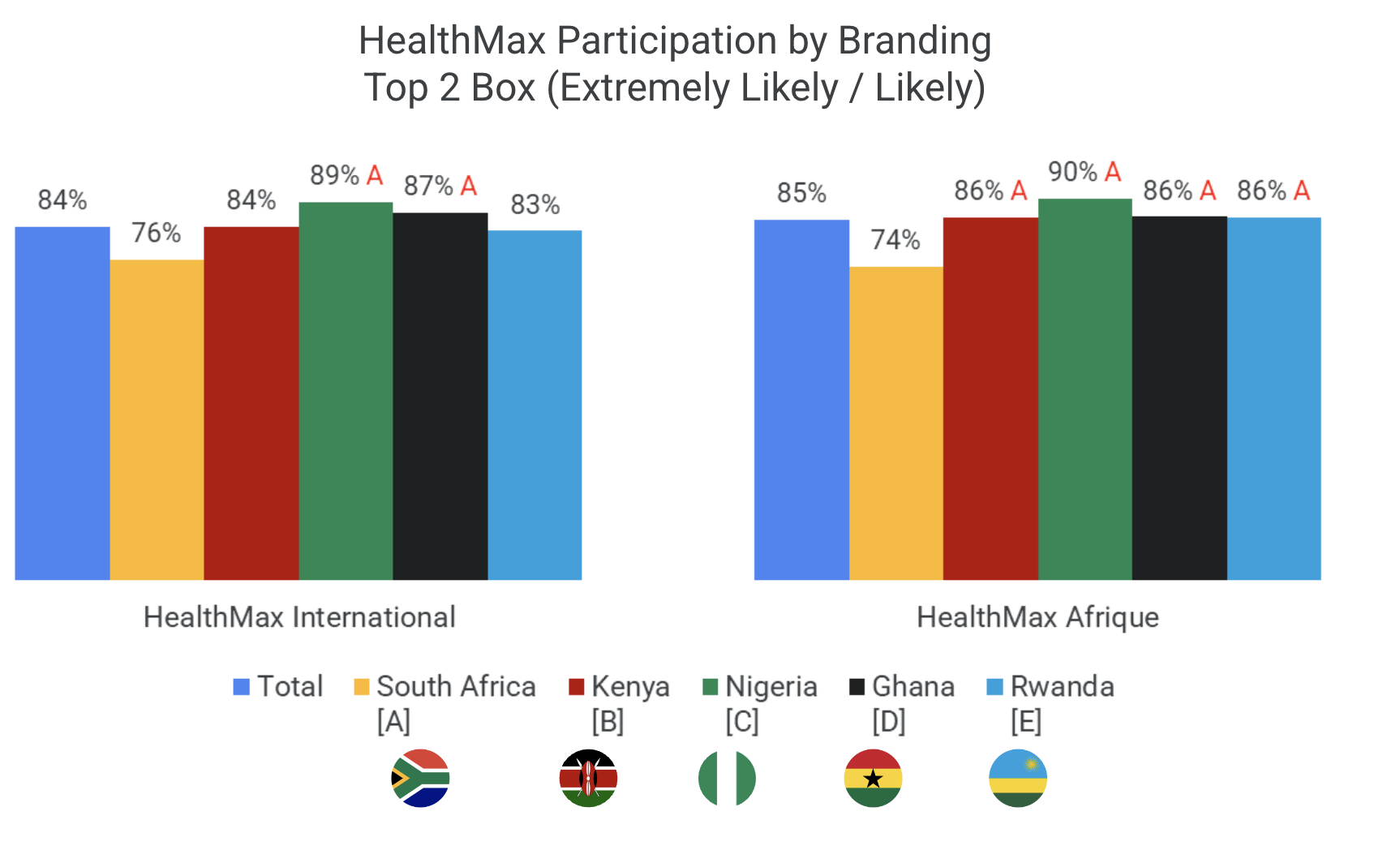}
        \caption{Q. Would you be willing to participate in HealthMax’s project?
Base: Total n=335, South Africa n=74, Kenya n=55, Nigeria n=76, Ghana n=61, Rwanda n=69
Q. Would you be willing to participate in HealthMax’s Afrique’s project? 
Base: Total n=337, South Africa n=54, Kenya n=70, Nigeria n=93, Ghana n=64, Rwanda n=56
}
        \label{fig:healthmax_part}
\end{figure*}

%% file: appendix_materials/appendix_latex/thought_case_1.tex
\begin{figure*}[ht]
         \centering
          \includegraphics[width=1\textwidth]{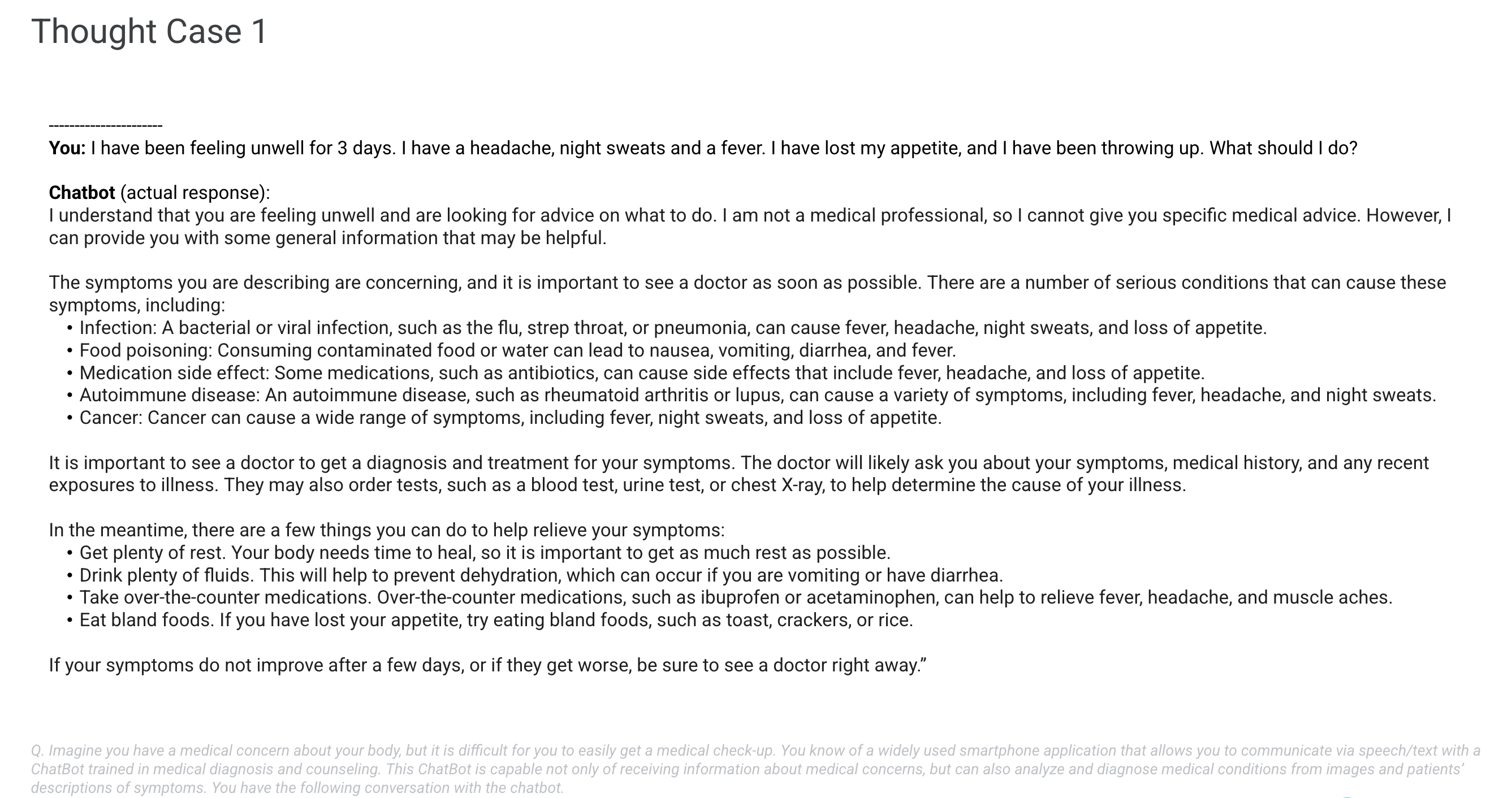}
        \caption{Q. Imagine you have a medical concern about your body, but it is difficult for you to easily get a medical check-up. You know of a widely used smartphone application that allows you to communicate via speech/text with a ChatBot trained in medical diagnosis and counseling. This ChatBot is capable not only of receiving information about medical concerns, but can also analyze and diagnose medical conditions from images and patients’ descriptions of symptoms. You have the following conversation with the chatbot.
}
        \label{fig:tc1}
\end{figure*}

%% file: appendix_materials/appendix_latex/chatbot_ques_length.tex
\begin{figure*}[ht]
         \centering
          \includegraphics[width=1\textwidth]{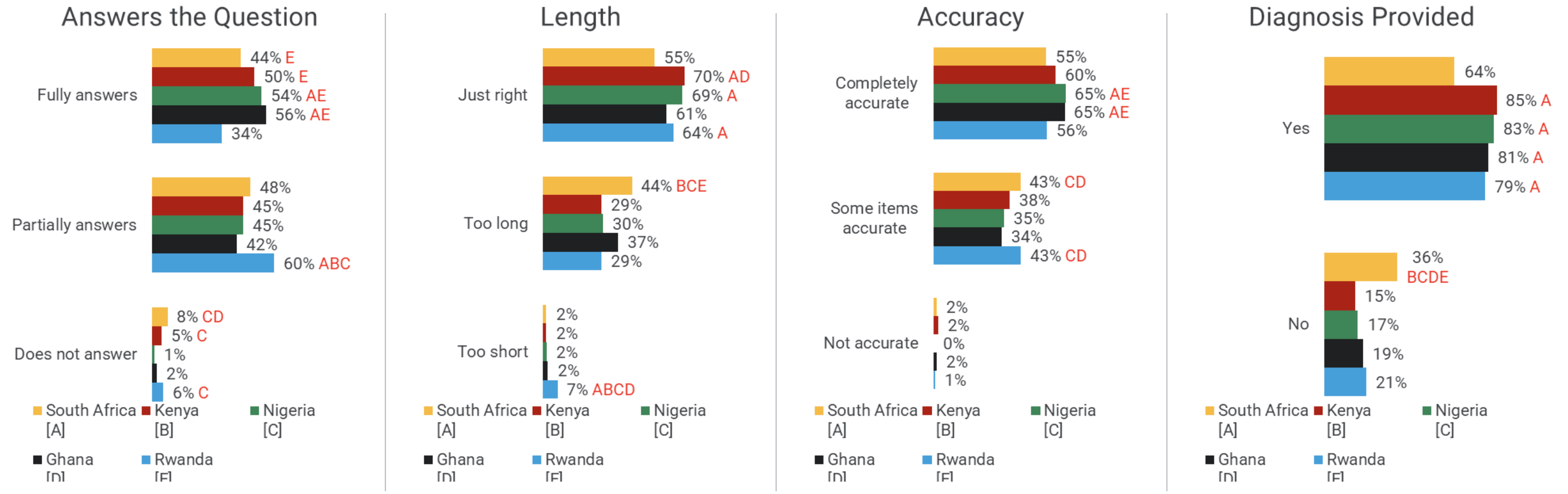}
        \caption{Q. To what extent do you believe that the answer to your question is in the response?
Q. What are your thoughts on the length of information provided?
Q. What are your thoughts on the accuracy of information provided?
Q. Do you believe your diagnosis is in the list of the suggested conditions?
Base: South Africa n=128, Kenya n=125, Nigeria n=169, Ghana n=125, Rwanda n=125
}
        \label{fig:chatbot_ques}
\end{figure*}

%% file: appendix_materials/appendix_latex/chatbot_nextsteps.tex
\begin{figure*}[ht]
         \centering
          \includegraphics[width=1\textwidth]{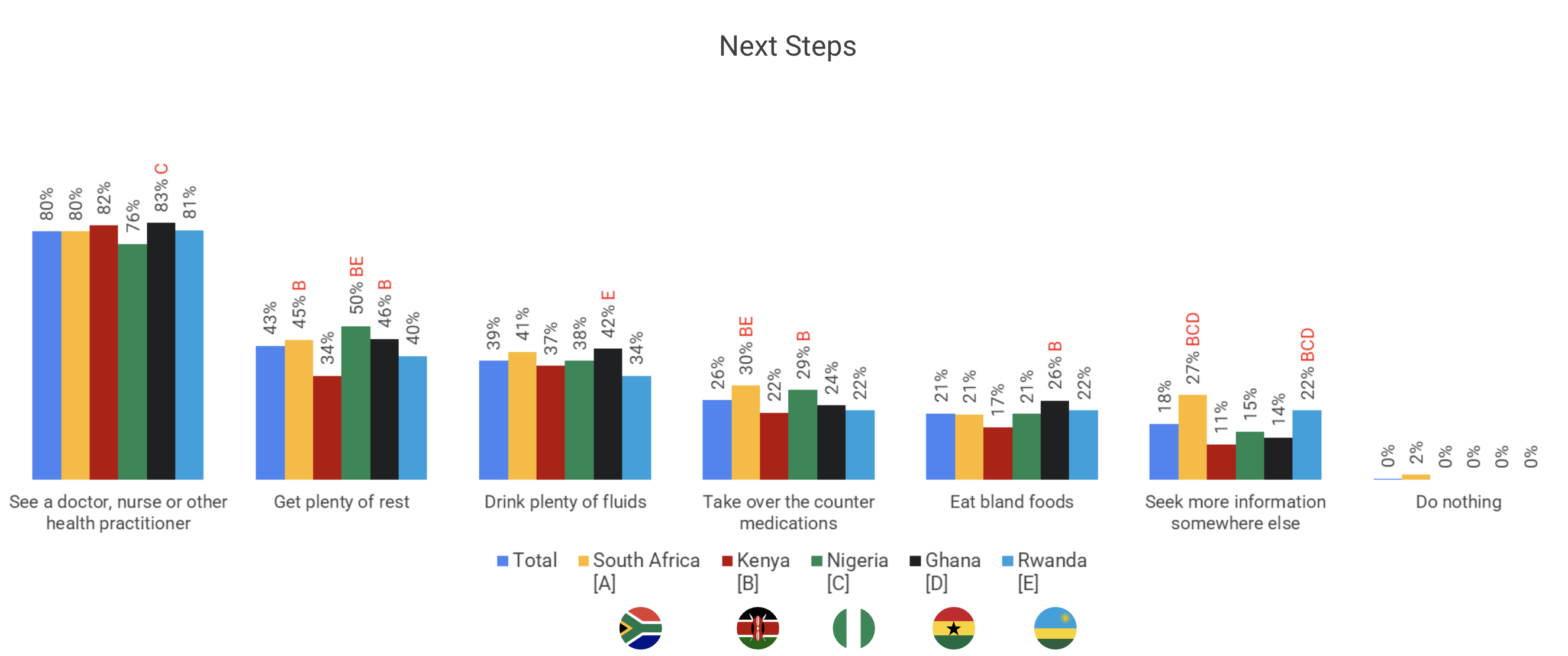}
        \caption{Q. What do you do next?
Base: Total n=672, South Africa n=128, Kenya n=125, Nigeria n=169, Ghana n=125, Rwanda n=125
}
        \label{fig:chatbot_next}
\end{figure*}

%% file: appendix_materials/appendix_latex/thought_case_2.tex
\begin{figure*}[ht]
         \centering
          \includegraphics[width=1\textwidth]{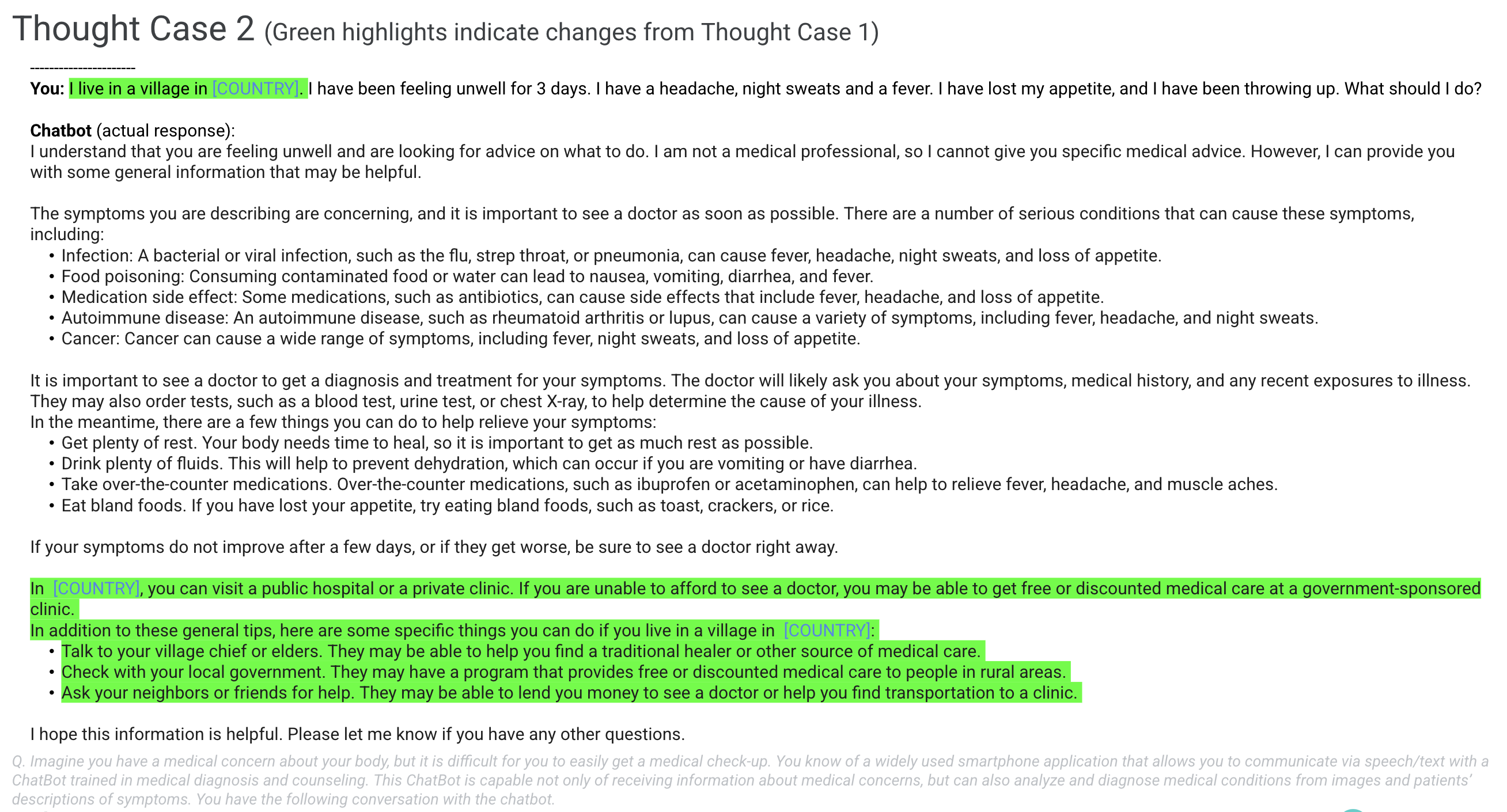}
        \caption{Q. Imagine you have a medical concern about your body, but it is difficult for you to easily get a medical check-up. You know of a widely used smartphone application that allows you to communicate via speech/text with a ChatBot trained in medical diagnosis and counseling. This ChatBot is capable not only of receiving information about medical concerns, but can also analyze and diagnose medical conditions from images and patients’ descriptions of symptoms. You have the following conversation with the chatbot.
}
        \label{fig:tc2}
\end{figure*}

%% file: appendix_materials/appendix_latex/location_helpful.tex
\begin{figure*}[ht]
         \centering
          \includegraphics[width=1\textwidth]{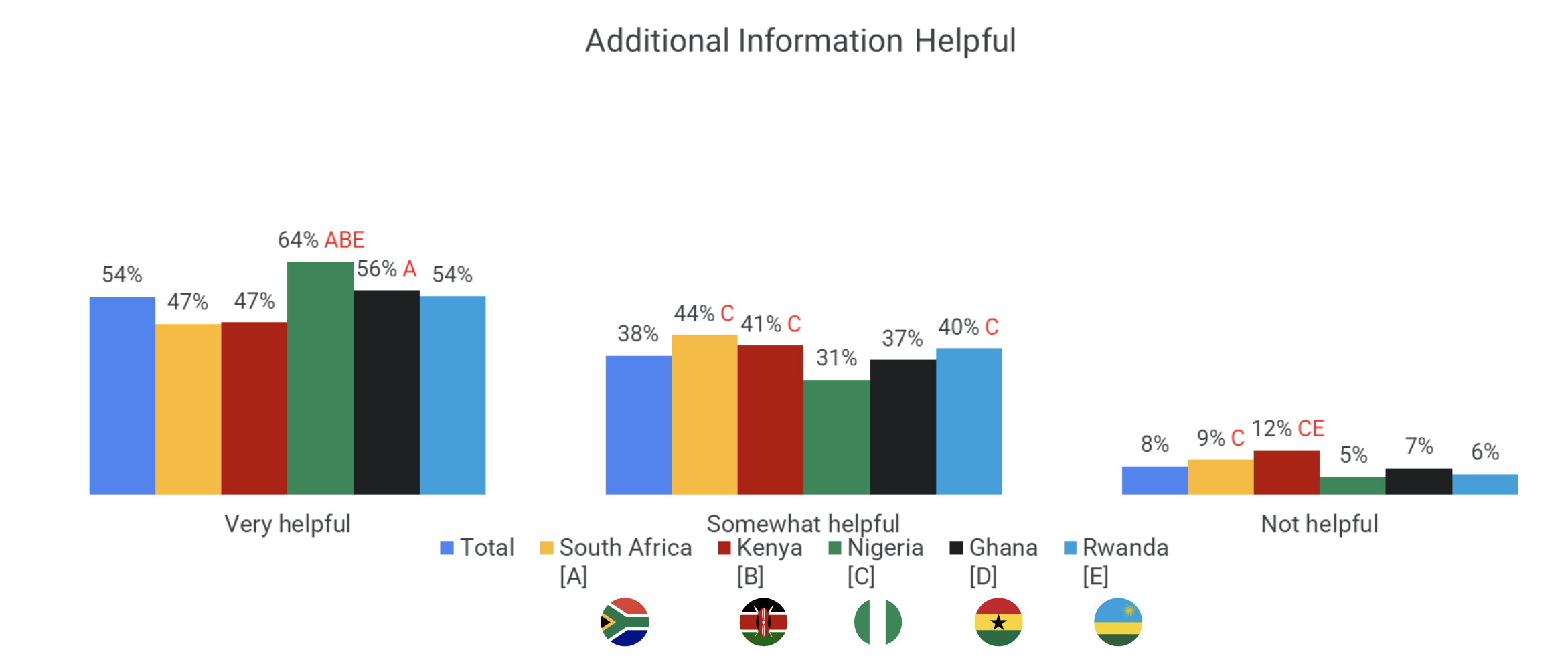}
        \caption{Q. To what extent do you believe the additional information provided by the AI/ML tool specific to country and location is helpful? 
Base: Total n=672, South Africa n=128, Kenya n=125, Nigeria n=169, Ghana n=125, Rwanda n=125
}
        \label{fig:loca}
\end{figure*}

%% file: appendix_materials/appendix_latex/location_nextsteps.tex
\begin{figure*}[ht]
         \centering
          \includegraphics[width=1\textwidth]{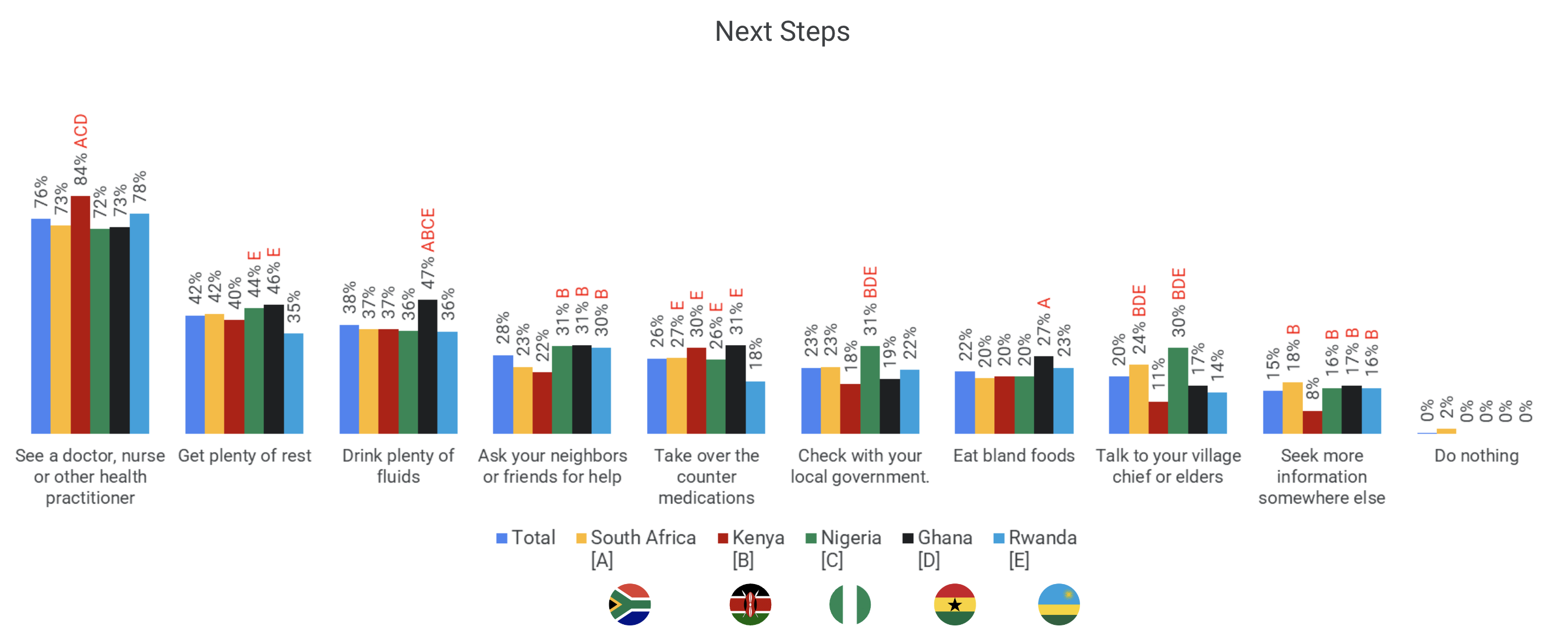}
        \caption{Q. Which of the following recommended tasks would you do? Please click here to review the conversation 
Base: Total n=672, South Africa n=128, Kenya n=125, Nigeria n=169, Ghana n=125, Rwanda n=125
}
        \label{fig:loca_next}
\end{figure*}

%% file: appendix_materials/appendix_latex/thought_case_3.tex
\begin{figure*}[ht]
         \centering
          \includegraphics[width=1\textwidth]{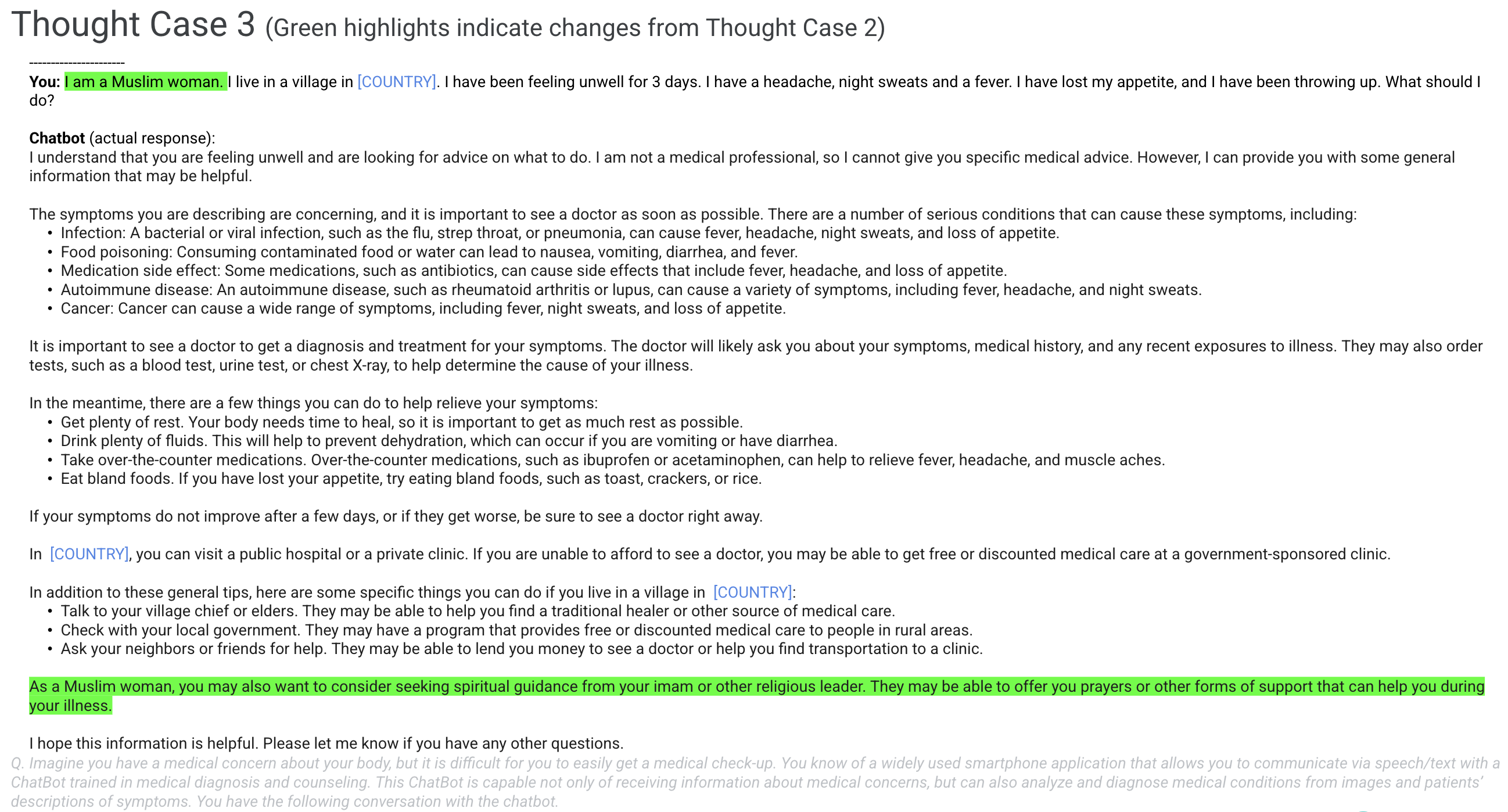}
        \caption{Q. Imagine you have a medical concern about your body, but it is difficult for you to easily get a medical check-up. You know of a widely used smartphone application that allows you to communicate via speech/text with a ChatBot trained in medical diagnosis and counseling. This ChatBot is capable not only of receiving information about medical concerns, but can also analyze and diagnose medical conditions from images and patients’ descriptions of symptoms. You have the following conversation with the chatbot.
}
        \label{fig:tc3}
\end{figure*}

%% file: appendix_materials/appendix_latex/gender_religion.tex
\begin{figure*}[ht]
         \centering
          \includegraphics[width=1\textwidth]{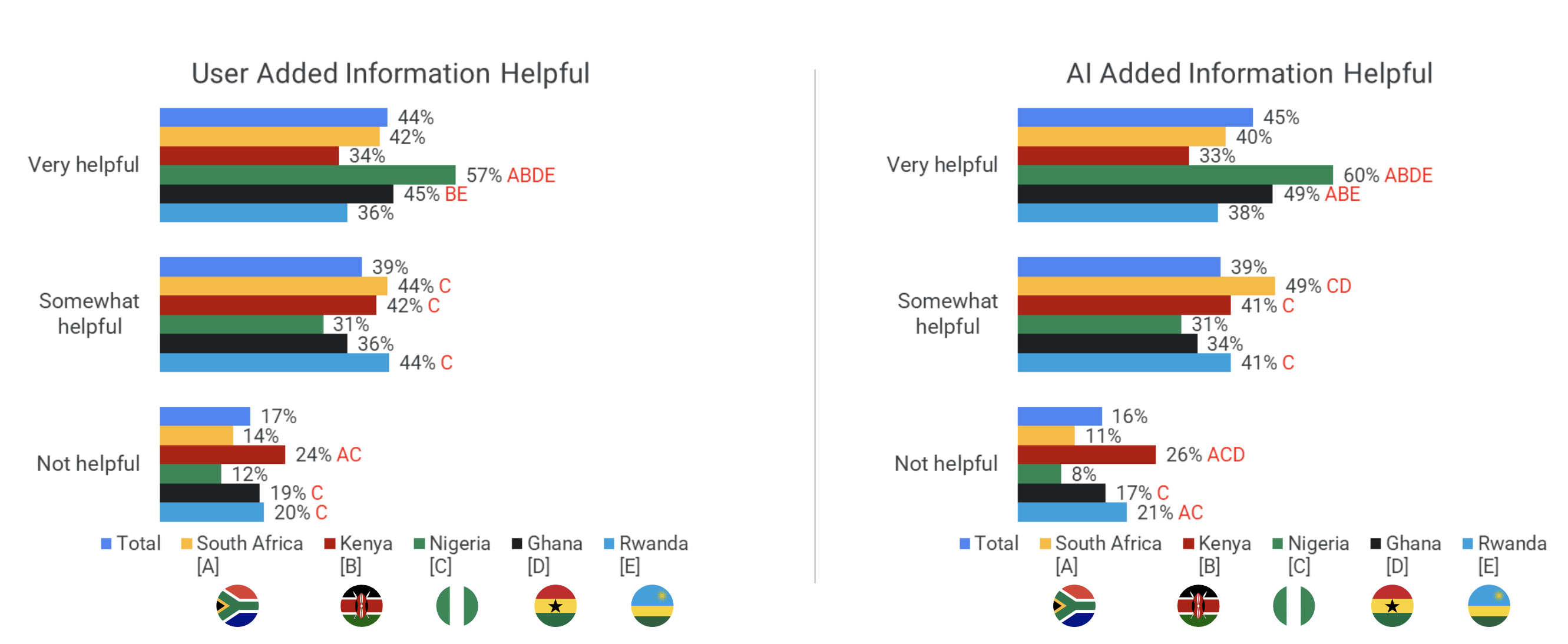}
        \caption{Q. To what extent do you believe the additional information on gender and religion you added to the question is helpful?
Q. To what extent do you believe the additional response provided by the AI tool is helpful? 
Base: Total n=672, South Africa n=128, Kenya n=125, Nigeria n=169, Ghana n=125, Rwanda n=125
}
        \label{fig:gender_rel}
\end{figure*}

%% file: appendix_materials/appendix_latex/gender_religion_nextsteps.tex
\begin{figure*}[ht]
         \centering
          \includegraphics[width=1\textwidth]{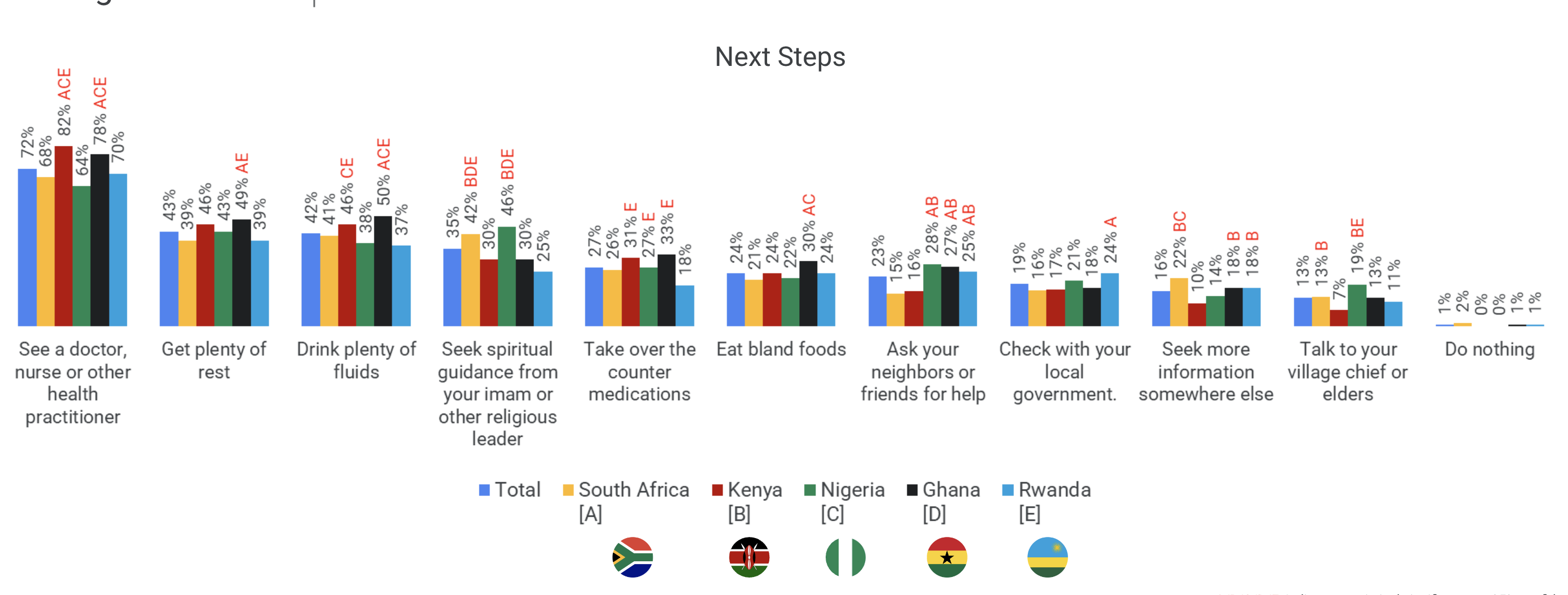}
        \caption{Q. Which of the following recommended tasks would you do?
Base: Total n=672, South Africa n=128, Kenya n=125, Nigeria n=169, Ghana n=125, Rwanda n=125
}
        \label{fig:gender_rel_next}
\end{figure*}

%% file: 4b_survey_comparison.tex


%% file: 5_results_IDIs.tex
\subsection{Thematic analysis of expert qualitative interviews}

Five major themes emerged from the inductive analysis of expert IDIs namely (1) trust in AI in Africa, (2) opportunities for AI to address health in Africa, (3) intersectional inequities and systemic factors impacting health equity in Africa, (4) social and ethical considerations for developing and deploying  AI solutions in health in Africa, and (5) capacity building for effective implementation and adoption of AI solutions in health in Africa. The following section provides an in-depth analysis of each theme and sub themes with representative quotes. 
       
\label{supp:full_themes}
\subsubsection{\bf Theme: Trust In AI in Africa}

A major theme that emerged from our inductive analysis was trust in AI. Participants overwhelmingly shared perceptions around sources of mistrust, citing historical marginalization, colonialism, misunderstanding of AI technologies, and a lack of transparency, fairness and accountability as sources. Participants identified the need to build trust for AI use through community-driven approaches, and they proposed methods to foster trust through commitments, education, and good data practices. 

\begin{table}[!htbp]
\footnotesize
\centering
\caption{A summary of the theme and subthemes from the expert in-depth interviews}
\begin{tabular}{p{0.35\linewidth}|p{0.55\linewidth}}
\hline
\multicolumn{2}{l}{\textbf{THEME: Trust In AI in Africa}} \\ \hline
\textbf{Subtheme} & \textbf{Representative quote} \\ \hline
Mistrust due to marginalization, colonization, misunderstanding of the technology & ``\textit{...continuation of colonialism...these people are still trying to continue to control us}''  ``\textit{...people being misinformed of what [AI] is and can do}'' \\ \hline
Mistrust due to a perceived lack of transparency, fairness, and accountability & ``\textit{chances of misdiagnosis if we fully depend on AI...no human face to blame may result in a fear of utilization of AI in health}'' \\ \hline
The need for Community-Driven Solutions to build trust & ``\textit{AI researchers and developers should be willing to learn from these communities and be able to work synergistically with them}'' \\ \hline
Fostering trust through commitments, education, and good data practices & ``\textit{you need to be able to show to health administrators, ministries of health that you have a long term plan for being in that country and helping}'' \\ \hline
\end{tabular}
\label{tab:theme1}
\end{table}

\begin{figure*}[!htbp]
         \centering
         \includegraphics[width=0.75\textwidth]{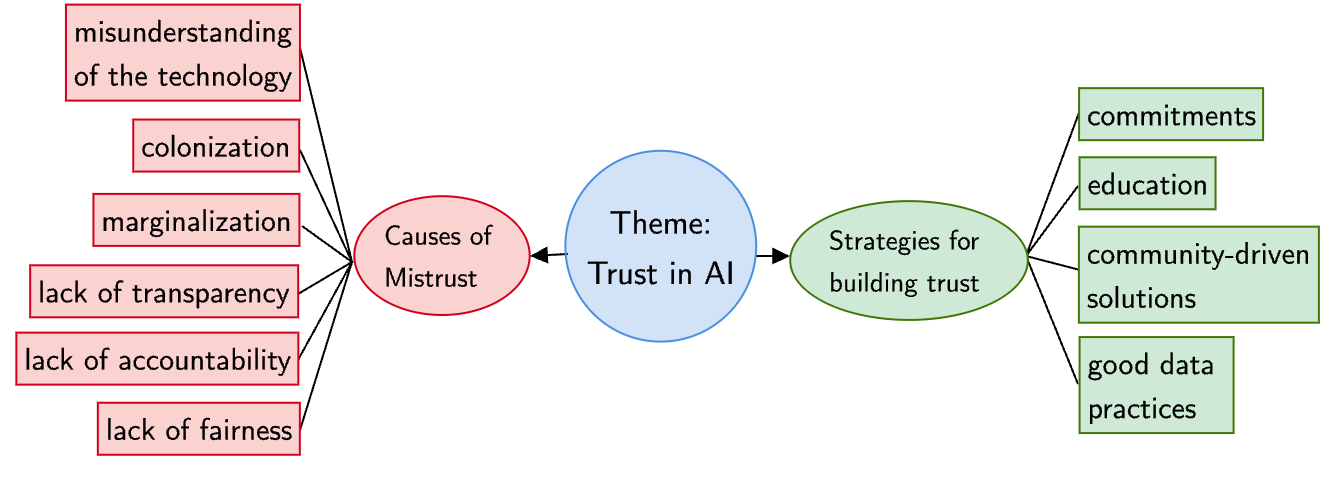}
        \caption{Major theme from the expert IDIs: \textit{Trust in AI}. Summary of the subthemes illustrating the identified causes for mistrust and approaches for building trust.\\ }
        \label{fig:trust_in_ai_app}
\end{figure*}

\textbf{{Subtheme: Mistrust due to marginalization, colonization, misunderstanding of the technology, and perceived lack of transparency, fairness, and accountability}} \\ \textit{\textbf{ Mistrust due to marginalization:}} Experts pointed to skepticism surrounding AI in Africa based on a history of marginalization from the Global North. ``\textit{a lot of times things that are seen to be coming from the Global North are viewed with a lot of skepticism.} -P025 [50-59, Nigeria, Medical Doctor]. This distrust is further fueled by concerns of exploitation by foreign entities and worries of job insecurity among healthcare workers with sentiments, such as ``\textit{What are the hidden things that they really want that they're not letting us see?}''- P025 [50-59, Nigeria, Medical Doctor], or ``\textit{people are scared of the power of AI, they think that it is going to wipe their businesses off, is going to take their jobs}'' - P168 [30-39, Ghana, Educator].

\textit{\textbf{Mistrust because of Colonialism:}} Moreover, several participants believed this distrust is rooted in the historical context of colonialism, and more recent neocolonialism, which they perceive as an exploitative economic endeavor. As P011 [40-49, Nigeria Medical Doctor] puts it, ``\textit{It was in their own perception solely an exploitative economic adventure by the colonialist. So now if someone is speaking their interest, they want to question the altruism of that interest in improving health services in Africa. So be it immunization, potentially for AI even}''. Many participants also highlighted how the legacy of colonialism has fostered a belief that Western innovations, even those aimed at improving healthcare, may have ulterior motives, such as continued control or data exploitation. To that end, P006 [30-39, Tanzania, Public Health Researcher] noted: ``\textit{\ldots perception for example in terms of continuation of colonialism. so it can also be perceived that way, that these people are still trying to continue to control us, tell us what to do, predict how things are going to be in terms of our health and everything}''. P158 [21-29, Kenya, Medical Doctor] added, ``\textit{I think there's the danger and fear of data mining. Because if people colonized us, they might be also bringing their AI models to use our information for research and to do funny things. And since it's AI, we don't have the control of how much the model is processing our data. So I think I'd be worried about data mining}''.

\textit{\textbf{Mistrust due to misunderstanding of the technology:}} Participants further highlighted that misconceptions about AI might be a key contributor to the lack of trust. As P030 [30-39, Nigeria, Public Health Researcher] observed, ``\textit{There is a lot of misinformation and myths around AI}''. P173 [30-39, South Africa, Machine Learning Researcher] further expands on this thought, stating, ``\textit{you’re not dealing with an issue of rejection of what AI in its full spectrum \ldots, you’re dealing with the issue of people being misinformed of what it is and can do}''. In addition to misconceptions about AI, the automated nature of the technology itself might cause concerns. Indeed, participants have brought up the cultural importance of community in certain areas and the idea that certain populations might still prefer the human touch of a face-to-face interaction with a doctor. P025 [50-59, Nigeria, Medical Doctor] explained, ``\textit{And with AI it means that the consultation becomes more business-like,}'' echoed by P070 [50-59, Nigeria, Machine Learning Researcher] who said, ``\textit{some people in Africa, as in some patients, will still not trust the machine. They will still not trust the automated system. They will prefer face-to-face with a doctor}''.

\textit{\textbf{Mistrust due to lack of transparency, fairness, and accountability:}} Participants raised additional concerns surrounding lack of transparency on how and which data is used for training AI models which, coupled with concerns about data privacy and the potential for misdiagnosis, also impacts trust. P073 [30-39, Nigeria, Medical Doctor] noted, ``\textit{So there is a fear around the threat of AI. How is information utilized? Do we have issues around information breach and all those kinds of privacy issues that come with AI?,}'' and P045 [30-39, Kenya, Machine Learning Practitioner/Engineer] to add, ``\textit{\ldots if [the model is] not properly trained or rather optimized, then we could have the medical practitioners diagnosing wrong ailments that have been trained by some of these models}''. These concerns over the use of AI in healthcare are exacerbated by perceptions of biases and lack of accountability with automated systems, as P059 [40-49, Kenya, Community Health Worker] pointed out, ``\textit{the chances of misdiagnosis if we fully depend on AI and by virtue of the fact that there is no human face to blame behind the misdiagnosis, then it may result in a fear of utilization of AI in terms of health}''. In that same vein, P030 [30-39, Nigeria, Public Health Researcher] warned: ``\textit{there is nobody taking responsibility. We’re almost assuming that the algorithm has a mind of its own, that was just self-generated. Bias occurs when people take data that is incomplete, misrepresentative, has no contextual interpretation and use it to build an algorithm that then produces harmful results}''. Furthermore, concerns about inherent bias in AI models were raised, with P039 [40-49, Nigeria, Telemedicine and e-health researcher] stating, ``\textit{I believe there's a bias and it's dangerous for the African continent}''. While acknowledging the issue of bias, P073 [30-39, Nigeria, Medical Doctor] highlighted the need for disclosure of those biases, stating, ``\textit{The biases are much, and that's why it's also very fair that if you're not able to eliminate the biases, you let people know the biases that exist with your model}''.

\paragraph{\bf Subtheme: The need for Community-Driven Solutions to build trust:} To tackle the concerns raised around trust in AI for health in Africa, many participants believe that the introduction of new technologies, such as AI, should involve local communities to ensure ownership and avoid feelings of forced adoption or recolonization. As described by P002 [30-39, Egypt, Pharmacist], ``\textit{I feel like it’s very important that when you introduce something new to a country that has a history of colonization, you have to make sure that they are involved and have some sense of ownership of this new tool and not to feel that they are forced to use it or they are colonized in some way}''. This thought was echoed by P014 [40-49, Malawi, Policy Maker] who said, ``\textit{The communities need to be empowered. We need the community voices so that they understand what is their role and responsibility, what are their rights to demand}''. Participants particularly emphasized the importance of ``home-grown'' solutions to increase adoption, citing, ``\textit{if it is homegrown and it's solving the problems that Africa is facing then it's easier for people to accept it}'' - P105 [30-39, Nigeria, Policy Maker]. 
Moreover, participants overwhelmingly stressed the importance of involving local communities to ensure culturally and contextually relevant solutions for AI for health in Africa, which will facilitate adoption of said solutions. This entails involving communities directly in the problem scoping and development phase, as stated by P059 [40-49, Kenya, Community Health Worker], ``\textit{if the communities are engaged at the point of the system then they can provide background information on the region the communities come from, the kind of cases that have been there, the kind of environment they live in and even the cultural perspectives that affect health thereby}''. P177 [21-29, Ghana, Medical Doctor] added to that sentiment with saying, ``\textit{AI researchers and developers should be willing learn from these communities and be able to work synergistically with them \ldots I think that is one way of ensuring that you’re getting different groups of people \ldots being open to understanding what the needs of the people are, their customs, their practices is the first step in ensuring that you’re being fair to them when you’re building AI products for them}''. Additionally, participants highlighted how involving local communities can ensure that solutions are correctly adapted to the customs of the communities they serve to facilitate adoption. As P158 [21-29, Kenya, Medical Doctor] put it: ``\textit{the learning model is made in regards to what people in the community will understand. So for example, go to a village and ask them what \ldots so how would you explain fever? What does fever feel for you? So there’s a way you can actually understand terms that people in the local area call it and then you’re going to be able to describe it that way}''. 

\paragraph{\bf Subtheme: Fostering trust through commitments, education, and good data practices:} In addition to involving local communities, participants highlighted the importance for providers of the technology to garner the trust of target users and stakeholders. As P036 [40-49, USA, Public Health Researcher] noted: ``\textit{when we implement any kind of intervention, digital health or not, having trust in the provider actually matters the most}''. Ways of building that trust include establishing long-term commitments, engaging with stakeholders to showcase the benefits of using AI tools in health, and providing assurance of the effectiveness of these models. P036 [40-49, USA, Public Health Researcher] explained: ``\textit{\ldots if you're going to introduce AI and machine learning [ \ldots ] you need to be able to show to health administrators, ministries of health that you plan to stay, and that you have a long term plan for being, you know, in that country and helping \ldots}''. P025 [50-59, Nigeria, Medical Doctor] also noted, ``\textit{So I think maybe having many pilot studies that can actually demonstrate success would be helpful}''.

Participants also mentioned how essential it is to address concerns about data ownership and sovereignty, ensuring data is used responsibly. While P025 [50-59, Nigeria, Medical Doctor] believes, ``\textit{at the end of the day if their data is collected and used to develop the algorithms, they would benefit because there is a lot of mistrust,}'' other participants addressed the need to collect representative data as well as utilize existing data for building effective models. For example, P177 [21-29, Ghana, Medical Doctor] said, ``\textit{We have to make sure that we are getting the right data. Once you feed it the right data \ldots it's going to return the right output}''. P177 [21-29, Ghana, Medical Doctor] further explained, ``\textit{one of the things is engaging its stakeholders. Making them see the need for access to the data \ldots first of all if you have data that you’re not using then it doesn’t make sense to go and collect different data if you already have data collected. So we have to work on that and then we have to also ensure that we are collecting good data from a wide range of sources}''.

Additionally, participants believed raising awareness about AI's potential can further foster trust P056 [40-49, Kenya, Public Health Researcher] noted, ``\textit{They need to be educated, they need to be engaged about these AI products that are available or are being researched on. They need to be told how they will work, the benefits and the risks of these products}''. P059 [40-49, Kenya, Community Health Worker] echoed this thought, stating, ``\textit{Creation of awareness, sharing of information that is based on facts with the reasonable data that will show how AI has been used elsewhere and improved access to health, this could be done through opportunities like in conferences, symposiums and even media like televised ads and the like. We need to bring it or make the community aware, we need to make the population aware of the advantage of using AI in health}''.

\subsubsection{\bf Theme: Opportunities for AI to Address Health in Africa}

Another major theme that emerged was the expert suggestions of opportunities for AI to address health concerns in Africa, encompassing their views on the desired features of AI solutions for better adoption. Overall, experts highlighted AI’s potential to democratize healthcare, namely the idea that AI could offer greater autonomy to African communities, reducing reliance on outside support by providing tools for self-diagnosis, prevention, and community-based care.

\begin{table}[!htbp]
\footnotesize
\centering
\caption{A summary of the theme and subthemes from the expert in-depth interviews}
\begin{tabular}{p{0.35\linewidth}|p{0.55\linewidth}}
\hline
\multicolumn{2}{c}{\textbf{THEME: Opportunities for AI to Address Health in Africa}} \\ \hline
\textbf{Subtheme} & \textbf{Representative quote} \\ \hline
State of AI in health in Africa & ``\textit{penetration to healthcare providers is very difficult…doing it with local associations… researchers [tends towards smooth] adoption}''  \\ \hline
AI as an administrative support tool for better patient registration, facilitating personalized treatment plans, and optimizing resource allocation & ``\textit{…taking time off these repetitive administrative tasks that the healthcare system [is] currently burdened with}'' \\ \hline
AI as a decision support tool for better diagnostics, reduced medical errors and unbiased treatment & ``\textit{a community health worker will take this tool, and based on what the technology is learning about the different patients, they’ll predict whether or not this patient needs to get referred to a hospital}'' \\ \hline
Opportunities for AI to address health inequities and reduce structural barriers in healthcare &
        ``\textit{AI can help … identify factors that contribute to health disparities.}''
        ``\textit{can help a lot to overcome the shortage [of doctors}''
        ``\textit{....provide healthcare information and services to people who do not have access}'' \\ \hline
Desiderata for AI solutions 
        & ``\textit{Africa [has] the most diverse population in terms of genes… we now need to focus on customizing these technologies to our contexts}''\\ \hline
Generative AI - specific features
        &``\textit{[AI] will really need to learn how…as a person from a community or a local person, I will describe my symptoms. It really needs to pick the context of language, specific language, specific names of things}'' \\ \hline
\end{tabular}
\label{tab:theme2}
\end{table}

\begin{figure*}[!htbp]
         \centering
         \includegraphics[width=0.75\textwidth]{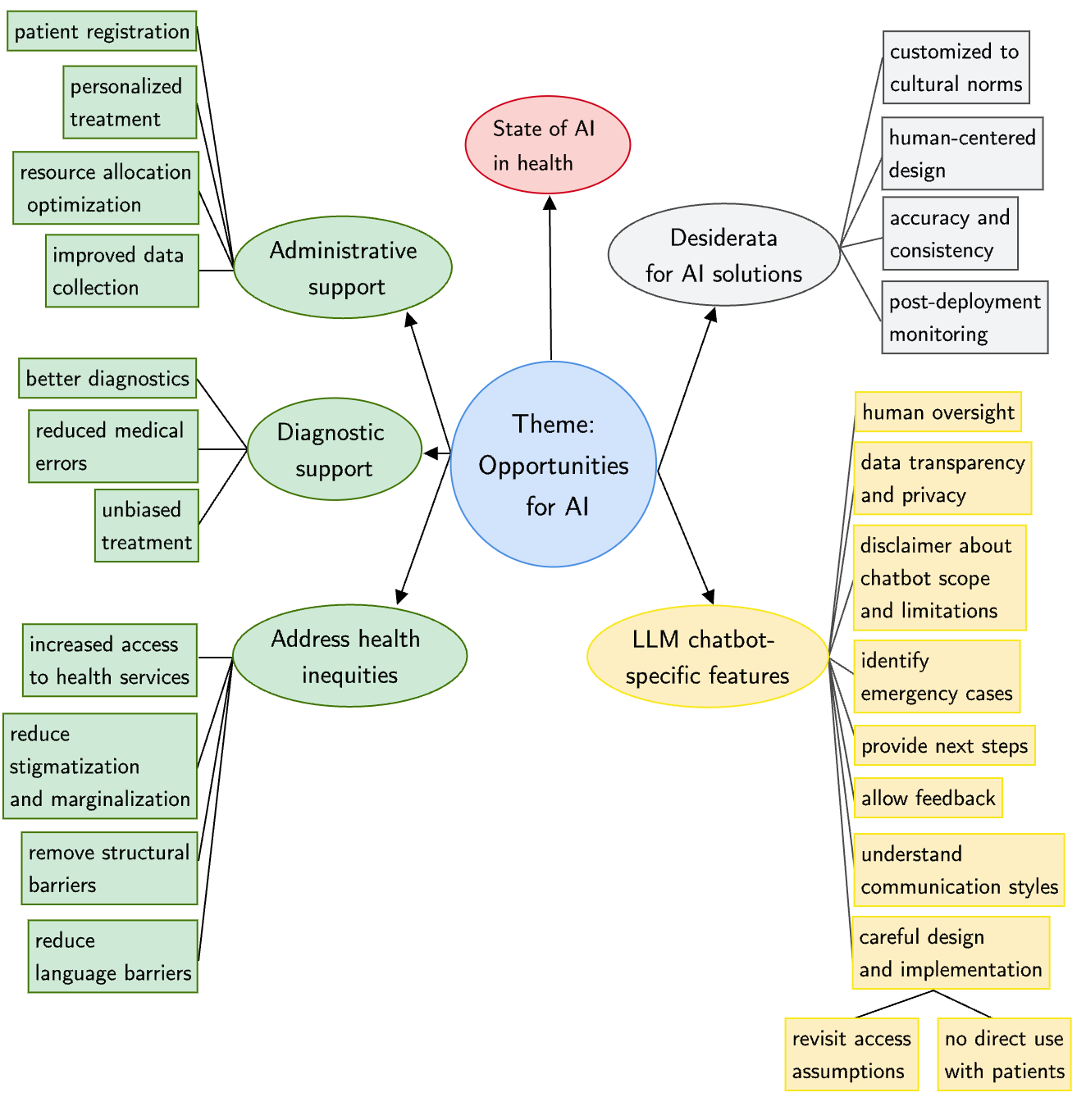}
        \caption{Major theme from the expert IDIs: \textit{Opportunities for AI to address health in Africa}. Summary of the subthemes illustrating the foremost use cases for AI in health and the desired features for AI solutions to improve efficacy and adoption. }
        \label{fig:trust_in_ai}
\end{figure*}

\paragraph{\bf Subtheme: State of AI in health in Africa:}  In discussing opportunities for AI to address health in Africa, most experts first referenced the current state of AI  for health in Africa. Many participants indicated that the technology was in its infancy on the continent and that many health practitioners were not aware of the scope of advancements AI can provide in healthcare: ``\textit{It is very, very early in African context}'' (P038 [30-39, USA, Machine Learning Practitioner/Engineer]), ``\textit{doctors have to be aware of AI and the use of AI, how to use AI, how to use mobile apps, how to use models, or desktop models, or APIs and so on. This is something I think lots of doctors are not aware of \ldots  I think they know about AI, but they don't understand what is AI. So they are afraid}'' (P176 [40-49, Morocco, Machine Learning Researcher]). Other participants noted that even in countries where AI has been adopted, use cases are limited to data collection and digitalization. This can be attributed to concerns around regulations, as noted by P176 [40-49, Morocco, Machine Learning Researcher], ``\textit{the penetration to the hospital or healthcare providers is very, very difficult in terms of regulation, and so on. And doing it with local associations, and local researchers and so on can be done smoothly and can tend towards adoption}''. When reflecting on the state of AI in healthcare, participants emphasized the need for more awareness campaigns at the level of governments, health workers, and the general population to facilitate adoption of AI solutions. P117 said, ``\textit{So, once they understand the importance of AI. I think they should be able to identify gaps that can be may resolved by AI applications,}'' echoed by P133 [30-39, Uganda] who recommended ``\textit{adoption of AI and machine learning practices into the medical curricula all over and this would allow the institutions that churn out medical practitioners to produce AI or ML competent practitioners and eventually an informed and empowered group of practitioners would allow us [to] harvest the full potential of AI}''. 

\paragraph{\bf Subtheme: AI as a support tool for better patient registration, facilitating personalized treatment plans, and optimizing resource allocation:} Participants expressed a shared view of AI’s potential to streamline administrative tasks such as patient registration and help in the prioritization of patients for care and management given the limited health workforce.  As P030 [30-39, Nigeria, Public Health Researcher] explains, ``\textit{I think that will be a game changer instead of taking time off these repetitive administrative tasks that the healthcare system [is] currently burdened with and allowing health workers to work at the scope of practice}''. P038 [30-39, USA, Machine Learning Practitioner/Engineer] also notes other ways in which AI can help facilitate personalized treatment plans: ``\textit{Being able to use AI whether it's population analytics, whether it is just being able to understand your history and provide you with recommendations, monitoring your care plan, making sure you're doing the right things, those are all places where AI can actually play a role}''.

In addition to simplifying administrative tasks, participants mentioned the role AI can play in improving data collection, which can offer insights into local health needs and conditions. For example, P029 [40-49, USA, Digital Health Lead for an NGO] said, ``\textit{\ldots helping alleviate their documentation burden. So making it easier to document the care that's received to the patients without them having to spend time typing or writing things down}''. P006 [30-39, Tanzania, Public Health Researcher] added, ``\textit{AI can be used to collect and analyze data about African communities. This data can be used to improve understanding of the needs of African communities and to develop policies and programs that address those needs}''.

Tangentially, participants noted that a key benefit of AI solutions is their ability to empower communities by offering essential health information on sensitive topics. This is particularly relevant for stigmatized communities who would otherwise not have access to critical health information. P038 [30-39, USA, Machine Learning Practitioner/Engineer] shared the anecdote based on a work experience: ``\textit{People were talking to the LLM in a way that was super comfortable. They were asking great questions. I'll give you an example. Someone said: ‘My sister is HIV positive. I might have a problem if I use her toothbrush’ \ldots  These are real questions that they're afraid to actually ask anyone else, but they were super comfortable asking an LLM about it \ldots  Things like that are super critical to where AI can actually play a role. Being a simple teaching tool with empathy, being there for diagnostics through the course of providing insights}''. AI’s ability to help provide contextual and empathetic suggestions for communication with patients was also mentioned by participants, including P029 [40-49, USA, Digital Health Lead for an NGO], who said ``\textit{and so there's no one there to mentor them and to tell them, this is how you take care of patients. This is how you deal with the community \ldots So figuring out how to use AI and like natural language processing to say \ldots  ‘don't talk that way to a patient’, you know, like ‘you talk this way’, that's nice and makes them feel like they can open up to you}''.

\paragraph{\bf Subtheme: Opportunities for AI to address health inequities and reduce structural barriers in healthcare:} Another opportunity for AI in health in Africa that participants stressed was surrounding health equity. Indeed, participants brought up the multiple ways in which AI can potentially reduce existing inequities in Africa, from increasing access to health services to enhancing the health workforce in remote areas. Specifically, participants recognized AI’s potential to decentralize healthcare access which reduces inequities due to geographic location and financial status. P018 [30-39, Mauritius, Medical Doctor] noted, ``\textit{AI can be used to develop telemedicine programs that provide healthcare services to people in remote areas. AI can also be used to develop mobile health applications that provide healthcare information and services to people who do not have access to healthcare facilities}''. Similarly, participants mentioned AI’s potential to ``\textit{improve health equity by providing access to healthcare services to people who live in rural areas and people who are poor}'' (P018 [30-39, Mauritius, Medical Doctor]), which would allow underserved populations to ``\textit{still access the same diagnostic services without necessarily incurring the economics or money expense to physically travel}'' (P059 [40-49, Kenya, Community Health Worker]). Participants noted how AI-powered solutions can streamline medical diagnosis, and thus ``\textit{reduce the time that you have to spend in a hospital}'' (P105 [30-39, Nigeria, Policy Maker]). Notably, this improved access ``\textit{would easily translate to better health outcomes and even better equity because the persons who would benefit most from faster and cheaper services are usually the disadvantaged,}'' as P011 [40-49, Nigeria Medical Doctor] pointed out. 

Beyond improving access to healthcare regardless of location and financial status, participants noted AI’s potential to reduce inequities due to stigmatization and marginalization. By providing an anonymous outlet, AI can provide useful health information on sensitive topics to communities who would otherwise be marginalized: ``\textit{for conditions which are stigmatized like abortion, like the sexual minorities, sexual orientation minorities, perhaps AI could come in to address their needs in a manner which is anonymous and confidential}'' (P011 [40-49, Nigeria Medical Doctor]). Participants also highlighted AI as a useful tool to reduce language barriers and improve better access to language minorities. While P018 [30-39, Mauritius, Medical Doctor] noted, ``\textit{AI can also help to translate medical information into different languages. This can help to improve access to healthcare for people who speak a language other than the dominant language,}'' P177 [21-29, Ghana, Medical Doctor] pointed out AI’s potential to reduce communication barriers by adapting to local languages and customs, ``\textit{I think AI for natural language processing. When taking into account African languages, right? I think that's also an area of application so that, for even those who cannot read and write, at least many of them can speak. And so you can, they can interact with these tools and get some benefit from them}''. 

Moreover, several participants mentioned opportunities for AI to remove structural barriers that plague most healthcare systems in Africa, including ``\textit{the lack of resources, poor infrastructure, and a shortage of healthcare workers}'' (P115 [21-29, Cameroon, Machine Learning Researcher]). The need to curb the shortage of skilled labor was a particularly important problem that many participants believe AI could address. As P002 [30-39, Egypt, Pharmacist] pointed out: ``\textit{in Africa we suffer somehow from the shortage of qualified healthcare workers so relying on AI tools that can help in diagnosis and how to manage certain diseases or conditions can help a lot to overcome the shortage}''. AI-aided diagnostics can enable quicker diagnostics and reduce heavy workload for health workers, as P158 [21-29, Kenya, Medical Doctor] noted, ``\textit{AI is going to help with the seeing of more patients and quicker diagnosis and yeah, so lesser mortalities in those regions. So I think just increasing the number of patients who can be seen quickly with minimal errors, because the healthcare workers are overworked at a short period of time. So I think that's basically what it can do to reduce inequities}''. This is particularly important in rural areas where the ``\textit{doctor-to-patient ratio is very low}''. Finally, participants discussed how AI can serve as a tool for health monitoring and data collection which could lead to more ``tailored solutions'' (P105 [30-39, Nigeria, Policy Maker]) and policies. As P018 [30-39, Mauritius, Medical Doctor] puts it, ``\textit{AI can help to identify and reduce health disparities. AI can be used to collect data on health outcomes, and it can be used to identify factors that contribute to health disparities. This information can be used to develop policies and programs that aim to reduce health disparities}''.

\paragraph{\bf Subtheme: AI as a decision support tool for better diagnostics, reduced medical errors and unbiased treatment:} A key area of opportunity for AI in health in Africa is in decision-support according to the data. Participants frequently discussed opportunities for AI to improve medical diagnosis, enhance disease monitoring and forecasting, and provide decision support especially in regions with a reduced number of health specialists. Many participants believe AI can ``\textit{aid in early disease detection and outbreak management}'' (P111 [30-39, Nigeria, Medical Doctor]). They also see potential for AI to ``\textit{intervene a lot in data interpretation and data analysis or forecasting}'' (P002 [30-39, Egypt, Pharmacist]). This could improve understanding of health needs and ``\textit{emerging infectious diseases}'' (P002 [30-39, Egypt, Pharmacist]). Participants also described AI’s potential as a decision-support tool, providing valuable insights into health history, treatment, and patient management. For example, P036 [40-49, USA, Public Health Researcher] said ``\textit{a community health worker will take this tool, and based on, you know, what the technology is learning about the different patients, they’ll predict whether or not this patient needs to get referred to a hospital}''. Given the ``\textit{limited infrastructure}'' (P129) in some areas, participants noted the importance of early detection using AI, explaining that ``\textit{if you're able to diagnose diseases quite early, I think you'd be able to make a decision about the patient or about the clinic in general}'' (P117).

They also emphasized the importance of using AI tools to ``augment'' human doctor capabilities rather than replace available doctors. P038 [30-39, USA, Machine Learning Practitioner/Engineer] stated, ``\textit{We believe that providers and the AI can work really well together. The AI is not looking to replace their expertise \ldots  but when they work together, they're actually augmenting each other}''. This collaborative approach could be particularly beneficial in areas with limited access to specialists, as AI can provide a valuable second opinion and help doctors make more informed decisions. For example, P158 [21-29, Kenya, Medical Doctor] stated, ``\textit{I would want an AI machine that helps me, you know, second-think my decisions. So it's going to give me a lot more confidence,}'' echoed by P010 [30-39, Kenya, Research and Policy Analyst], who believes AI would be helpful ``\textit{in terms of really increasing the accuracy, the confidence of the healthcare providers in terms of just giving the right medication and providing the right treatment}''.

When discussing the use of AI for medical diagnosis, participants presented a view of AI as potentially providing more accurate diagnoses and fair medical assessments. P055 [30-39, Kenya, Clinical Officer]  believes that with AI ``\textit{you are going to reduce misdiagnosis that is common in the current healthcare system}''. P111 [30-39, Nigeria, Medical Doctor] shared this sentiment, stating that AI ``\textit{could reduce subjectivity and bias, leading to better diagnosis. And in addition to better diagnosis, leads to better decision making}''.  
Overall, participants expressed optimism about the potential of AI to enhance healthcare in Africa, viewing it as a means to increase efficiency, accessibility, and equity, as summarized by P115 [21-29, Cameroon, Machine Learning Researcher]: ``\textit{AI has the potential to make healthcare more efficient, accessible, and equitable}''.

\paragraph{\bf Subtheme: Desiderata for AI solutions:} When discussing AI solutions in health in Africa, participants specified several desiderata for solutions to meet before and after deployment. They highlighted the need to develop contextually relevant solutions that are adapted to the specificities and customs of the population they target. For example, P036 [40-49, USA, Public Health Researcher] said ``\textit{\ldots being sensitive to \ldots  the cultural norms of each country is always important}''. P038 [30-39, USA, Machine Learning Practitioner/Engineer] shared an example of adaptation from their work, ``\textit{our primary goal is, ‘You probably already use WhatsApp to communicate with a lot of people. You already probably use an application for other things. Just add us as part of your workflow.’ We don't have to actually replace it}''. P038 [30-39, USA, Machine Learning Practitioner/Engineer] mentioned useful questions that developers should ask when designing AI tools, including ``\textit{What are the digital adoption? What are people actually using over there?,}'' to ensure that ``\textit{when you're building AI, you're building it for that particular context}''. Other participants favored a human-centered design approach, proposing to build  ``\textit{user friendly}'' tools (P025 [50-59, Nigeria, Medical Doctor]) and ensuring that these are ``\textit{user-informed and user-engaged}''. (P030 [30-39, Nigeria, Public Health Researcher])

In addition to being contextually relevant, participants emphasized that accuracy and consistency are crucial factors before deploying AI tools in health. As P039 [40-49, Nigeria, Telemedicine and e-health researcher] stated ``\textit{in decision support, especially when it comes to patients' decisions where life is involved \ldots you have the accuracy to be 70 something point something that's not acceptable in medicine}''. P055 [30-39, Kenya, Clinical Officer]  added ``\textit{AI should aim for consistent results across different groups}''. This is particularly important in the African context given the genetic and cultural diversity of the population, ``\textit{Africa is known to have the most diverse population in terms of genes, in terms of all these other characteristics patients might have \ldots what we now need to focus more on is customizing some of these technologies to our contexts}'' - P143 [21-29, Zimbabwe, Medical Doctor].

Finally, some participants mentioned the need for ``post deployment monitoring'' to address any downstream bias (P038 [30-39, USA, Machine Learning Practitioner/Engineer]). This adds to P133 [30-39, Uganda]’s call for regulation to ensure safe use of these AI solutions in Health: ``\textit{there should be regulators of AI \ldots but the ultimate authority over acceptance and allowing usage of healthcare AI would be vested in the ministries of health’s hands and this comes with responsibility of making sure that different AI systems have been tested against all extreme unacceptable stereotypes, and this would allow a normal user to use the system based on trust and with assurance}''.

\paragraph{\bf Subtheme: Generative AI/LLM chatbot specific features:} When presented with the case study around a hypothetical chatbot providing health diagnosis through text and speech communication, participants discussed several chat-bot specific features that would be important in increasing the reliability and trust of such a tool.  
Participants overwhelmingly shared requests for safeguards like human oversight, strong data privacy and transparency as critical components of these tools. They highlighted the need for reassurances regarding confidentiality of the chatbot discussion and transparency around how the data will be collected and used. For example, P008 [30-39, Ethiopia, Public Health Researcher] said, ``\textit{it has to assure also the data privacy that Chatbot or the application will not use or will not disclose their medical histories or whatever the data the user is feeding to another or third party}''. This reassurance is particularly important for users who are likely to share sensitive information, as P019 [40-49, France, Public health NGO chief executive ] pointed out, ``\textit{would the interaction with the machine actually help me express myself without the fear of judgment and therefore receive potential recommendation that would be stronger?}'' To increase trust, participants also referred to transparency in how the chatbot reaches its conclusions as a critical feature. P038 [30-39, USA, Machine Learning Practitioner/Engineer] described it as \textit{it's going to be really important that the Chatbot when it provides the analysis and diagnosis, actually talks about how it got to that, what factors it took into consideration and why this is relevant for the patient}''. Still with regards to increasing trust, several participants addressed the need for disclaimers around the AI’s purpose and limitations. This is important for setting correct expectations with the user, as P002 [30-39, Egypt, Pharmacist] explains, ``\textit{I believe that it should have some sort of disclaimers as well, like what you can or cannot expect from the chatbot or describing the extent of its abilities in the diagnoses or medical counseling}''. 
 
Prior to discussing user-specific needs for interacting with the chatbot, participants emphasized the importance of ensuring that the chatbot can rapidly identify urgent and emergency cases as well as provide clear next steps after assessment. P025 [50-59, Nigeria, Medical Doctor] shared, ``\textit{the chat-bot needs to be able to know what is an emergency and how do they counsel and ensure this person gets to care}''. P038 [30-39, USA, Machine Learning Practitioner/Engineer] added, ``\textit{Whether it is handing me off to a doctor to continue the conversation with, whether it is telling me that, ‘Hey, you should go to a local health facility and here's the closest health facility.’ Whether it's saying, ‘Hey, you know, I can take you to a site where you can order something.’ But there needs to be a clear next step, if there's a particular suggestion, just so the patient is not left hanging and guessing as to what they need to do}''.

To ensure successful communication with the users, participants cited the need for the chatbot to be able to take into account interpersonal variations in communication styles and cultural contexts. They highlighted how ``\textit{the machine will really need to learn how as a community member or as a person from a community or a local person how I will describe even my symptoms. It really needs to pick the context of language, specific language, specific names of things}'' (P010 [30-39, Kenya, Research and Policy Analyst]). With speech communication, the ability to comprehend diverse pronunciations and accents is also crucial, as regional variations can significantly impact interpretation, as noted by P006 [30-39, Tanzania, Public Health Researcher], ``\textit{The pronunciation of words as we said here, text and speech, how some words can be pronounced differently. So I think these are some of the things to be considered that the locality and how the English Kenyans speak may be different from how Ugandans speak}''. Additionally, participants emphasized the importance of having the ChatBot's communication style be natural and user-friendly, mimicking human interaction. For example, P038 [30-39, USA, Machine Learning Practitioner/Engineer] shared, ``\textit{I should be able to communicate with the ChatBot in a natural way, right? So that I don't have to think about how I should phrase this particular thing?}'' This is important to build trust and encourage open communication.

Finally, several participants mentioned that having personalized communication with the chatbot and being able to provide feedback on the accuracy of the assessment should be available in order to improve future communications. This can help prevent future misdiagnoses, as P038 [30-39, USA, Machine Learning Practitioner/Engineer] described, ``\textit{Being able to actually take back feedback when a medical provider actually disagrees with the AI and using that to actually either feedback in to tune better to ensure that any biases that were in play are taken care of so that it doesn't happen again in the future}''.

A few participants contested the assumption of widespread smartphone access and reiterated  the critical need for careful design and implementation: ``\textit{\ldots when I think about a chatbot, it's like, you already are assuming a couple of things, that this person has a smartphone [laughs]. Because there are still simple cell phones, right?}'' - P036 [40-49, USA, Public Health Researcher]. Some participants also objected to using such a chatbot directly with patients. They propose to use such chatbots as a preliminary step for patient preparation and insight while leaving the final decision to a physician. To that end, P019 [40-49, France, Public health NGO chief executive ] noted ``\textit{if it could be used as a preliminary, as a way to prepare your patient for the visit, for example, so it’s done as a preliminary activity before the person even reaches the hospital}''. P002 [30-39, Egypt, Pharmacist] added, ``\textit{I also don’t want the chatbot to directly answer the patient but instead the chatbot to discuss things or give options to a physician then a physician to decide the final decision for the patient}''.

\subsubsection{\bf Theme: Intersectional Inequities and Systemic Factors Impacting Health Equity in Africa}

Experts described various factors, both individual and systemic, that impact health equity in Africa, as well as proposals for addressing existing inequities. Experts broadly identified social determinants of health, infrastructure limitations, and socioeconomic disparities steaming from the history of colonialism as key contributors to health inequities. Experts stressed that while targeted interventions can address specific barriers like geographical limitations or lack of funding, there is a need for a comprehensive approach to sustainably tackle health inequities in Africa.

\begin{table}[!htbp]
\footnotesize
\centering
\caption{A summary of the theme and subthemes from the expert in-depth interviews}
\begin{tabular}{p{0.35\linewidth}|p{0.55\linewidth}}
\hline    
\multicolumn{2}{c}{\textbf{THEME: Intersectional Inequities and Systemic Factors}}\\
\multicolumn{2}{c}{\textbf{Impacting Health Equity in Africa}} \\ \hline
\textbf{Subtheme} & \textbf{Representative quote} \\ \hline
Attributes which impact health equity & ``\textit{we have some cultures that prohibit a woman…to seek healthcare in the health facilities}''
        -``\textit{LGBTQ communities face health inequities in general because of stigma and judgment}''
        -``\textit{People who speak a language other than the dominant language have less access to healthcare}'' \\ \hline
Systemic barriers to health equity
        & ``\textit{Our biggest challenge or inequities in health is on access, by access I mean the capacity of community members or people to access different health services, for different reasons, this is contributed to by the poor road network, the lack of adequate equipment and human resource in terms of skill}'' \\ \hline
Addressing health inequities in Africa
        & ``\textit{It’s like a whole systems approach and it’s not a one-time, five-year, 10-year project…you’re talking investments over the next, at least five decades}'' \\ \hline
\end{tabular}
\label{tab:theme3}
\end{table}

\begin{figure*}[!htbp]
         \centering
         \includegraphics[width=0.75\textwidth]{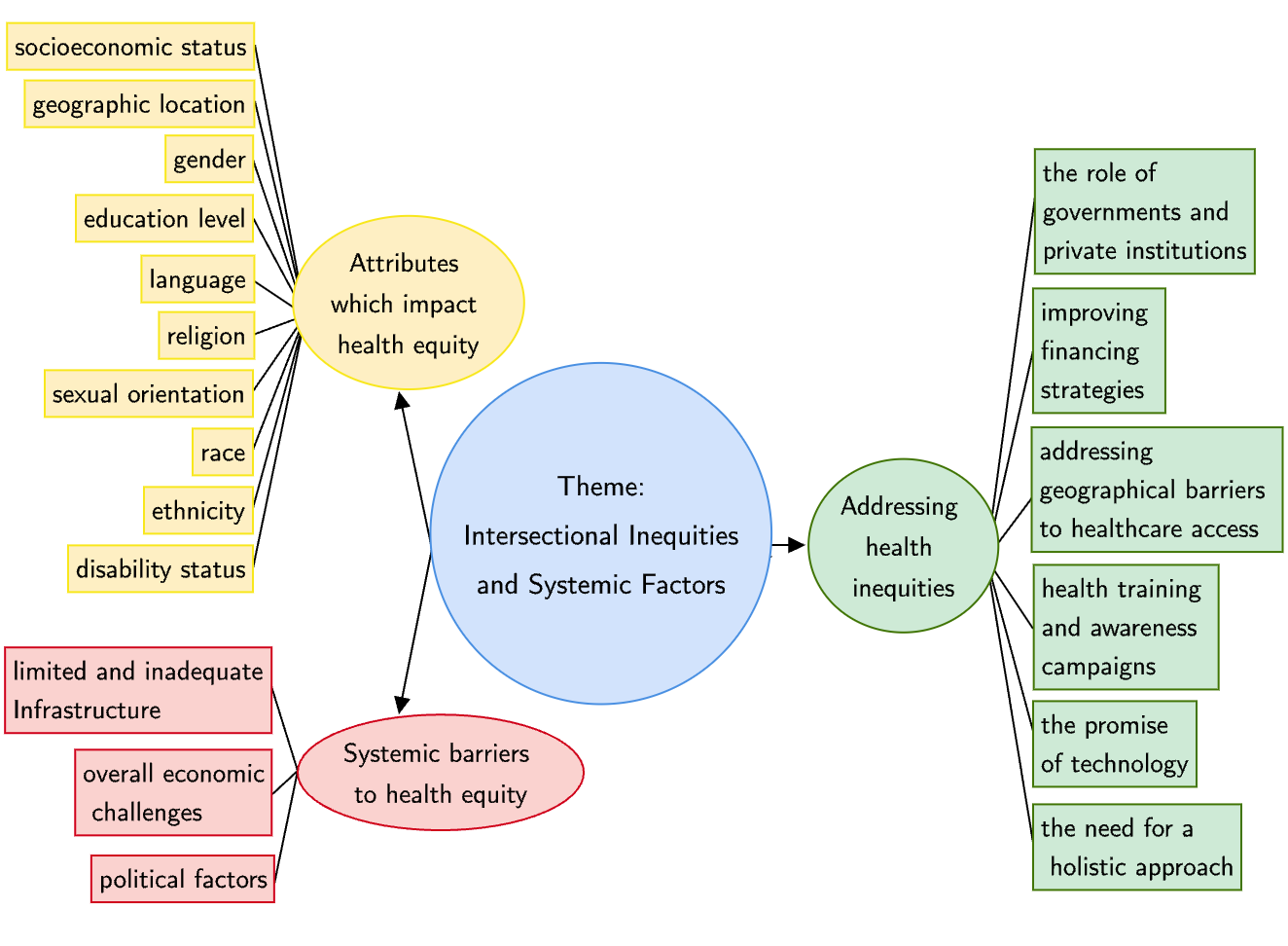}
        \caption{Major theme from the expert IDIs: \textit{Intersectional Inequities and Systemic Factors impacting health equity in Africa}. Summary of the subthemes illustrating the existing causes of health inequities and proposed approaches for addressing them. }
        \label{fig:trust_in_ai}
\end{figure*}

\paragraph{\bf Subtheme: Attributes which impact health equity:} Participants identified several factors which contribute to health inequities in Africa, namely socioeconomic status, geographic location, gender, religion, education level, language, sexual orientation, race, ethnicity, and disability status. 

\textbf{\textit{Socioeconomic status}} was frequently mentioned as a significant driver of health inequities due to most patients having to ``pay out-of-pocket'' for healthcare services. As P018 [30-39, Mauritius, Medical Doctor] noted, ``\textit{Income is a major factor that contributes to health inequities. People who are poor are less likely to have access to healthcare, and they are less likely to receive quality healthcare when they do have access}''. These financial disparities lead to a ``social-economic stratified'' system where  ``\textit{the rich can afford to go to private hospitals, while the poor have to rely on public hospitals, which are often underfunded and understaffed}'' (P066 [40-49, Nigeria, Medical Doctor]). 

\textbf{\textit{Geographic location:}} Participants also mentioned geographic location as another key determinant of health equity. Indeed, healthcare access is heavily skewed towards urban areas creating a rural vs. urban divide where communities in rural areas often lack access to the same quality of care available in urban centers. P158 [21-29, Kenya, Medical Doctor] described this divide, stating ``\textit{people in rural areas don't have the same kind of medical care, they don't have the same access to the kind of doctors and consultants that people in the urban centers have. So one it's quality of care, quality of services, equipment or facilities, the facilities in urban centers look better than the facilities in rural areas}''. The lack of infrastructure in rural areas further deepens health inequities by requiring additional expenses to travel to available healthcare providers. P066 [40-49, Nigeria, Medical Doctor] shared a personal account to this effect: ``\textit{I live in a rural area, and there is no hospital nearby. If I get sick, I have to travel a long way to the nearest hospital, which is very expensive}''.

\textbf{\textit{Gender}} was also mentioned as a source of inequity in healthcare access. As P029 [40-49, USA, Digital Health Lead for an NGO] said, ``\textit{There is a lack of access to healthcare, [and] difficulty for women getting access to services}''. This inequity is often compounded by socioeconomic factors and cultural norms that restrict women's autonomy. For example, P029 [40-49, USA, Digital Health Lead for an NGO] shared how ``\textit{in some places they can't go without their husband}''. P073 [30-39, Nigeria, Medical Doctor] shared a similar thought, stating ``\textit{We do not have equal access to quality health care in Nigeria \ldots  and it's worse among \ldots  women and children. There are some places where the cultural norms do not allow for equitable access to quality healthcare. So for instance, we have some cultures that prohibit a woman \ldots to seek healthcare in the health facilities}''. On the other side, P056 [40-49, Kenya, Public Health Researcher] shared how men can also be affected by social and cultural factors that may prevent them from visiting health facilities: ``\textit{a lot of men do not seek health services until they’re seriously ill, that’s the time they’ll be able to go to a health facility for services or for treatment}''.

\textbf{\textit{Education level:}} Several participants highlighted the impact of education and literacy on healthcare access, honing on the point that limited education can restrict opportunities and perpetuate health inequities. P143 [21-29, Zimbabwe, Medical Doctor] explained that ``\textit{the level of education of someone in Africa especially affects the level of opportunities you get, which at the end of the day affects the income people have, affects what kind of services they can access, who they can access them from}''. The degree of literacy and education can hinder people’s understanding of health issues and available resources, prompting them to turn to traditional medicine instead of seeking formal medical care. As P133 [30-39, Uganda] pointed out: ``\textit{Another is literacy. In Africa, we still have the belief in traditional medicine, and this somehow contributes to other people choosing not to go to medical facilities and instead go into traditional shrines}''.

\textbf{\textit{Language:}} Given the diversity of languages spoken on the African continent, language barriers can lead to further disenfranchisement from groups who do not speak the dominant language. This point was exemplified by P018 [30-39, Mauritius, Medical Doctor], who stated ``\textit{People who speak a language other than the dominant language have less access to healthcare than those who speak the dominant language}''. Participants further underscored how language barriers can lead to misunderstandings and compromise the quality of care, ultimately hindering effective healthcare delivery. As P039 [40-49, Nigeria, Telemedicine and e-health researcher] puts it, based on personal experience: ``\textit{[Language barriers] put a limit to communication between the healthcare practitioner and the patients. For instance, there are a number of languages in Nigeria \ldots So if I were to be a medical doctor, I don't understand Hausa language, and I'm posted to the north now, and I'm posted to a rural area to serve. Now, if the individual I'm going to attend to doesn't speak English, there's a limit to the medical service that the individual can assess from me as a doctor who doesn't speak their language and again, there's going to be a loss in transcription of language}''.

\textbf{\textit{Religion:}} Some participants noted religion as a factor impacting health equity with certain religious beliefs acting as a barrier to seeking healthcare in Africa. For example, P045 [30-39, Kenya, Machine Learning Practitioner/Engineer] shared that ``\textit{a lot of people will tend to seek \ldots the religious intervention in terms of  \ldots  being cured of a particular disease}''. Others mentioned barriers due to specific religious beliefs which discourage them from using health facilities. As P056 [40-49, Kenya, Public Health Researcher] explained ``\textit{in Africa, there are some religions that do not allow you to go to the hospital even when you’re sick. You have to come and seek guidance from your religious leader first, so a lot of people lose their lives at this  \ldots  Some religious beliefs do not allow more than medicine or more than health seeking behaviors}''. As a result, many people may delay or avoid seeking necessary care, leading to negative health outcomes.
However, one participant disagreed with the extent to which religion impacts health equity, downplaying its significance compared to other factors like ethnicity and income. P176 [40-49, Morocco, Machine Learning Researcher] stated: \textit{I collaborate with people from Africa, but I never find that religion is a problem. The most problem maybe is the ethnicity \ldots And in terms of income, people are treated differently}''.

\textbf{\textit{Sexual orientation, Race, Ethnicity, Disability status:}} Some participants mentioned inequity in healthcare access for sexual minorities, mostly due to stigma and discrimination: ``\textit{when you actually think about the LGBTQ communities, they actually face a lot of health inequities in general because of stigma and judgment. Those are the key population groups, I think, in Africa that are impacted by the health inequities}'' - P038 [30-39, USA, Machine Learning Practitioner/Engineer]. This inequity can be exacerbated for these individuals when seeking care for issues related to their sexuality as P055 [30-39, Kenya, Clinical Officer]  noted ``\textit{In Kenya, the sexual minorities find access to healthcare very difficult. These are the things where they go for healthcare for situations or conditions that are related to their sexuality, they will be discriminated}''. A few participants also noted that certain ethnic groups and people with disabilities can experience unequal access to healthcare. For example, P172 [21-29, Tunisia, Medical Doctor] stated ``\textit{sometimes in some countries there is something related to race, and that's not related to white versus black, for example, but also some ethnicities inside the same country actually can suffer from that}''. For P111 [30-39, Nigeria, Medical Doctor], ``\textit{persons with lower education and persons with disability \ldots usually have the worst outcomes}''.

\paragraph{\bf Subtheme: Systemic barriers to health equity:} Beyond individual factors, participants identified broader systemic issues that significantly contribute to health inequities in Africa. These include infrastructural challenges, continental-level socioeconomic factors, and political incentives. 

\textbf{\textit{Infrastructural Barriers:}} Several participants noted how inadequate infrastructure, such as poor road networks and ill-equipped facilities, coupled with a shortage of skilled healthcare workers, create substantial barriers to healthcare access. For example, P039 [40-49, Nigeria, Telemedicine and e-health researcher] shared a personal anecdote to this point, saying ``\textit{If I'm supposed to have a class in about five minutes’ time, I will leave four hours before that time. Because you can just be held in traffic. Now, imagine if there's an emergency and you put a patient in a car and you try to rush the person to the hospital in Lagos. I don't know if the person survives anyways, but I feel the congestion in major cities is also a problem that we are dealing with in accessing healthcare}''. This lack of access is further compounded by inadequate equipment in many facilities and lack of qualified healthcare workers as P059 [40-49, Kenya, Community Health Worker] added ``\textit{Our biggest challenge or inequities in health is on access, by access I mean the capacity of community members or people to access different health services, for different reasons, this is contributed to by the poor road network, the lack of adequate equipment and human resource in terms of skill}''.

\textbf{\textit{Socioeconomic Factors:}} When discussing the state of health equity in Africa, some participants attributed the lack of equitable healthcare access to the continent’s overall economic challenges which limit the availability and distribution of resources. As P010 [30-39, Kenya, Research and Policy Analyst] explained ``\textit{as a continent we are economically disadvantaged and therefore that also attributes more to this growing gap of the inequities of the health status and health outcomes and the distribution of resources}''. The pervasive poverty in many African countries is a significant contributor as well, as it directly impacts individuals' ability to afford healthcare as highlighted by P066 [40-49, Nigeria, Medical Doctor],``\textit{Poverty is the main cause of health inequities in Africa. People living in poverty are less likely to have access to healthcare, nutritious food, and clean water}''. Rural areas are particularly affected due to profit incentives that discourage establishing healthcare facilities in those areas. This leaves these communities with only basic primary healthcare centers as P039 [40-49, Nigeria, Telemedicine and e-health researcher] pointed out:  ``\textit{Individuals who want to establish do not also want to put their health facilities there because of gains. So that means in the rural community, you only have primary health centers and what they do is just basic healthcare care delivery}''. 

\textbf{\textit{Political factors}} were also brought up as another systemic barrier. P038 [30-39, USA, Machine Learning Practitioner/Engineer] noted how changes in political leadership can impact healthcare access through shifting regulations, ``\textit{because of changes in government, sometimes regulations actually change, sometimes access to preventative medication changes}''. Additionally, P055 [30-39, Kenya, Clinical Officer]  highlighted that favoritism due to tribal affiliation can potentially affect equitable resource allocation, explaining ``\textit{there is always that inequality when the sitting president is not from your tribe, resources to your community will not be as much compared to the people whose president come from their tribe}''. This leads to communities whose members are not aligned with the ruling party receiving fewer resources.

\paragraph{\bf Subtheme: Addressing health inequities in Africa:} To address the individual and systemic barriers to health equity in Africa, experts proposed various solutions, ranging from governmental interventions to health awareness campaigns and the utilization of technology.

\textbf{\textit{The role of governments and private institutions:}} Participants, for the most part, pointed to the critical role of government and policy in addressing health inequities in Africa. Many called for better health resource allocation in order to ``\textit{avail the healthcare service throughout the population on equitable basis}'' (P008 [30-39, Ethiopia, Public Health Researcher]). This would involve increasing health funding, especially in rural areas, as P056 [40-49, Kenya, Public Health Researcher] noted ``\textit{one main thing i want our government to do is to make sure that we fund our health sector, we employ the necessary workers, healthcare workers to work not just within the city center but in the rural areas where services are needed}''. P033 [30-39, USA, Doctoral candidate ] called for improved financing strategies by ``\textit{lobbying parliament to contribute more to the national health budgets,  \ldots  holding policymakers accountable for any declarations that they make in terms of contributing to health financing budget}''. Additionally, participants stressed the need for proactive policies aimed at ``restructuring healthcare delivery'' and promoting healthcare efficiency. As P045 [30-39, Kenya, Machine Learning Practitioner/Engineer] put it ``\textit{African governments are able to streamline health, \ldots formulating the right policies, policies that are geared towards improving health, health facilities, health services}''.  
At the private sector level, participants highlighted  the need for social responsibility from private companies, ``\textit{making sure that they are contributing to the communities that they're taking from}'' (P029 [40-49, USA, Digital Health Lead for an NGO]).  They also emphasized the essential role of health insurance in expanding access to healthcare services, stating that ``\textit{health insurance will become very key in terms of addressing the issue of health services provision}'' (P045 [30-39, Kenya, Machine Learning Practitioner/Engineer]). Beyond independent governmental and private solutions, P059 [40-49, Kenya, Community Health Worker] advocated for collaboration between different sectors to address health inequities at the community level, stating \textit{local community, the government and non-governmental organizations need to work hand in hand  \ldots  the local community and non-governmental organizations can support the government to ensure health services or health inequities are tackled at the community level}''.

\textbf{\textit{Improving financing strategies:}}Several participants expressed concerns about the current approach to funding health initiatives in Africa. For example, P014 [40-49, Malawi, Policy Maker] noted that while \textit{donors have been pushing lots of money to Africa \ldots the impact in terms of addressing health inequities hasn't been really meaningful}''. They recommended some improvements including focusing on ``\textit{supporting programs that address the underlying social determinants of health, rather than simply providing disease-specific treatment}'' (P014 [40-49, Malawi, Policy Maker]). Additionally, some participants raised the issue of high interest rates from international institutions, which create a significant burden on healthcare funding in many African countries. They proposed collective action from these countries to solve this issue: ``\textit{collectively it's not an individual country, but I think collectively. If countries come together and say, look, we can’t pay this. It's not equitable and we're not gonna do it anymore}'' - (P029 [40-49, USA, Digital Health Lead for an NGO]). A few participants underscored the importance of fostering innovation and supporting local health startups. For example, P045 [30-39, Kenya, Machine Learning Practitioner/Engineer] noted the importance of ``\textit{being able to power the startups, especially the health startups and startups within the health industry}''.

\textbf{\textit{Addressing geographical barriers to healthcare access:}} In order to address inequities due to geographical barriers, participants proposed exploring novel approaches to healthcare delivery. For example, P013 [30-39, Zambia, Public Health Researcher] highlighted the potential of telemedicine and video conferencing, stating ``\textit{It really boils down to being able to provide them services without having to be physically there. Till we figure out someone else using the telephone or video conferencing to be able to provide health care to people who are far off}''. In that same vein, P117 shared a concrete example of such innovation, stating, ``\textit{we've seen programs where for example, vaccines are delivered via drones, in hard to reach areas. So that has proved to be a solution}''. Other participants have discussed improving the planning and placement of health facilities to ensure equitable resource distribution. They suggest utilizing census data to ``\textit{see how many people are not having access or how many health facilities are in a certain region versus how people are spread. And then we decide to set up those health facilities based on the information we have gathered}'' (P138 [30-39, Uganda, Electrical engineering]).

\textbf{\textit{Health training and awareness campaigns:}} Participants have noted a lack of general health literacy among the general population. As P014 [40-49, Malawi, Policy Maker] pointed out, ``\textit{I think that's one of the biggest challenges in Africa, is that people don't really know what their rights are when it comes to health care. They don't know how to access care, and they don't know how to advocate for themselves}''. A possible solution raised was investing in education and awareness campaigns to empower populations to make informed decisions about their health. This is particularly important given the many barriers to accessing healthcare that are tied to cultural beliefs. For example, P059 [40-49, Kenya, Community Health Worker] believes that ``\textit{awareness creation and creating forums where knowledge is shared with different target groups will assist in demystifying the wrong perceptions that specific genders should be allowed to access health services}''. Such campaigns are also important in decreasing the knowledge gap between certain communities. P059 [40-49, Kenya, Community Health Worker] proposed having ``\textit{health talk sessions with different ethnic groups on different health topics, thereby identifying where they have got knowledge gaps on information regarding health issues and thereby encouraging them to access the health services despite the long distances they may have to cover}''. In addition to educating the general population, training clinical health workers was also highlighted as a crucial step in reducing health inequities. Indeed, P145 [40-49, Egypt, Pharmacist] emphasized the shortage of qualified physicians, stating ``\textit{one of the most challenges we face in Africa is the presence of qualified, especially physicians in every facility}''. Proposed solutions include increasing the ``\textit{training [of] clinical health workers}'' (P038 [30-39, USA, Machine Learning Practitioner/Engineer]), particularly in rural areas, and developing ``\textit{customized medical training programs}'' (P172 [21-29, Tunisia, Medical Doctor]) to address specific local needs. 

\textbf{\textit{The promise of technology:}} While discussing solutions for health inequities in Africa, participants touched on the potential of AI to improve healthcare access and efficiency. For example, P033 [30-39, USA, Doctoral candidate ] believes existing initiatives that ``\textit{are using chatbots and different mobile devices to triage care \ldots can really assist in terms of keeping costs down and so really leveraging these– the use of mobile phones \ldots can help for urban and for rural folks}''. Participants also described AI-aided decision making as a solution for improved health outcomes. 

\textbf{\textit{The need for a holistic approach:}} Participants acknowledged that addressing health inequities requires a long-term, holistic approach that considers the historical context of colonialism. P030 [30-39, Nigeria, Public Health Researcher] stressed that solutions must first ``\textit{take into account the colonial historical structures that have displaced power from communities, and in which current health inequities are rooted}'' and ``\textit{dismantling those structures}''. Furthermore, they emphasized the broader systemic challenges to health equity in Africa and the need for long-term, sustainable solutions. P030 [30-39, Nigeria, Public Health Researcher] questions ``\textit{How do we ensure adoption, implementations, sustainability, skill and spread?}'', highlighting the need for long-term commitments, ``\textit{It’s like a whole systems approach and it’s not a one-time, five-year, 10-year project \ldots you’re talking investments over the next, at least five decades just to [establish] the right architecture for the right framework and governance model}''.

\subsubsection{\bf Theme: Social and Ethical Considerations for Developing and Deploying AI Solutions in Health in Africa} Experts articulated the importance of addressing social and ethical factors when introducing AI technologies in healthcare in Africa. They highlighted the necessity of developing localized solutions that are sensitive to social behaviors, advocating for community engagement throughout the development and deployment of AI solutions. Experts also emphasized addressing existing inequities through equitable AI deployment and prioritizing language accessibility. The data underscores that the safe and effective implementation of AI in healthcare in Africa hinges on ethical considerations, bias mitigation strategies, establishing clear regulatory frameworks, and addressing data representativeness, scarcity, collection and privacy. We detail these considerations below. 

\begin{table}[!htbp]
\footnotesize
\centering
\caption{A summary of the theme and subthemes with representative quotes from the expert in-depth interviews}
\begin{tabular}{p{0.35\linewidth}|p{0.55\linewidth}}
\hline
\multicolumn{2}{c}{\textbf{THEME: Social and Ethical Considerations for Developing}} \\ 
\multicolumn{2}{c}{\textbf{and Deploying AI solutions in Health in Africa}} \\ \hline
\textbf{Subtheme} & \textbf{Representative quote} \\ \hline
Social considerations
        &``\textit{I don't want AI to take the place of a bedside carer...  African cultures are far more communal … and I would hate to see elders and people in our community losing that…relationship with doctor, nurse, midwife that's built over time}''\\ \hline
Ethical considerations
        & ``\textit{[governments are] not really talking about regulating these tools and how these tools should be used and the data that is used to build these tools}''\\ \hline
Data issues
        & ``\textit{some of these models that we are expected to adopt in Africa, have been developed using datasets that don't really reflect our populations…}''\\ \hline
Security and privacy
        & ``\textit{There are already data privacy acts in most African countries…whatever data one is collecting has to be in line with the data privacy act of that particular country, or state, or city}''\\ \hline
\end{tabular}
\label{tab:theme4}
\end{table}

\begin{figure*}[!htbp]
         \centering
         \includegraphics[width=0.75\textwidth]{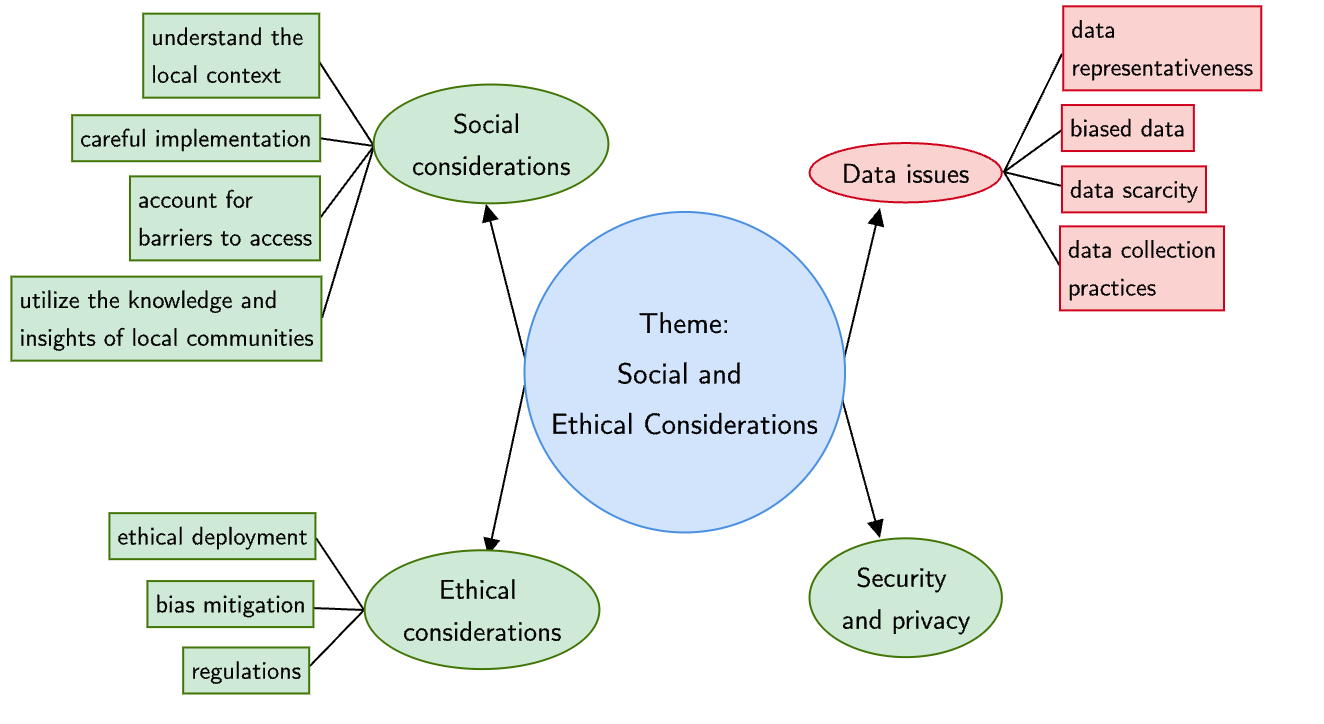}
        \caption{Major theme from the expert IDIs: \textit{Social and Ethical considerations for developing and deploying AI solutions in Health in Africa}. Summary of the subthemes illustrating the social, ethical, privacy and data-related considerations. }
        \label{fig:trust_in_ai}
\end{figure*}

\paragraph{\bf Subtheme: Social considerations:} When implementing AI in healthcare in Africa, participants emphasized the importance of accounting for many social considerations specific to the continent. \\

\textbf{\textit{Context:}} One of such considerations is the need to develop localized solutions that consider social behaviours and cultural sensitivities. This includes ``\textit{understanding the context like the social behavior, the context of healthcare in that particular country, the leadership structure as well}'' (P006 [30-39, Tanzania, Public Health Researcher]) and ``\textit{making sure the model eliminates all these}'' (P006 [30-39, Tanzania, Public Health Researcher]). Furthermore, participants underlined the necessity of accounting for diverse practices, even in seemingly simple tasks such as taking photographs of a test. P038 [30-39, USA, Machine Learning Practitioner/Engineer] illustrated this, stating, ``\textit{It's not just: take a photograph of a test. It is: take a blurry photograph, take a photograph from different backgrounds. Put the test on your jeans and take a photograph over there because people do that}''. Participants also stressed that AI should complement rather than replace the human touch in healthcare, particularly by preserving the valued bedside care and communal relationships that are important in African cultures. P033 [30-39, USA, Doctoral candidate ] stated ``\textit{I don't want AI to take the place of bedside carer \ldots  I think African cultures in general are far more communal than a lot of western European anglo-judeo Christian cultures \ldots  and I would hate to see elders and people in our community losing that. I would really hate to see them lose that bedside care, that relationship with doctor, nurse, midwife that's built over time}''.

\textbf{\textit{Utilize the knowledge and insights of local communities:}} Participants strongly advocated for community engagement during the development and deployment of AI solutions in healthcare in Africa. They described multiple benefits to involving local communities, from providing valuable insights into community health practices to ensuring reliable data collection. For example, P091 [21-29, Kenya, Health Equity Activist] noted that even ``\textit{traditional herbalists}'' could be involved to provide ``\textit{data and information on what actually goes on in terms of health, how people experience different diseases or different conditions}''. P033 [30-39, USA, Doctoral candidate ] shared similar views, stating that civil society groups and focus groups ``\textit{are really the folks who really know how to collect data from different populations, especially because they’ve been advocating for a lot of these marginalized populations for a long time}''. Moreover, participants emphasized how involving people with diverse lived experiences, including those who may not be AI professionals but possess valuable on-the-ground knowledge, is crucial for ensuring that AI solutions are relevant and effective. P038 [30-39, USA, Machine Learning Practitioner/Engineer] noted, ``\textit{One of the things is having someone local on the ground. I do believe that if you're going to have a program that's deployed in a particular community, you need someone who's actually familiar with that community involved in the program right from the get go}''. 
Additionally, participants lamented a perceived disregard towards African research and advocated for empowering and recognizing African researchers and practitioners in AI. As P033 [30-39, USA, Doctoral candidate ] shared, ``\textit{I would love to see deference to African practitioners, definitely lots of deference to them for building and executing and fairness}''. Participants overall shared that, to create solutions that meet the needs of African populations, it is essential to prioritize community engagement and foster collaboration with African researchers and practitioners. As P138 [30-39, Uganda, Electrical engineering] put it, ``\textit{We need the community, we can’t do without it}''. 

\textbf{\textit{Careful implementation and accessibility:}} Another consideration that participants brought up was the need for careful implementation to prevent AI from exacerbating existing inequities. They stressed the need to ensure that AI is deployed in a way that reaches vulnerable populations who are most in need of the technology, as well as taking steps to ensure accessibility in terms of language. P010 [30-39, Kenya, Research and Policy Analyst] noted how AI solution providers need to ``\textit{really look at having a balance between market profitability and how best to ensure that, other than profits, \ldots how does this really get into the general population who need it the most}''. Language accessibility was frequently brought up, with P046 [40-49, Kenya, Public Health Researcher] asking, ``\textit{Can we take into consideration first and foremost an AI that is able to interact using those particular languages?}'' Given the linguistic diversity in Africa, language accessibility is a crucial consideration for ensuring that AI tools are inclusive.

Furthermore, participants emphasized the need to account for the critical infrastructural challenges hindering AI adoption in Africa. Among those, participants cited inadequate internet connectivity, particularly in remote areas as a major obstacle, especially for AI applications requiring real-time online access. P118 [30-39, Zimbabwe, Machine Learning Practitioner/Engineer] explained ``\textit{access to [the] internet…in remote areas where you have to really search for the internet…because we understand some of these challenges, so you notice that some of the technology that…people are introducing to the market…they allow you to work offline and sync whenever you get access to the internet. But when it comes to maybe accessing AI…something that you need to access online in real time,…that it will be impossible in this type of setup.} The lack of suitable devices and reliable electricity further compounds this issue. 
Moreover, participants stressed the need to address the high cost of AI technologies and the computational demands of running AI systems.  As P111 [30-39, Nigeria, Medical Doctor] noted ``\textit{The cost of accessing advanced AI tools and apps can be a barrier.} P038 [30-39, USA, Machine Learning Practitioner/Engineer] followed with a personal anecdote, ``\textit{AI is expensive…a lot of the players in the AI space are for profit companies that are actually looking to make money. There are very few nonprofits..even with a nonprofit, even when we actually go and have these conversations, the first question we get is, how much is this going to cost us? Because AI is expensive, so that's a huge challenge.}

\paragraph{\bf Subtheme: Ethical considerations:} In addition to social considerations, participants emphasized the importance of responsible and ethical implementation of AI solutions to ensure fairness and avoid negative consequences. \\
\textbf{\textit{Ethical deployment:}} While the participants acknowledged that AI holds great promise for revolutionizing healthcare in Africa, potentially making it more ``\textit{accessible, efficient, and equitable}'' (P115 [21-29, Cameroon, Machine Learning Researcher]), they caution against uncritical adoption, stressing the need for ethical considerations to ensure ``\textit{safe and effective}'' (P079 [30-39, Zambia, Public Health Data Analyst]) implementation. P079 [30-39, Zambia, Public Health Data Analyst] cautiously warned, ``\textit{I think it's important to be cautious about the use of AI and machine learning in healthcare. We need to make sure that we are using these technologies in a way that is ethical and that does not have negative consequences for patients}''. Similarly P006 [30-39, Tanzania, Public Health Researcher] highlighted the importance of ethical data collection and analysis, stating ``\textit{It is important to ensure that AI is used to collect and analyze data in a way that is ethical and respectful of African communities}''.

\textbf{\textit{Bias mitigation:}} Another major concern from participants was towards the potential for bias in AI systems, particularly given the lack of representative data from Africa. They highlighted the importance of ensuring that AI solutions are unbiased as well as have bias mitigation techniques in place before and after deployment. As P115 [21-29, Cameroon, Machine Learning Researcher] put it, ``\textit{It is important to ensure that AI systems are developed and used in a way that is fair and equitable}''. P038 [30-39, USA, Machine Learning Practitioner/Engineer] echoed that sentiment, stating ``\textit{you have to ensure that before you deploy something, you are understanding what biases may exist in your model}''. They also underscored the importance of considering the potential risks of AI and taking proactive steps to mitigate them. In that sense, P030 [30-39, Nigeria, Public Health Researcher] asked ``\textit{Are we willing to cost-correct or pivot on another path, when along the way of implementation and rolling out Machine Learning models, we find out that we’ve made a mistake that is causing harm or perpetrating injustice?}''

\textbf{\textit{Regulations:}} Furthermore, concerns were also raised regarding the current lack of discussion and action surrounding AI regulation in Africa. P118 [30-39, Zimbabwe, Machine Learning Practitioner/Engineer] observed that African governments are ``\textit{not really talking about regulating these tools and how these tools should be used and the data that is used to build these tools}'' and warned about the ``\textit{the danger in the non-action from our African counterparts}''. Participants mentioned the need for regulations to put ``\textit{proper guardrails}'' (P133 [30-39, Uganda]) in the development of AI solutions in health. They called for proactive measures to regulate AI with P010 [30-39, Kenya, Research and Policy Analyst] noting the importance of ``\textit{defining priorities [and] creating the right regulatory and policy environment around AI development}''.  

\paragraph{\bf Subtheme: Data issues:} Participants overwhelmingly raised issues regarding data representativeness, scarcity, and collection methods when discussing considerations for implementing AI solutions in healthcare in Africa. 

\textbf{\textit{Data representativeness:}} Participants expressed concerns about potential biases in AI systems, particularly when applied to the African context. They emphasized that many existing machine learning models have been trained primarily on data from Western populations, leading to concerns about the applicability of these models in Africa. For example, P143 [21-29, Zimbabwe, Medical Doctor] noted, ``\textit{I feel like some of these models that we are expected to adopt in Africa, have been developed using datasets that don't really reflect our populations \ldots  So if a data set was created in Europe or in the US, the demographics of that part of that population is completely different from the demographics of let's say, Zambia, Nigeria, Kenya, South Africa. So that model is already coming in with a bias. So some of the patterns that it has picked, relationship between \ldots  certain diseases might not be applicable in my population}''. 
To mitigate bias, participants also highlighted the need for representative data which fully accounts for the diversity of the continent. AI tools must capture the continent's vast diversity, including various demographics and health challenges as P010 [30-39, Kenya, Research and Policy Analyst] stressed, ``\textit{ensure that as the tool is developed, it captures most of the groups}''. For P033 [30-39, USA, Doctoral candidate ], ``\textit{there are outliers \ldots a lot of people have not captured just how big the continent is and our diversities really just unaccounted for  \ldots so those are some other groups too that really became vulnerable to bias and stereotypes}''. Similarly, data quality was also highlighted as a major consideration when building AI solutions in Africa, given the limited access to comprehensive and representative datasets. P006 [30-39, Tanzania, Public Health Researcher] noted ``\textit{the quality of data is one of the areas that can cause a lot of bias in terms of correctness of the data but also completeness}''.

\textbf{\textit{Biased data:}} Participants cautioned that AI models may perpetuate biases and provide less accurate results for marginalized populations if trained primarily on data from advantaged groups. As P011 [40-49, Nigeria Medical Doctor] put it, ``\textit{AI comes to improve services, improve diagnosis, improve care, but that AI is only trained on data for persons who are already at an advantage and further disadvantaging the disadvantaged social class}''. This concern is compounded by potential inaccuracies in user input, leading to unreliable data that can undermine the trustworthiness of AI systems. P013 [30-39, Zambia, Public Health Researcher] shared an example to illustrate this point, saying ``\textit{We have a situation where people will answer affirmatively when asked \ldots things that they don't think are particularly important. For AI systems, that would be detrimental in the sense that if you can't trust the input of the user then it becomes really difficult to trust the outputs of the systems}''. 

\textbf{\textit{Data collection practices:}} Participants addressed existing data collections practices, lamenting that ``\textit{right now you find that data is collected not at the behest of the country or community [but it] is being collected by large organizations}'' (P013 [30-39, Zambia, Public Health Researcher]). Data collection considerations include understanding the challenges associated with the difficulty in accessing remote areas, potentially leading to biased and incomplete datasets. As P117 illustrated, ``\textit{if you're collecting data, you might find it difficult to collect data in some areas because maybe they're quite remote or they're hard to reach. So that again might end up into a bias}''. To mitigate this, participants suggest creating AI systems that are specifically designed for the African context, using locally sourced and representative data. For P143 [21-29, Zimbabwe, Medical Doctor], ``\textit{If we want to use ML significantly in Africa, we need to create datasets that are developed from Africa. A fair AI system \ldots would look like something that has been correctly sampled with \ldots representing much of the population. So that will include all these demographics, rural, peri-urban, slum, urban, women, children}''.

\textbf{\textit{Data scarcity:}} While many participants agreed that there is a data scarcity issue in Africa, some participants believe healthcare data is not inherently scarce in Africa, but accessing and using it effectively pose challenges. As P030 [30-39, Nigeria, Public Health Researcher] put it, ``\textit{I don’t think I fully agree with the fact that there’s data scarcity. I think there’s a scarcity of clean structured data}''. P059 [40-49, Kenya, Community Health Worker] echoed this sentiment, stating ``\textit{I don’t think data is scarce, what may be true is that this data has not been digitally covered}''. For P039 [40-49, Nigeria, Telemedicine and e-health researcher], while data is not scarce, reliably accessing this data is challenging. To address privacy concerns, they highlighted the importance of anonymizing medical data, making it ``\textit{available for researchers in the form that will not compromise the identity and security of the patients}''.

\paragraph{\bf Subtheme: Security and privacy:} Beyond issues related to data collection and representativeness, participants frequently discussed privacy and security considerations in AI for health in Africa. 

\textbf{\textit{Privacy:}} Participants stressed the importance of data privacy in AI healthcare applications given the pervasive ``\textit{concerns about data quality and privacy, particularly in the context of health information}''(P138 [30-39, Uganda, Electrical engineering]). They emphasized the need of data anonymization and encryption to safeguard patient privacy, as P045 [30-39, Kenya, Machine Learning Practitioner/Engineer] stated ``\textit{there has to be a way in which the data will be encrypted}''. P038 [30-39, USA, Machine Learning Practitioner/Engineer] added ``\textit{De identification is tricky, but it is possible. You want to make sure that there's no way this can be actually traced back to a person. And there are very clear rules around PII, PHI, and how you can make sure things like that are avoided}''. Participants also shared that local populations deeply understand the value of privacy. For example, P019 [40-49, France, Public health NGO chief executive ] noted that ``\textit{data privacy is key \ldots we’ve witnessed that ourselves. A young girl in an informal settlement in Kenya knows perfectly what that privacy is about \ldots she knows what the lack of privacy means, again in Kampala we all know that protecting his privacy is actually protecting his life}''. They emphasized the need to adhere to existing data privacy laws as P105 [30-39, Nigeria, Policy Maker] explained ``\textit{There are already data privacy acts in most African countries that I can think of. So, of course whatever data one is collecting has to be in line with the data privacy act of that particular country, or state, or city}''.

\textbf{\textit{Security:}} In addition to general privacy concerns, participants also expressed fear of data breaches and misuse when it comes to sensitive patient data. As P025 [50-59, Nigeria, Medical Doctor] noted, ``\textit{currently in the hospital, people are being very careful}'' and ask ``\textit{what are you doing to safeguard this data? Are you sure that the data won’t fall into the wrong hands?}''. This apprehension is compounded by the perceived lack of adequate protections for patient data, with P036 [40-49, USA, Public Health Researcher] acknowledging that \textit{there's no protections for patients and their data and what's being done with it}''.  To mitigate these concerns, some participants suggested implementing robust security protocols and policies. For example, P045 [30-39, Kenya, Machine Learning Practitioner/Engineer] believes that ``\textit{if proper systems or rather proper secure systems are put in place, and proper policies are put in place then the risk will reduce}''. Ultimately, participants highlighted the importance of tackling these security concerns to build trust in AI solutions in Africa.

\subsubsection{\bf Theme: Capacity Building for Effective Implementation and Adoption of AI Solutions in Health in Africa}
Finally, experts highlighted strategies to overcome the challenges associated with developing and integrating AI into the healthcare sector in Africa. These strategies encompass enhancing data collection and infrastructure, bridging the digital literacy gap, fostering strong local institutions, and implementing comprehensive training programs and awareness campaigns. Experts indicated that such efforts are crucial for nurturing the African AI ecosystem and building public trust for successful AI adoption in Healthcare in Africa. 

\begin{table}[!htbp]
\footnotesize
\centering
\caption{A summary of the theme and subthemes with representative quotes from the expert in-depth interviews}
\begin{tabular}{p{0.35\linewidth}|p{0.55\linewidth}}
\hline
\multicolumn{2}{c}{\textbf{THEME: Capacity Building for Effective Implementation and Adoption }} \\
\multicolumn{2}{c}{\textbf{of AI Solutions in Health in Africa}} \\ \hline
\textbf{Subtheme} & \textbf{Representative quote} \\ \hline
Dataset, Infrastructure building, and digitalization
        & ``\textit{we need to develop our own databases, our own data storage units to store our own data in Africa}'' \\ \hline
Digital Literacy
        & ``\textit{digital literacy programs, co-design it, make it multilingual, make it accessible}''\\ \hline
AI Education and Awareness 
        & ``\textit{the awareness creation activity should be based on their culture context, [and] on their educational status}''
        \\ \hline
\end{tabular}
\label{tab:theme5}
\end{table}

\begin{figure*}[!htbp]
         \centering
         \includegraphics[width=0.75\textwidth]{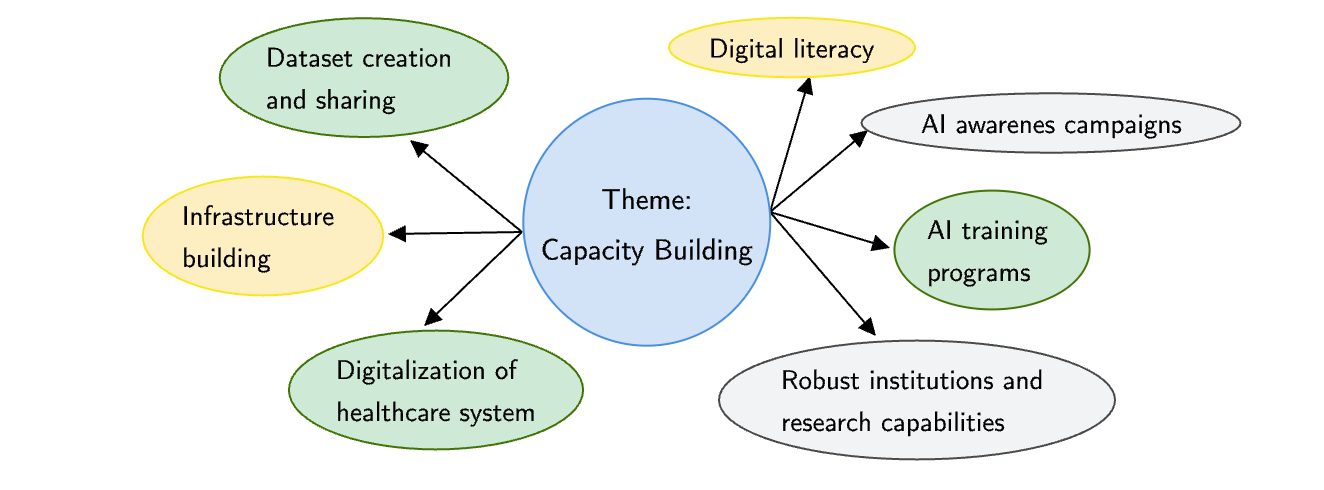}
        \caption{Major theme from the expert IDIs: \textit{Capacity building}. Summary of the subthemes illustrating the key capacity building initiatives to build an effective ecosystem for developing and adopting AI solutions in health in Africa .}
        \label{fig:trust_in_ai}
\end{figure*}

\paragraph{\bf Subtheme: Datasets, Infrastructure building and digitalization:} To overcome the data issues described in previous sections, participants advocated for a comprehensive approach to enhance data availability and quality. This includes promoting open data initiatives and encouraging data sharing among the scientific community as P028 [30-39, South Africa, Head of Engineering] noted ``\textit{I think the open data movement of trying to promote, publishing more data and making that widely available, can go a long way to building better data sets}''. To address issues of data scarcity, experts proposed several strategies: (1) healthcare providers should adopt digital technologies for efficient ``\textit{data capture, storage, retrieval and management}'' (P133 [30-39, Uganda] [30-39, Uganda]) and (2) consent mechanisms should be implemented to ensure ethical data collection and use. As P133 [30-39, Uganda] stated ``\textit{patients or people or owners of these data should be reached out to provide consent so that these data can be made available for AI training}''. They also proposed exploring alternative methods like crowdsourcing and offering incentives to motivate data contributions. 
Additionally, participants highlighted the importance of investing in local data infrastructure, including ``cloud hosting'' and data storage units. For P056 [40-49, Kenya, Public Health Researcher] ``\textit{we need to develop our own databases, our own data storage units to store our own data in Africa, then we'll be able to use this data to develop algorithms that will \ldots benefit the local communities}''. With regards to creating local datasets, participants recommended providing training on proper data collection techniques and supporting research projects focused on gathering and analyzing healthcare data. This will help ensure the availability of high-quality, representative data for AI development in Africa.
Moreover, participants stressed the importance of collaboration among African countries to ``\textit{share infrastructure and resources}'' (P029 [40-49, USA, Digital Health Lead for an NGO]). This collaborative approach, combined with ensuring accessibility to essential infrastructure, such as internet connectivity, will facilitate the widespread adoption of AI solutions. As P079 [30-39, Zambia, Public Health Data Analyst] noted, ``\textit{We need to be patient as we develop the infrastructure and expertise needed to use AI and machine learning effectively in healthcare in Africa}''.
Some participants also brought up the need to digitalize the healthcare system, recognizing the importance of electronic health records in fully utilizing AI in healthcare. P010 [30-39, Kenya, Research and Policy Analyst] noted commitments from some African leaders ``\textit{on digitalization \ldots in healthcare to improve healthcare and healthcare outcomes}''.  To that end, P029 [40-49, USA, Digital Health Lead for an NGO] suggested investing in electronic systems, noting the need to ``\textit{figure out how we can get electronic medical records that include \ldots provider narratives and patient narratives and use natural language processing to interpret that information}''. 

\paragraph{\bf Subtheme: Digital Literacy:} Participants frequently highlighted digital literacy as a prerequisite for the successful implementation and adoption of AI in healthcare in Africa. They highlighted the necessity of equipping both the general population and healthcare workers with these skills. For P030 [30-39, Nigeria, Public Health Researcher], \textit{digital literacy is a portal for empowerment, and that empowerment will enhance participation}''. 
However, many participants shared concerns about low levels of digital literacy in Africa with P073 [30-39, Nigeria, Medical Doctor] observing, ``\textit{even the most educated \ldots  are not digitally competent. So digital literacy is quite low}''. For P174 [30-39, Zimbabwe, Machine Learning Practitioner/Engineer], the ``\textit{gap within digitally enlightened people and communities and governments}''  poses a significant challenge to the application of AI interventions.
Nevertheless, participants suggested various approaches to promoting digital literacy in Africa. For P118 [30-39, Zimbabwe, Machine Learning Practitioner/Engineer], ``\textit{the digital gap is closing in Africa, but what is still, we are still trying to narrow it further. We can't really wait [for] it to close completely. It's high time we start talking about this AI technology}''. They focused on education and government involvement to ensure widespread digital literacy initiatives. They propose implementing \textit{digital literacy programs, co-design it, make it multilingual, make it accessible}'' (P030 [30-39, Nigeria, Public Health Researcher]). Particularly, participants believe governments and ministries of health should be actively involved in recognizing and promoting the importance of digital competence. For P143 [21-29, Zimbabwe, Medical Doctor], ``\textit{the Ministry has to take an active role in making sure health workers also have good digital health literacy. So this could be something that can be introduced as a course even in health programs}''. Participants also suggested ``\textit{incorporating digital literacy education in schools and universities}'' (P138 [30-39, Uganda, Electrical engineering]), particularly emphasizing the potential of AI in healthcare. 

\paragraph{\bf Subtheme: AI Education and Awareness:} Coming back to the issue of trust, several participants underscored the importance of educating the general public about AI and its potential benefits. P117 believes ``\textit{there is [a] need for people to understand what AI is all about, how it can improve their livelihood}''. P115 [21-29, Cameroon, Machine Learning Researcher] added ``\textit{The public needs to be educated about AI in healthcare. People need to understand the potential benefits and risks of AI, and they need to be able to make informed decisions about whether or not to use AI for their healthcare}''. However, in order to maximize the effectiveness of these awareness campaigns, they should be tailored to the communities they target. As P008 [30-39, Ethiopia, Public Health Researcher] put it, ``\textit{the awareness creation activity should be based on their cultural context, [and] on their educational status}''. 
Additionally, participants underscored the critical role of education and training to further equip individuals with the skills to understand and utilize AI tools effectively. They advocated for integrating AI education across various levels, from school curricula to training programs. As P028 [30-39, South Africa, Head of Engineering] noted, ``\textit{bringing training around AI into not only university level, but into STEM approaches at schools}'' is crucial, coupled with ``\textit{providing opportunities and funding around trying to build local skills}''. In the healthcare context, P115 [21-29, Cameroon, Machine Learning Researcher] emphasized the need to ``\textit{train healthcare professionals on how to use AI in a way that is safe and effective}'', ensuring they ``\textit{understand the capabilities and limitations of AI}''. To ensure accessibility, participants also stressed the importance of making such training affordable. 
Beyond education and training, participants advocated for building robust local institutions and research capabilities. They believe that collaboration across stakeholders – including users, local communities, and educational institutions – is vital in driving this development. As P010 [30-39, Kenya, Research and Policy Analyst] articulated, fostering partnerships and networking across these groups is essential for ``\textit{helping grow the capacities in Africa, the knowledge}''. P029 [40-49, USA, Digital Health Lead for an NGO] echoed this sentiment, asserting that establishing strong ``\textit{local institutions and research, whether public or private}'', is ``\textit{a key driver towards \ldots successfully implementing and maintaining AI}''.